\documentclass[mnsc,nonblindrev]{inform3_zp}
\usepackage{amssymb}
\usepackage{graphicx}
\usepackage{subfigure}
\usepackage{epstopdf}
\usepackage{multirow}
\usepackage[subnum]{cases}
\usepackage{color}
\usepackage{threeparttable}
\usepackage{algorithm}
\usepackage{algpseudocode}
\usepackage[normalem]{ulem}
\usepackage{float}
\usepackage{bm}
\usepackage{amsmath}
\usepackage{mathtools}
\usepackage{mathrsfs}
\usepackage{dsfont}
\usepackage[numbers,angle]{natbib}
\usepackage{url}
\usepackage{bbm}
\newcommand*{\QED}{\hfill\ensuremath{\square}}

\allowdisplaybreaks

\OneAndAHalfSpacedXI 

\usepackage{natbib}
 \bibpunct[, ]{(}{)}{,}{a}{}{,}%
 \def\bibfont{\small}%
\TheoremsNumberedThrough     
\ECRepeatTheorems

\MANUSCRIPTNO{}

\begin{document}



\RUNTITLE{Mostly Beneficial Clustering: Aggregating Data for Operational Decision Making}

\TITLE{Mostly Beneficial Clustering: Aggregating Data for Operational Decision Making}

\ARTICLEAUTHORS{%
\AUTHOR{Chengzhang Li}
\AFF{Antai College of Economics and Management, Shanghai Jiao Tong University, \EMAIL{cz.li@sjtu.edu.cn}} 
\AUTHOR{Zhenkang Peng}
\AFF{Antai College of Economics and Management, Shanghai Jiao Tong University, \EMAIL{shoucheng@sjtu.edu.cn}}
\AUTHOR{Ying Rong}
\AFF{Antai College of Economics and Management, Shanghai Jiao Tong University, \EMAIL{yrong@sjtu.edu.cn}}
} 

\ABSTRACT{%
With increasingly volatile market conditions and rapid product innovations, operational decision-making for large-scale systems entails solving thousands of problems with limited data. Data aggregation is proposed to combine the data across problems to improve the decisions obtained by solving those problems individually. We propose a novel cluster-based Shrunken-SAA approach that can exploit the cluster structure among problems when implementing the data aggregation approaches. We prove that, as the number of problems grows, leveraging the given cluster structure among problems yields additional benefits over the data aggregation approaches that neglect such structure. When the cluster structure is unknown, we show that unveiling the cluster structure, even at the cost of a few data points, can be beneficial, especially when the distance between clusters of problems is substantial. Our proposed approach can be extended to general cost functions under mild conditions. When the number of problems gets large, the optimality gap of our proposed approach decreases exponentially in the distance between the clusters. We explore the performance of the proposed approach through the application of managing newsvendor systems via numerical experiments. We investigate the impacts of distance metrics between problem instances on the performance of the cluster-based Shrunken-SAA approach with synthetic data. We further validate our proposed approach with real data and highlight the advantages of cluster-based data aggregation, especially in the small-data large-scale regime, compared to the existing approaches.

}%



\maketitle

%

\section{Introduction}

With the growing availability and accessibility of data, data-driven decision-making plays an increasingly critical role in various operational practices \citep{BertsimasKallus2020,  MisicPerakis2020, QiMakShen2020, ZhangChaoShi2020, ChenChaoShi2021, FengShanthikumar2022}. Many data-driven approaches are developed to facilitate operational decision-making in the big-data regime \citep[e.g.,][]{HaskellJainKalathil2016, BanRudin2019, ChenSongWei2020, WangCurrimRam2022, Dutta2022}. However, with the ever-changing commercial landscape and the unprecedented need for rapid product innovation, it is not uncommon that only a limited number of data can be acquired and analyzed by firms for operational decision-making \citep{LimShanthikumarShen2006, GuptaRusmevichientong2021}. For instance, when firms launch new products, especially for those with relatively short life cycles, only a few sales data points can be collected and utilized for inventory planning in the upcoming selling horizon. At the same time, firms often manage thousands of operations management problems simultaneously \citep{BayatiCaoChen2022}. For example, Shein, an ultra-fast-fashion brand, launches 6,000 new items per day \citep{HBR2022, Wired2022}. Not only does the firm need accurate demand predictions for multiple products, but also effective procurement and inventory management strategies to achieve operational efficiency. This gives rise to the tremendous need for operational decision-making under the small-data large-scale regime \citep{GuptaRusmevichientong2021, FengLiShanthikumar2023}.

In practice, firms could adopt one of three distinct approaches to deal with challenging decision-making with limited data. The first approach treats each problem in isolation, relying solely on respective data for operational decision-making. However, the decision quality of this decoupling approach may be compromised due to data scarcity. For instance, Sample Average Approximation (SAA) for each problem can recover the optimal solutions with a sufficient amount of data \citep{ShapiroDentchevaRuszczynski2021}, but may perform poorly with limited data \citep{BesbesMouchtaki2023}.
The second approach combines data from all problems to derive a shared estimate applicable to all. While this approach effectively tackles the data scarcity issue that hinders the first approach, it introduces potential bias in the estimate for each problem due to inherent differences among them.
The third approach segments problems into clusters and subsequently combines data within each cluster to derive a common estimate for problems within the same cluster. This approach alleviates bias introduced by the second strategy, yet its effectiveness hinges on the degree of similarity among problems within the same cluster.


In light of the operational challenges faced by firms and limitations of those three conventional approaches in the small-data large-scale regime, some recent advancements in data analytics emphasize the necessity of developing data-driven approaches to obtain effective decisions with limited data by leveraging additional statistical structures among those problems \citep[e.g.,][etc.]{GuptaRusmevichientong2021, Bastani2021, GuptaKallus2022, FengShanthikumar2023}. Data aggregation with clustering (DAC), proposed by \cite{CohenZhangJiao2022} aiming to improve product demand prediction, employs hypothesis testing to ascertain whether estimates should be obtained at the individual product level, across all products, or at the cluster level. If hypothesis testing indicates that an estimate should be derived by combining data across all products or within the same cluster, the resulting estimate will be the same across all products or within the cluster, respectively. Taking a different perspective, \cite{GuptaKallus2022} proposes a novel Shrunken-SAA algorithm (referred to as {\em direct data pooling} hereafter) that determines operational decisions with a weighted combination of the data from the individual problem and the aggregated data from all problems. Instead of providing a uniform estimate when combining data from all problems or at the cluster level, the weighting mechanism in the direct data pooling approach not only alleviates bias arising from data aggregation approaches but also tackles the data scarcity issue encountered in decoupling approaches. However, in instances where the underlying differences among problems are considerable, the value of direct data pooling with all problems may be limited \citep[see also the discussion in Section \ref{sec_DataPooling_MSE}]{GuptaKallus2022, FengLiShanthikumar2023}.


Inspired by the existing developments, we propose a {\em cluster-based data pooling} approach that can exploit the cluster structure among problems and carefully choose the amount of data for aggregation. Cluster-based data pooling involves a two-phase procedure. It first estimates the cluster structure by some clustering algorithm. After identifying which problems are aggregated together, direct data pooling is implemented to problems within each cluster. We present the performance of our proposed approach benchmarked with the existing approaches in Figure \ref{Fig_AlgorithmComparisionMSE}. Leveraging the cluster structure can generate additional benefits over direct data pooling. Furthermore, within each cluster, the carefully shrunken SAA solutions outperform that of the DAC approach, which mandates that estimates be identical within the cluster. Research questions naturally arise: does cluster-based data pooling always provide additional benefits over direct data pooling? In addition, can the cluster-based approach be extended to general cost functions that are representative of various operations management settings?

We first analytically investigate the performance of the cluster-based data pooling approach under the commonly used cost function, mean square error (MSE). When the cluster structure of problems is given, we show that, as the number of problems grows, the cluster-based data pooling approach always provides additional benefits over the direct data pooling approach that neglects such structure. This provides implications for practitioners who can incorporate their domain knowledge, e.g., the cluster structure of problems, in the implementation of data pooling approaches. 

When cluster structure is unknown, identifying such structure requires a certain amount of data points as input and reduces the sample size for the subsequent data pooling procedure. We find that, when the distance between clusters is substantial, the clustering analysis can still generate additional benefits at the cost of using a few data points for clustering. Nevertheless, when the distance between clusters is small, e.g., the distance between clusters is zero, the cluster-based data pooling approach cannot outperform direct data pooling. This suggests that cluster-based data pooling may not necessarily outperform direct data pooling, for example, when the underlying distributions for problems are relatively concentrated.

We further extend our cluster-based data pooling approach to general cost functions under mild conditions. The performance of our proposed approach is measured by the optimality gap compared with the cost of an oracle who knows the underlying cluster structure and tunes the shrinkage parameter with the true distribution information. We show that when the number of problems gets large, the optimality gap decreases exponentially in the distance of clusters. This provides theoretical guarantees on the performance of our cluster-based data pooling approach in numerous operations management settings.

To further validate the cluster-based data pooling approach, we explore its benefits for newsvendor systems using both synthetic and real data. Two key aspects of flexibility in implementing cluster-based data pooling are examined: the choice of distance measure between problem instances and the determination of the number of clusters for data aggregation. 
Our investigation begins by assessing the impact of different metrics measuring the distance between problem instances on the performance of cluster-based data pooling. Surprisingly, we observe that the distance metric suggested by the newsvendor cost, that is, the difference between sample quantiles of observed demand data, does not necessarily identify a more favorable cluster structure than the sample mean-based distance metric in a data-driven setting. When the coefficients of variation associated with those problems are the same, the cluster structures obtained by two distance metrics with true distribution information are the same, while the latter can provide a more robust estimate, especially with limited data. Subsequently,  since the number of clusters is unknown in practice a priori, we devise a bisect clustering algorithm that automatically determines the number of clusters. Utilizing real data, we demonstrate that the cluster-based data pooling outperforms the direct one, especially when the number of problems is large and the number of data points for each problem is small. This highlights the advantages of our proposed approach that flexibly harnesses the cluster structure and carefully aggregates the data within the cluster in the small-data large-scale regime. 

The remainder of the paper is organized as follows. In Section \ref{sec_Literature}, we review the related studies. In Section \ref{sec_Setup}, we introduce the problem setup.  In Section \ref{sec_DataPooling_MSE}, we present the direct data pooling approach under MSE. In Section \ref{sec_ClusterMSE}, we propose the cluster-based data pooling approach and demonstrate the benefits of clustering over data pooling under MSE. In Section \ref{sec_ClusterGeneral}, we investigate the cluster-based data pooling approach under the general cost functions. In Section \ref{sec_Numerical}, we conduct numerical experiments with synthetic and real data to validate our approach. We conclude the paper in Section \ref{sec_conclusion}. 

\section{Literature Review}\label{sec_Literature}

Our work relates to the stream of literature on small-data and large-scale learning and data-driven optimization, and the stream of studies on cluster-based learning and decision-making.

{\bf Small-data and large-scale regime.}
The goal of our work is to solve simultaneously a large number of problems where the available data for each problem is limited. This problem setup is defined as the small-data large-scale regime by \cite{GuptaRusmevichientong2021}. One classical example is the simultaneous estimation of multiple Gaussian means. \cite{Stein1956} proposes a data pooling procedure that outperforms the decoupled procedure with respect to the total expected estimation error, where the latter adopts the sample mean as the estimate for the corresponding distribution. While many follow-up works in Statistics aim to explain Stein's phenomenon \citep[]{Brown1971,EfronMorris1977,Stigler1990,Beran1996,Brown2012}, \cite{GuptaKallus2022} proposes a novel Shrunken-SAA algorithm that generalizes Stein's method to data-driven optimization problems. Their approach requires mild distributional assumptions and can be applied to various optimization problems. As the number of problems grows, the Shrunken-SAA algorithm is shown to outperform the decoupled approaches under the mean squared loss. There are several other studies that leverage the idea of shrinkage to solve data-driven optimization problems with specific optimization problem structures \citep[]{Jorion1986, DeMiguelMartin-UtreraNogales2013, DavarniaCornuejols2017} or statistical conditions such as Gaussian or near-Gaussian assumptions \citep[]{MukherjeeBrownRusmevichientong2015, GuptaRusmevichientong2021}. The idea of shrinkage in \cite{GuptaKallus2022} is also related to Stacking in Machine Learning \citep{DvzeroskiZenko2004, Pavlyshenko2018, JiangLiHaqSaboorAli2021}. While  \cite{GuptaKallus2022} chooses the shrinkage parameter that determines how to combine the data set of one problem with those of other problems, in Stacking, the meta learner is trained to combine the predictions from the base learners that are obtained with the same data set. Our work is closely related to \cite{GuptaKallus2022} pointing out that the benefit of data pooling can be limited when the underlying distributions are dispersed. We propose to implement a clustering analysis to identify the underlying problem similarity structure before pooling the data, aiming to further enhance the advantage of data pooling approaches over the decoupled ones.

Our work is also related to Multi-task learning (MTL) in Machine Learning, which improves the generalization ability by utilizing domain information from related tasks, potentially using a shared representation to enhance learning outcomes across tasks \citep{Caruana1997}. The multi-task learning has been used in various applications, including speech recognition \citep{LuLuSehgalGuptaDuThamGreenWan2004}, computer vision  \citep{CaruanaO'Sullivan1998}, and medical diagnosis \citep{AmyarModzelewskiLiRuan2020}. This stream of studies aims to learn the shared representation across prediction tasks, for example, in the form of neural networks, and emphasize the improved prediction outcomes. Our work, on the other hand, focuses on data-driven optimization problems and provides theoretical guarantees for cluster-based data pooling approaches for decision-making of multiple problems. Transfer learning, a special case of multi-task learning, improves a learner from one domain by transferring information from a related domain \citep[]{WangMahadevan2011, DuanDongTsang2012, HarelMannor2011, NamKim2015, ZhouTsangPanTan2014}. \cite{Bastani2021} proposes a novel two-step estimator to efficiently combine a large amount of proxy data and a small amount of target data, and achieves the same accuracy for the target task with exponentially less amount of data. \cite{BastaniSimchi-LeviZhu2022} apply transfer learning to the problem of learning shared structure via a sequence of dynamic pricing experiments for related products. \cite{NabiNassifHongMamanimbens2022} propose a hierarchical
empirical Bayes approach to learn empirical meta-priors, which are used to decouple the learning rates of first-order and second-order features in a generalized linear model.   Different from those studies based on transfer learning, our work focuses on solving multiple problems simultaneously, and each of the problems can access only a limited number of data. \cite{FengLiShanthikumar2023} develop cross-learning solutions under the parametric Operational Data Analytics (ODA) framework that utilizes the ample data from related systems and demonstrates the proposed solution outperforms any transfer learning solution. For multiple systems with limited data, they derive the co-learning solutions that are asymptotically optimal for both the aggregate system and individual systems. \cite{FengLiShanthikumar2023} assumes the random variable associated with an individual system is known up to its scale, and the objective function satisfies the homogenous property. Our approach does not require the statistical knowledge of the distribution family and allows for implementation under general convex cost functions. 

{\bf Cluster-based learning and decision-making.} 
Clustering analysis, an unsupervised learning paradigm including the K-means clustering \citep{MacQueen1967} and Hierarchical clustering \citep{Nielsen2016}, is adopted as a common exploratory data analysis with applications in image segmentation \citep{ColemanAndrews1979}, bioinformatics \citep{KarimBeyanZappaCostaRebholz-SchuhmannCochezDecker2021}, community detection \citep{MalliarosVazirgiannis2013}, and etc. Many recent works also leverage clustering analysis to improve the performance of supervised learning algorithms. \cite{DuanCuiQiaoYuan2018} proposes a cluster-based supervised learning framework that identifies the exemplars, the data instance closest to the cluster center, and partitions the data into different classes via discriminative learning. \cite{PeikariSalamaNofech-MozesMartel2018} propose a cluster-then-label method to identify high-density regions in the feature space, and then implement the support vector machines for prediction tasks. Furthermore, different from the sequential clustering and supervised learning of the above literature, \cite{ChenXie2022} proposes a cluster-aware supervised learning (CluSL) framework, which integrates the clustering analysis with supervised learning. CluSL aims to achieve dual objectives: identifying optimal clusters for a data set while concurrently minimizing the cumulative loss functions within each cluster. Our proposed approach estimates the cluster structure and derives the data-driven decisions sequentially, which renders more interpretability. Furthermore, our work not only provides an asymptotic analysis of the cluster-based approach but also decomposes the optimality gap into components associated with the clustering analysis and that of the data pooling process. This illustrates a delicate trade-off between the benefit of exploring the problem similarity structure and that of data pooling. 

Cluster-based learning and decision-making have received increasing attention in operations management and management science applications. \cite{CohenZhangJiao2022} propose the Data Aggregation with Clustering (DAC) approach to balance the trade-off between the benefit of data aggregation and that of model flexibility, and show that the DAC algorithm can improve prediction accuracy compared to a wide range of benchmarks with both synthetic and real data. \cite{BernsteinModaresiSaure2019} propose a dynamic clustering policy, which estimates the mapping of customer profiles to clusters as well as the preference parameters for each cluster by adaptively adjusting customer segments. The case study based on a real data set suggests that the benefit of the dynamic clustering policy can be substantial and result in more than 37\% extra revenues. \cite{LiuHeMaxShen2021} apply a hierarchical clustering algorithm to each batch of orders based on geographical proximity to predict the travel time of last-mile delivery services.  \cite{MiaoChenChaoLiuZhang2022} propose pricing policies that concurrently perform clustering analysis over product demands and set individual pricing decisions on the fly. They show that online clustering is an effective approach to tackle dynamic pricing problems associated with low-sale products and validate their pricing policies using a real data set from Alibaba. \cite{LiLamPeng2022} studies a context-dependent optimization problem for selecting the best design. They devise a sampling procedure to learn both the global cluster information and local design information based on a Gaussian mixture model and prove the consistency and the asymptotic optimality of the proposed sampling policy. \cite{BhatLyonsShiYang2022} present a framework for trust-aware sequential decision-making in a human-robot team and cluster the participants’ trust dynamics into three categories according to their personal characteristics. Different from these aforementioned papers, our proposed approach can be applied to solve multiple optimization problems with general cost functions. More importantly, our approach is also applicable to the setting with limited data as we leverage the data pooling approach for making decisions within each cluster.




\section{Problem Setup}\label{sec_Setup}
Consider the stochastic optimization problem
\begin{equation}\label{eqn_SP}
 \min_{x \in \mathcal{X}} \quad \mathbb{E}_{\xi \sim \mathbb{P}}[c(x,{\xi})],
\end{equation}
where the probability measure $\mathbb{P}$ governs the random variable ${\xi}$ and $c(x, \hat{\xi})$ is the cost associated with the decision $x \in \mathcal{X}$ and the realized random variable $\hat{\xi}$. 
In practice, one may need to solve thousands of unrelated stochastic optimization problems in the form of \eqref{eqn_SP} simultaneously. Let $\mathcal{K} = \{1,2,\ldots,K\}$ denotes the set of $K$ problems. 
\begin{equation}\label{eqn_SPLargeScale1}
    \min_{x_k \in \mathcal{X}_k, k \in \mathcal{K}}\quad \frac{1}{K} \sum_{k=1}^K \mathbb{E}_{\xi_k \sim \mathbb{P}_k}[c_k(x_k, {\xi}_k)],
\end{equation}
where $x_k \in \mathcal{X}_k$ denotes the decision variable and ${\xi}_k$ the random variable, and $c_k(x_k, \hat{\xi}_k)$ is the cost associated with decision $x_k$ and realized random variable $\hat{\xi}_k$ for the $k$-th problem. As there are no coupling constraints across the feasible regions $\mathcal{X}_k$, and the random variables ${\xi}_k$ for each problem are independent, equivalently, we can solve each problem separately. 
\begin{equation}\label{eqn_SPLargeScale2}
     \frac{1}{K} \sum_{k =1}^K \min_{x_k \in \mathcal{X}_k} \mathbb{E}_{\xi_k \sim\mathbb{P}_k}[c_k({x}_k, {\xi}_k)].
\end{equation}
In data-driven settings, the probability measures of random variables, denoted by $\{{\mathbb{P}}_k$, $k\in \mathcal{K}\}$, are typically unknown, and one may have access to data sets of $N$ samples, $S_k = \{\hat{\xi}_{k,1}, \hat{\xi}_{k,2}, \ldots, \hat{\xi}_{k,N}\}$, for each subproblem $k \in \mathcal{K}$.\footnote{For brevity, we assume that the number of observations for each problem is the same. Our analysis can be extended to the case where the numbers of observations for each problem are different.} The goal is to find the mapping from these data to the feasible regions to minimize the objective function defined in \eqref{eqn_SPLargeScale1}. 

A canonical policy that one can adopt is Sample Average Approximation (SAA) where the probability measure $\mathbb{P}_k$ is approximated by the empirical distribution of the observed values $S_k = (\hat{\xi}_{k,1}, \hat{\xi}_{k,2}, \ldots, \hat{\xi}_{k,N})$.
\begin{equation}
    x_k^\mathrm{SAA}(S_k) \in \argmin_{{x}_k \in \mathcal{X}_k} \frac{1}{N}\sum_{i=1}^N c_k({x}_k, \hat{\xi}_{k,i}). 
\end{equation}

In light of the form of Equation \eqref{eqn_SPLargeScale2}, one may decouple the problem, and apply data-driven procedures, such as SAA, separately to each subproblem. However, \cite{GuptaKallus2022} proposes the data pooling approach that shrinks the SAA solutions by parameters estimated from pooling data across subproblems and finds that it can outperform the decoupling approach, even if the subproblems are unrelated and the random variables are independent. 

Before proceeding to the analysis, we elaborate on and clarify the evaluation metrics of the data-driven decisions. For a given problem indexed by $k$, we validate the data-driven decision $x_k(S_k)$  by computing $\mathbb{E}_{{\xi}_k}[c_k( x_k(S_k), {\xi}_k)]$, which is referred as the {\em out-of-sample cost}. The expectation is taken with respect to the test sample ${\xi}_k$ given the decision $x_k(S_k)$. To evaluate the average performance of the decision $x_k(S_k)$ that depends on historical datasets $S_k$, we consider the {\em expected out-of-sample cost}, that is,  $\mathbb{E}_{S_k}[\mathbb{E}_{\xi_k}[c_k( x_k(S_k), {\xi}_k)]]$, where the outer expectation is taken with respect to the observed samples \citep[e.g.,][]{BesbesMouchtaki2023}. 

{
\section{Data Pooling with Mean Squared Error (MSE)}\label{sec_DataPooling_MSE}
In this section, we present the data pooling approach proposed by \cite{GuptaKallus2022} and illustrate the potential benefits and limitations of data pooling under the mean squared error (MSE). 
In particular, \cite{GuptaKallus2022} proposes to shrink the empirical distribution of limited observed data with an anchor distribution, denoted by $\mathbb{P}_0$, and the degree of shrinkage is controlled by a parameter denoted by $\alpha > 0$. Specifically, the decision for problem $k$ is derived by solving the following optimization problem. 
\begin{equation}
   x_k(\alpha, \mathbb{P}_0, S_k) \in \argmin_{x_k \in \mathcal{X}_k}  \sum_{i=1}^N c_k({x}_k, \hat{\xi}_{k,i}) + \alpha \cdot \mathbb{E}_{ \mathbb{P}_0}[c_k({x}_k,{\xi})].
\end{equation}
Under MSE, the data pooling approach essentially prescribes the following class of policies that are parameterized by $\alpha$ and $\mu_0$, where $\mu_0$ is the mean of the anchor distribution $\mathbb{P}_0$.
\begin{equation}\label{eqn_ShrunkenSAAMSE}
    x_k(\alpha, \mu_0, \hat{\mu}_k) = \frac{N}{N+\alpha} \hat{\mu}_k +  \frac{\alpha}{N+\alpha} \mu_0.
\end{equation}
For ease of exposition, under  MSE, the anchor distribution is substituted by the corresponding mean $\mu_0$ and the dataset $S_k$ is substituted by the sample mean $\hat{\mu}_k$ in the decision function $x_k$. From the above equation, the data pooling approach shrinks the sample mean, $\hat{\mu}_k$, towards the mean of the anchor distribution, $\mu_0$, via the shrinkage parameter $\alpha$. In particular, as $\alpha \to 0\mbox{ }(\infty)$, $x_k(\alpha, \mu_0, \hat{\mu}_k)$ converges to the sample mean (anchor mean).

Given a shrinkage parameter $\alpha$ and an anchor mean $\mu_0$, the total expected out-of-sample cost of all problems in $\mathcal{K}$ with the data pooling approach, denoted by $Z(\alpha, \mu_0, \mathcal{K})$, is computed as
\begin{equation}
    Z(\alpha, \mu_0, \mathcal{K}) = \sum_{k=1}^K\mathbb{E}_{S_k}[\mathbb{E}_{\xi_k}[c_k(x_k(\alpha, \mu_0, \hat{\mu}_k), \xi_k)]].
\end{equation}
If the mean and variance of the random variable $\xi_k$, denoted by $\mu_k$ and $\sigma_k^2$, are known, the above expected out-of-sample cost can be explicitly expressed as, $Z(\alpha,\mu_0, \mathcal{K})  = \sum_{k=1}^K \sigma_k^2 +  \sum_{k=1}^{K}\Big[\Big(\frac{\alpha}{N+\alpha}\Big)^2(\mu_k-\mu_0)^2+\Big(\frac{N}{N+\alpha}\Big)^2 \frac{\sigma_k^2}{N} \Big].$
When the anchor mean $\mu_0$ is given, we can minimize the out-of-sample cost over $\alpha$, which derives the following shrinkage parameter,
\begin{equation}\label{eqn_OptAlphaPooling}
    \alpha^\mathrm{AP}_{\mu_0} = \frac{\sum_{k=1}^K \sigma_k^2}{\sum_{k=1}^K (\mu_k-\mu_0)^2} >0,
\end{equation}
where $\mathrm{AP}$ stands for a priori, meaning $\alpha^\mathrm{AP}_{\mu_0}$ is the optimal a priori choice of shrinkage parameter before observing any data given the anchor mean $\mu_0$. 
With the optimal parameter $\alpha^\mathrm{AP}_{\mu_0}$, the expected out-of-sample cost of the data pooling approach is
\begin{equation} \label{eqn_GuptaCost}
    Z(\alpha^\mathrm{AP}_{\mu_0},\mu_0, \mathcal{K}) = \sum_{k=1}^{K}\sigma_k^2+  \Big(\frac{1}{N} \sum_{k=1}^{K}  \sigma_k^2 \Big)\frac{N}{N+\alpha^\mathrm{AP}_{\mu_0}} =  \sum_{k=1}^{K}\sigma_k^2 + \frac{(\frac{1}{N}\sum_{k=1}^K \sigma_k^2)(\sum_{k=1}^K (\mu_k - \mu_0)^2)}{\frac{1}{N}\sum_{i=1}^K \sigma^2_k + \sum_{k=1}^K (\mu_k - \mu_0)^2}.  
\end{equation}

To demonstrate the benefit of data pooling, we consider solving each problem separately via the SAA approach as the benchmark. When the cost function is in the form of mean squared error, the SAA solution is the sample mean for each subproblem, that is, $x_k^\mathrm{SAA}(S_k) = \hat{\mu}_k = \frac{1}{N} \sum_{i=1}^N \hat{\xi}_{k, i}, k \in \mathcal{K}.$
A direct computation gives the expected out-of-sample cost of SAA solutions, $Z_\mathrm{SAA}=\sum_{k=1}^K\mathbb{E}_{S_k}[\mathbb{E}_{\xi_k}[c_k(x_k^\mathrm{SAA}(S_k), \xi_k)]] =  \sum_{k=1}^{K}\big(\sigma_k^2+\frac{\sigma_k^2}{N}\big).$
Comparing the expected out-of-sample costs of the SAA solutions and the Shrunken-SAA solutions,
\begin{equation} 
Z_\mathrm{SAA} - Z(\alpha_{\mu_0}^\mathrm{AP}, \mu_0, \mathcal{K}) = \Big(\frac{1}{N}\sum_{k=1}^{K} \sigma_k^2 \Big)\frac{\alpha^\mathrm{AP}_{\mu_0}}{N+\alpha^\mathrm{AP}_{\mu_0}} =\frac{(\frac{1}{N}\sum_{k=1}^K \sigma_k^2)^2}{\frac{1}{N}\sum_{i=1}^K \sigma^2_k + \sum_{k=1}^K (\mu_k - \mu_0)^2} > 0. \label{shrunken_saa_compare}
\end{equation}
This suggests that the benefit of the data pooling approach is strictly positive if one can select the optimal shrinkage parameter $\alpha_{\mu_0}^\mathrm{AP}$ for any given anchor mean $\mu_0$. Since the benefit of data pooling increases as the shrinkage parameter increases, the anchor that yields the lowest possible out-of-sample cost is the average of the true means of each subproblem, i.e., $\mu_{0}^{\mathrm{AP}} = \frac{1}{K} \sum_{k=1}^K \mu_{k}$. Replacing $\mu_0$ by $\mu_{0}^{\mathrm{AP}}$, the optimal a priori shrinkage parameter is,
\begin{equation}
\alpha^\mathrm{AP} = \frac{\sum_{k=1}^K \sigma_k^2}{\sum_{k=1}^K (\mu_k-\mu_0^\mathrm{AP})^2} = \frac{\sum_{k=1}^K \sigma_k^2}{\sum_{k=1}^K (\mu_k-\frac{1}{K} \sum_{k=1}^K \mu_{k})^2}.
\end{equation}

While we have demonstrated the potential benefit of the data pooling approach over SAA by Equation \eqref{shrunken_saa_compare}, the impacts of shrinking the SAA solutions on the subproblems are not uniformly beneficial. Therefore, we unfold the discussion on the benefit or loss incurred for each problem resulting from shrinking the SAA solutions. Suppose one implements the data pooling approach with the optimal a priori  shrinkage parameter and anchor distribution, $(\alpha^\mathrm{AP}, \mu_0^\mathrm{AP})$, the expected out-of-sample cost for problem $k$ is, 
\begin{equation}
     Z(\alpha^\mathrm{AP}, \mu_0^\mathrm{AP}, k) =\underbrace{\vphantom{\bigg(\Big(\frac{N}{N+\alpha^\mathrm{AP}}\Big)^2 -1 \bigg) \frac{\sigma_k^2}{N}}\frac{\sigma_k^2}{N} +\sigma_k^2}_{\text{SAA cost}} +  \underbrace{\bigg( \frac{\alpha^\mathrm{AP}}{N+\alpha^\mathrm{AP}}\big(\mu_k-\mu_0^\mathrm{AP}\big)\bigg)^2}_{\text{Bias induced by shrinkage}}-\underbrace{\bigg(1 - \Big(\frac{N}{N+\alpha^\mathrm{AP}}\Big)^2  \bigg) \frac{\sigma_k^2}{N}}_{\text{Variance reduced by shrinkage}}. \label{cost_decompose}
\end{equation}
As indicated by the above equation, the difference between the cost of the data pooling approach and that of the SAA approach boils down to a trade-off between the bias induced by shrinkage and the variance reduced by shrinkage with the shrinkage parameter $\alpha^\mathrm{AP}$ and the anchor mean $\mu_0^\mathrm{AP}$. Therefore, whether the data pooling approach can reduce the expected out-of-sample cost for the problem $k$ depends on the magnitudes of the impacts of shrinkage on bias and variance.
}

\begin{figure}[htbp!]
\centering
\subfigure[Shrinking concentrated means]{
\begin{minipage}[t]{0.33\linewidth}
\centering
\includegraphics[width=2.2in]{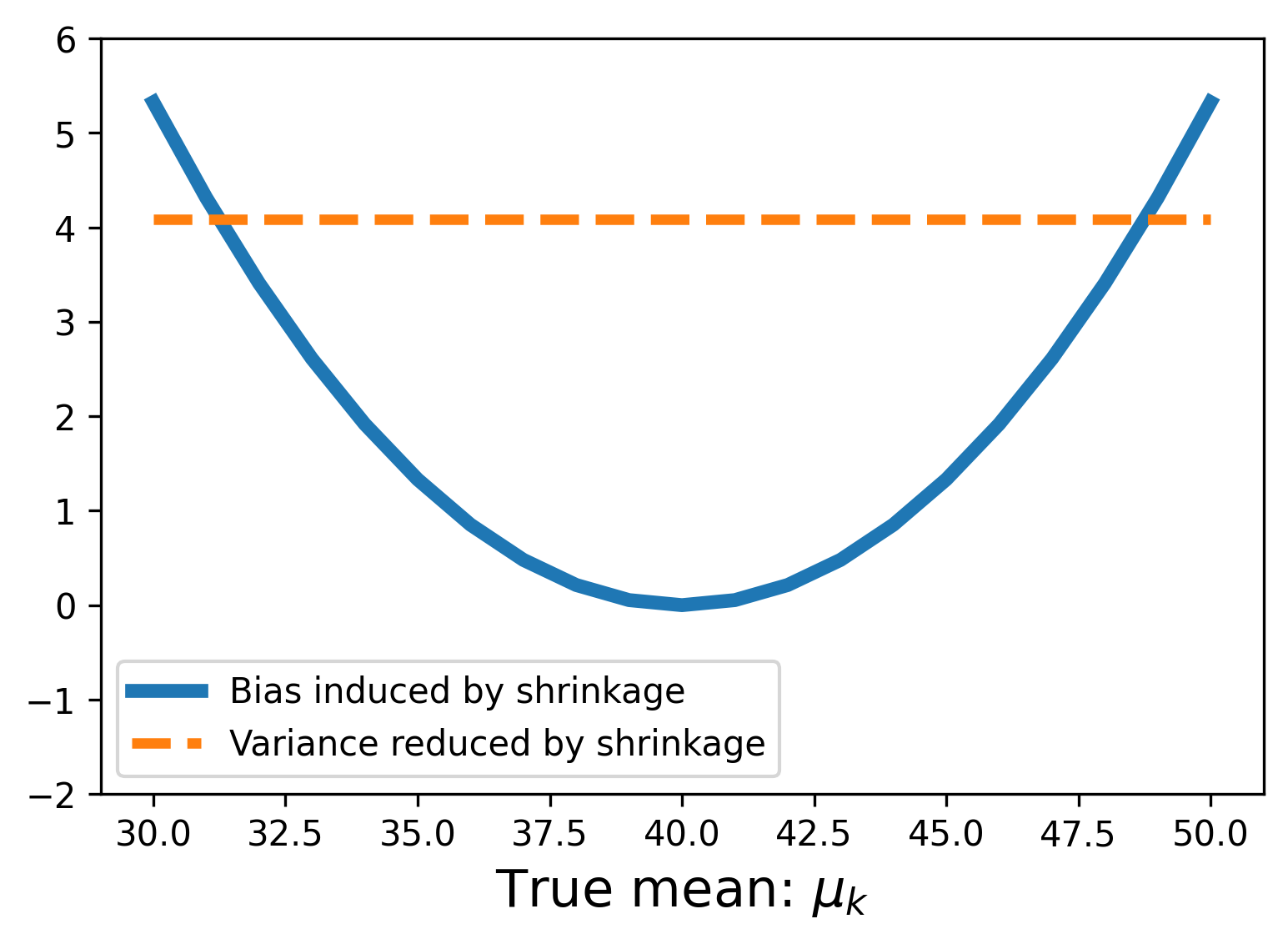}
\end{minipage}%
}%
\subfigure[Shrinking dispersed means]{
\begin{minipage}[t]{0.33\linewidth}
\centering
\includegraphics[width=2.2in]{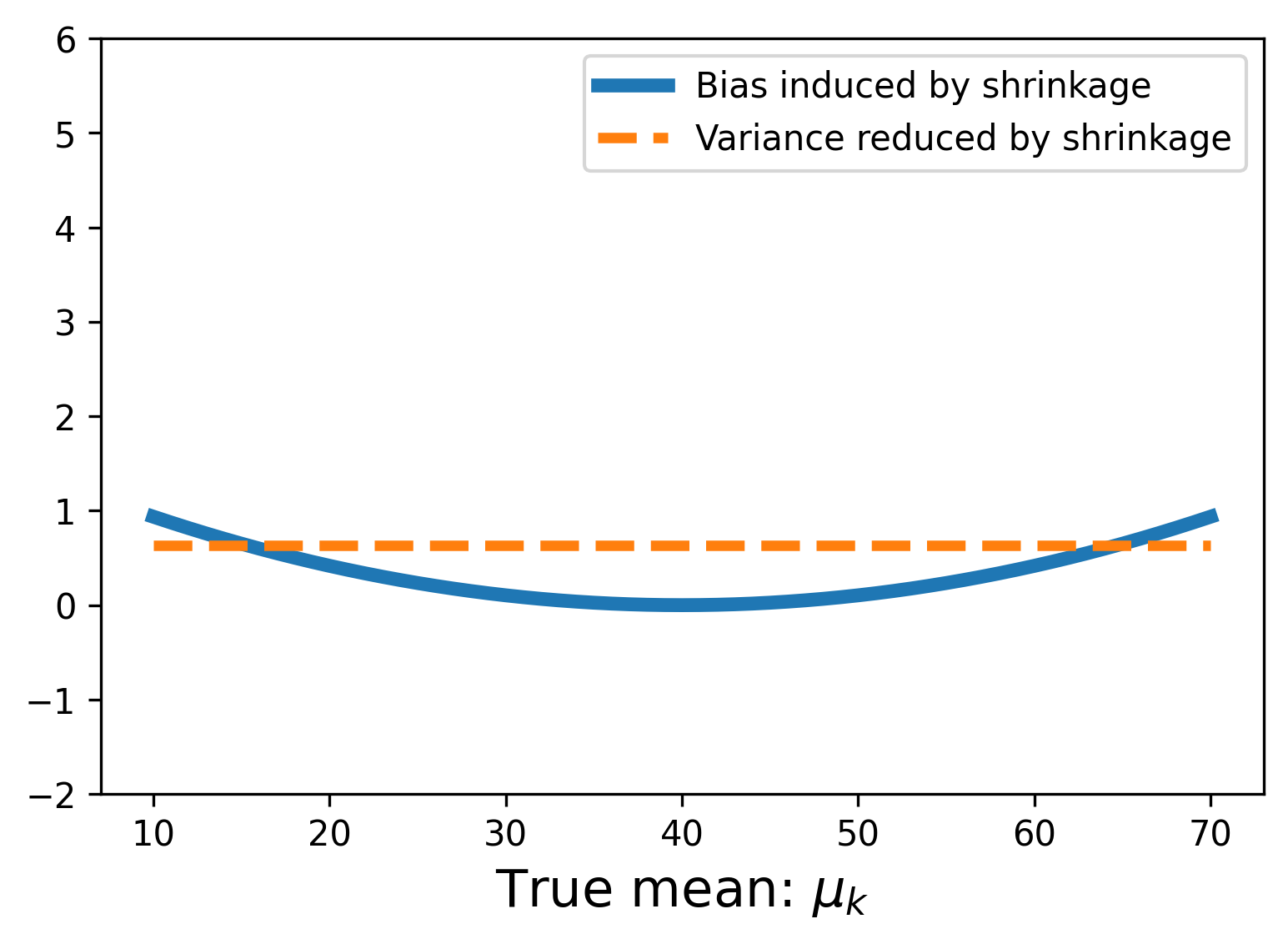}
\end{minipage}%
}%
\subfigure[Shrinking clustered means]{
\begin{minipage}[t]{0.33\linewidth}
\centering
\includegraphics[width=2.2in]{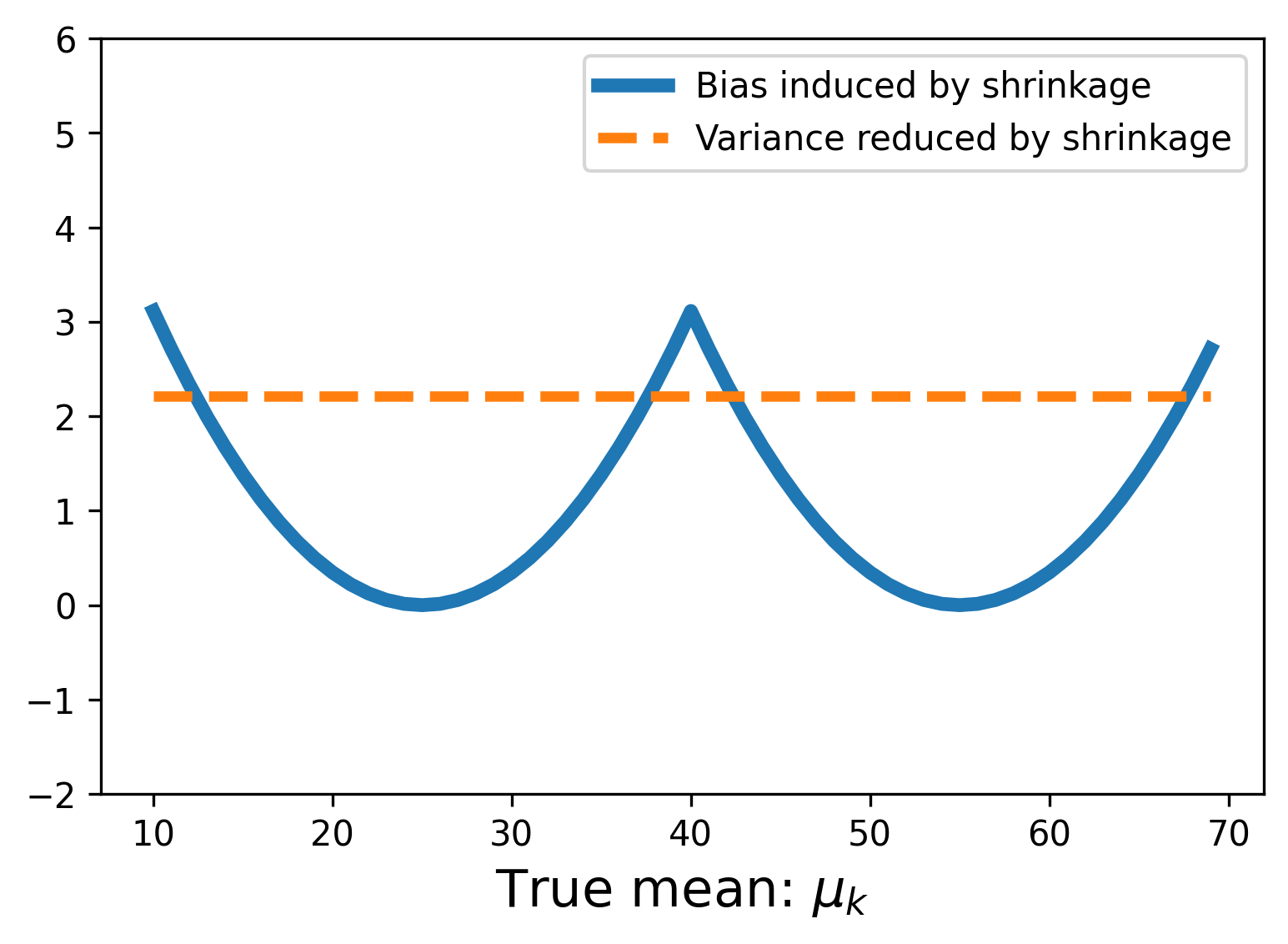}
\end{minipage}
}%
\centering
\caption{Impacts of Shrinkage on Bias and Variance under MSE}
\label{fig_ShrinkageBiasVariance}
 \parbox[t]{0.95\linewidth}{\footnotesize Notes. For the panel (a) ((b) and (c)), the true mean of the random variables associated with each problem is randomly sampled from a uniform distribution, $\mu_k \sim \mathrm{U}(30, 50)$ ($\mu_k \sim \mathrm{U}(10, 70)$). The random variable follows a Gaussian distribution $\xi_k \sim N(\mu_k, 10^2)$. The number of observations for each problem, $N=10$.}
\end{figure}

We illustrate the ``bias-variance trade-off'' regarding the benefit of the data pooling approach in Equation \eqref{cost_decompose} via a numerical example. The true mean of each problem is randomly drawn from an interval, $\mu_k \sim \mathrm{U}(30,50)$, and the corresponding random variable $\xi_k$ follows a Gaussian distribution $N(\mu_k,10^2)$. We assume the number of observations of the random variable for each problem is $N = 10$. As demonstrated in Figure \ref{fig_ShrinkageBiasVariance} (a), the variance reduction is constant for each problem, and the bias term increases as the true mean deviates from the optimal anchor mean, which is 40 in this case. Therefore, the problems with true means centered around the anchor mean benefit the most from shrinkage, whereas, those with true means significantly different from the anchor mean may, in fact, face losses due to shrinkage. Furthermore, when the distribution of true mean $\mu_k$ is expanded from $\mathrm{U}(30,50)$ to $\mathrm{U}(10,70)$, as shown in Figure \ref{fig_ShrinkageBiasVariance} (b), the total benefit of data pooling over SAA is limited when the true means are more dispersed. This is because when true means are widely dispersed, it necessitates a smaller optimal shrinkage parameter to prevent excessive bias in problems on both ends of the true mean distribution. Consequently, this also diminishes the advantages of variance reduction. The intuition suggests that clustering the problems under consideration may help reduce the dispersion of the distribution of true means. Following this intuition, we group the problems into two clusters according to their true means, and then apply the data pooling approaches to each cluster accordingly. As demonstrated in Figure \ref{fig_ShrinkageBiasVariance} (c), by leveraging the cluster structure information, some problems can obtain significantly more benefits, although some are still subject to minor losses due to shrinkage. Therefore, if one can identify appropriate cluster structures, implementing a clustering-then-pooling approach may help reduce the total cost compared to applying data pooling directly to all problems.

\section{Cluster-based Data Pooling under MSE}\label{sec_ClusterMSE}
Inspired by the previous discussion on the bias-variance trade-off in implementing the data pooling approach, especially with dispersed true mean distribution, in this section, we propose the cluster-based data pooling approach aiming to enhance the performance of the data pooling approach. 
We consider two scenarios: (i) the cluster structure is \textit{given}; (ii) the cluster structure is \textit{unknown}. 
For case (i), we leverage the cluster structure and implement the data pooling approach for each cluster of subproblems to reduce the total cost. For case (ii), we devise a data-driven procedure to estimate the cluster structure and investigate the benefit of the clustering procedure for the direct data pooling approach. 
We assume that there exist two clusters of problems, $C_1$ and $C_2$, such that $\mathcal{K} = C_1 \cup C_2$ and $C_1 \cap C_2 = \emptyset$. We consider a sequence of systems, indexed by $K = |C_1| + |C_2|$. We assume that when $K \to \infty$, we have $|C_i| \to \infty$, $i=1,2$, and $\lim_{K\to\infty} \frac{|C_i|}{|C_j|} > 0$ for $i=1,2$, $j = 3-i,$ to avoid trivial cases. 

\subsection{When Cluster Structure is Given}\label{sec_KnownClusterMSE}
Given the cluster structure $(C_1, C_2)$, we implement the data pooling approach for the subproblems within each cluster. The decision function for each cluster based on the data pooling approach is given by,
\begin{equation}\label{eqn_ClusterShrunkenSAAMSE}
x_k(\alpha_i,\mu_{i,0}, \hat{\mu}_k) = \frac{N}{N+\alpha_i} \hat{\mu}_k +\frac{\alpha_i}{N+\alpha_i}\mu_{i,0}, \quad k \in C_i, \quad i = 1,2,
\end{equation}
where $\alpha_i$ denotes the shrinkage parameter and $\mu_{i,0}$ denotes the anchor for the subproblems in the cluster $C_i$, $i = 1,2$. The expected out-of-sample cost for all subproblems in cluster $C_i$ is given by
\begin{equation}\label{eqn_OutOfSampleCost}
    Z(\alpha_i,\mu_{i,0}, C_i) = \sum_{k \in C_i} \mathbb{E}_{S_k}[\mathbb{E}_{\xi_k}[ c(x_k(\alpha_i, \mu_{i,0}, \hat{\mu}_k), \xi_k)]],
\end{equation}
and consequently, the expected out-of-sample cost for all $K$ subproblems is given by $\sum_{i=1}^2 Z(\alpha_i,\mu_{i,0}, C_i)$. If one can evaluate the expected out-of-sample costs of all subproblems within each cluster, the optimal shrinkage parameters can be obtained by solving the following optimization problem.
\begin{equation}\label{eqn_OptimalShrinkageCluster}
(\alpha_{\mu_{1,0}}^\mathrm{AP},\alpha_{\mu_{2,0}}^\mathrm{AP}) \in \argmin_{\alpha_1,\alpha_2} \quad Z(\alpha_1,\mu_{1,0}, C_1) + Z(\alpha_2,\mu_{2,0}, C_2).
\end{equation}
Suppose the means and variances of subproblems from each cluster are known, similar to the shrinkage parameter defined in \eqref{eqn_OptAlphaPooling}, we can compute the optimal shrinkage parameters given the anchor means ${\mu}_{i,0}, i = 1, 2$, as follows,
\begin{equation}\label{eqn_APShirnkageKnwonCluster_FixAnchor}
    \alpha^\mathrm{AP}_{\mu_{i,0}} = \frac{\sum_{k \in C_i} \sigma_k^2}{\sum_{k \in C_i}(\mu_k-\mu_{i,0})^2}, \quad i = 1,2.
\end{equation}
With the optimal shrinkage parameter $\alpha^\mathrm{AP}_{\mu_{i,0}}$, the expected out-of-sample cost for all $K$ problems is computed as,
\begin{equation}\label{eqn_APMSEKnownCluster}
    \sum_{i=1}^2 Z(\alpha_{\mu_{i,0}}^\mathrm{AP}, \mu_{i,0}, C_i) =  \sum_{k=1}^{K}\sigma_k^2+ \sum_{k\in C_1}\frac{\sigma_k^2}{N+\alpha_{\mu_{1,0}}^\mathrm{AP}}+\sum_{k\in C_2}\frac{\sigma_k^2}{N+\alpha_{\mu_{2,0}}^\mathrm{AP}} .
\end{equation}
From the above equation, the total cost decreases as the shrinkage parameter increases. Similar to the direct data pooling approach, the optimal anchor that results in the smallest MSE is the average of means of the subproblems from each cluster, that is, 
\begin{equation}\label{eqn_APAnchorKnwonCluster}
    \mu_{i,0}^\mathrm{AP} = \frac{1}{|C_i|} \sum_{k\in C_i}\mu_k, \quad i = 1,2. 
\end{equation}
If $\mu_{i,0}$ is replaced by $\mu_{i,0}^{\mathrm{AP}}$,  the optimal a priori  shrinkage parameter is computed as, 
\begin{equation}\label{eqn_APShirnkageKnwonCluster}
    \alpha^\mathrm{AP}_{i} = \frac{\sum_{k \in C_i} \sigma_k^2}{\sum_{k \in C_i}(\mu_k-\mu_{i,0}^\mathrm{AP})^2} = \frac{\sum_{k \in C_i} \sigma_k^2}{\sum_{k \in C_i}(\mu_k-\frac{1}{|C_i|} \sum_{k\in C_i}\mu_k)^2}, \quad i = 1,2.
\end{equation}
The following proposition formally states that when the cluster structure is given, the cluster-based data pooling approach with the optimal a priori shrinkage parameters and anchors can outperform the data pooling approach that shrinks all problems with the same shrinkage and anchor parameter. 
\begin{proposition}\label{prop_GivenClusterBenefit} Given the cluster structure $(C_1, C_2)$ and the cost function for each subproblem is mean squared error, that is, $c_k(x_k, \xi_k) = (x_k-\xi_k)^2$, we have, 
\begin{equation}
    Z(\alpha^\mathrm{AP}, \mu_0^\mathrm{AP}, \mathcal{K}) - \sum_{i=1}^2 Z(\alpha_i^\mathrm{AP}, \mu_{i,0}^\mathrm{AP}, C_{i})\ge 0.
\end{equation}
When $\mu_{1,0}^\mathrm{AP} \neq \mu_{2,0}^\mathrm{AP}$, the above inequality holds strictly.
\end{proposition}

With the anchor mean $\mu_0^\mathrm{AP}$ and the shrinkage parameter $\alpha^\mathrm{AP}$, as noted in \cite{GuptaKallus2022}, the cost $Z(\alpha^\mathrm{AP}, \mu_0^\mathrm{AP}, \mathcal{K})$ provides the lowest cost that the data pooling approach can possibly achieve. Similarly, the expected total out-of-sample cost $\sum_{i=1}^2 Z(\alpha_{i}^\mathrm{AP}, \mu_{i,0}^\mathrm{AP}, C_{i})$  provides the lowest cost that the cluster-based data pooling approach can possibly achieve, which not only leverages the distribution information but also the cluster structure. Proposition \ref{prop_GivenClusterBenefit} indicates the potential improvement on the direct data pooling approach by leveraging the given cluster structure information. However, the cost $\sum_{i=1}^2 Z(\alpha_{i}^\mathrm{AP}, \mu_{i,0}^\mathrm{AP}, C_{i})$  cannot be achieved in practice as the distribution information is not known. In the following, we propose a data-driven procedure to estimate $\alpha_i^\mathrm{AP}$ and $\mu_{i,0}^\mathrm{AP}$. When the number of problems $K$ grows, the cluster-based data pooling approach with the estimated shrinkage parameters and anchor means can result in the expected out-of-sample cost that is close to that of the optimal a priori parameters under mild conditions, and thus, can outperform the direct data pooling approach that ignores the cluster structure.  

\begin{assumption}\label{asmp_1}
 The random variable $\xi_k, k \in \mathcal{K}$, satisfy one of following two conditions:
\begin{itemize}
    \item[(i)] there exist constants $a_\mathrm{max} < \infty$ and $\sigma_\mathrm{max}<\infty$ such that $|\xi_k| \le a_\mathrm{max}$ and $\sigma^2_k \leq \sigma_\mathrm{max}^2$;
    \item[(ii)] $\xi_k$ follows the Gaussian distribution with mean $\mu_k$ and variance $\sigma_k^2$. These exist constants $b_\mathrm{max} < \infty$ and $\sigma_\mathrm{max}<\infty$ such that $|\mu_k| \le b_\mathrm{max}$ and $\sigma_k^2 \le \sigma_\mathrm{max}^2$.
\end{itemize}
\end{assumption}
In the next proposition, we characterize the limits of the data-driven shrinkage parameter and the corresponding expected out-of-sample cost for a cluster. Specifically, as $K \to \infty$, the data-driven anchor and shrinkage parameters are close to the optimal a priori parameters, and thus, the expected out-of-sample cost with the data-driven parameters is close to that of the optimal a priori parameters. For ease of exposition, we drop the subscript $i$ and present the result for an arbitrary cluster $C$. Let $|C|$ denote the cardinality of the set $C$.

\begin{proposition} \label{prop_DataDrivenPooling}
 Suppose Assumption \ref{asmp_1} holds, and the cost function for each subproblem is the mean squared error, that is, $c_k(x_k, \xi_k) = (x_k-\xi_k)^2$. For a set of problems denoted by $C$, let
\begin{align*}
    &\hat{\mu}_0 = \frac{1}{|C|}\sum_{k\in C} \hat{\mu}_k,\\
    &\hat{\alpha} = \frac{\sum_{k \in C}\frac{1}{N-1}\sum_{j = 1}^{N}(\hat{\xi}_{kj}-\hat{\mu}_k)^2}{\sum_{k\in C}(\hat{\mu}_0 -\hat{\mu}_k)^2-\frac{1}{N}\sum_{k \in C}\frac{1}{N - 1}\sum_{j = 1}^{N}(\hat{\xi}_{kj}-\hat{\mu}_k)^2}.
\end{align*}
Then, as $|C| \to \infty$, we have,
\begin{itemize}
    \item[(i)]  $\hat{\mu}_0 \to_p \mu_0^{\mathrm{AP}}$, $\hat{\alpha} \to_p \alpha^\mathrm{AP}$, where   $\mu_0^\mathrm{AP} = \frac{1}{|C|}\sum_{k \in C}\mu_k$ and  $\alpha^\mathrm{AP} = \frac{\sum_{k\in C}\sigma_k^2}{\sum_{k\in C}(\mu_k - \mu_0^{\mathrm{AP}})^2}$;
    \item[(ii)] $ \frac{1}{|C|}\sum_{k \in C}  (x_k(\hat{\alpha},\hat{\mu}_0,\hat{\mu}_k)-\mu_k)^2  \to_p \frac{1}{|C|}\sum_{k \in C}  (x_k(\alpha^\mathrm{AP},\mu_0^{\mathrm{AP}},\hat{\mu}_k)-\mu_k)^2$.
\end{itemize}
\end{proposition}

The results and the proof in Proposition \ref{prop_DataDrivenPooling} are related to Theorem 2.1 in \cite{GuptaKallus2022}. 
Our results are different from that of  \cite{GuptaKallus2022} in the following two aspects. First, as \cite{GuptaKallus2022} essentially considers the setting where Part (i) of Assumption \ref{asmp_1} satisfies, i.e., the random variables and the second moments are bounded, our results can also be applied to Part (ii) of Assumption \ref{asmp_1} where the random variables are unbounded, e.g., Gaussian random variables. Second, while we consider the data-driven anchor parameter $\hat{\mu}_0$, \cite{GuptaKallus2022} analyzes the asymptotic performance of the data pooling algorithm with a fixed anchor parameter. The dependence across the terms $(\hat{\mu}_0 - \hat{\mu}_k)^2$ in the equation that computes the shrinkage parameter $\hat{\alpha}$ makes the analysis more involved (see the proof of Proposition \ref{prop_DataDrivenPooling} in Appendix \ref{appx_A}). By decomposing the difference in expected out-of-sample costs and deriving the convergence results for each component with concentration inequalities for bounded random variables and Sub-Gaussian (Sub-Exponential) random variables, we are able to show that the expected out-of-sample cost of the data pooling algorithm is close to that of the optimal a priori parameters when the size of the cluster goes to infinity even with unbounded random variables and data-driven anchors.

\begin{theorem}\label{thm_DataDrivenPooling} When Assumption \ref{asmp_1} holds and the cost function for each subproblem is the mean squared error, that is, $c_k(x_k, \xi_k) = (x_k-\xi_k)^2$. For the cluster $C_i$, $i=1,2$, let
\begin{align*}
    \hat{\mu}_{i,0} &= \frac{1}{|C_i|}\sum_{k\in C_i} \hat{\mu}_k,\\
    \hat{\alpha}_i&= \frac{\sum_{k \in C_i}\frac{1}{N-1}\sum_{j = 1}^{N}(\hat{\xi}_{k,j}-\hat{\mu}_k)^2}{\sum_{k\in C_i}(\hat{\mu}_{i,0} -\hat{\mu}_k)^2-\frac{1}{N}\sum_{k \in C_i}\frac{1}{N - 1}\sum_{j = 1}^{N}(\hat{\xi}_{k,j}-\hat{\mu}_k)^2}.
\end{align*}
We have,
\begin{equation}
\frac{1}{K}\sum_{i=1}^2 \sum_{k \in C_i}  (x_k(\hat{\alpha}_i,\hat{\mu}_{i,0},\hat{\mu}_k)-\mu_k)^2  \to_p \frac{1}{K} \sum_{i=1}^2 \sum_{k \in C_i}  (x_k(\alpha^\mathrm{AP}_{i},\mu_{i,0}^\mathrm{AP},\hat{\mu}_k)-\mu_k)^2. 
\end{equation}

 \end{theorem}

Theorem \ref{thm_DataDrivenPooling} suggests that if we implement the data pooling approach to each cluster with the data-driven parameters, $\hat{\mu}_{i,0}$ and $\hat{\alpha}_{i}$, the expected out-of-sample cost is close to that of the one with optimal shrinkage parameters and anchors as the size of clusters grows. The latter is no larger than the expected out-of-sample cost of the direct data pooling approach, even using the optimal shrinkage parameter and anchor (recall Proposition \ref{prop_GivenClusterBenefit}). This underscores the advantage of the cluster-based data pooling approach over the direct data pooling approach when the cluster structure is given.

\subsection{When Cluster Structure is Unknown}\label{sec_mse_unknown}
\begin{figure}[htbp!]
    \centering
     \includegraphics[width=0.6\linewidth]{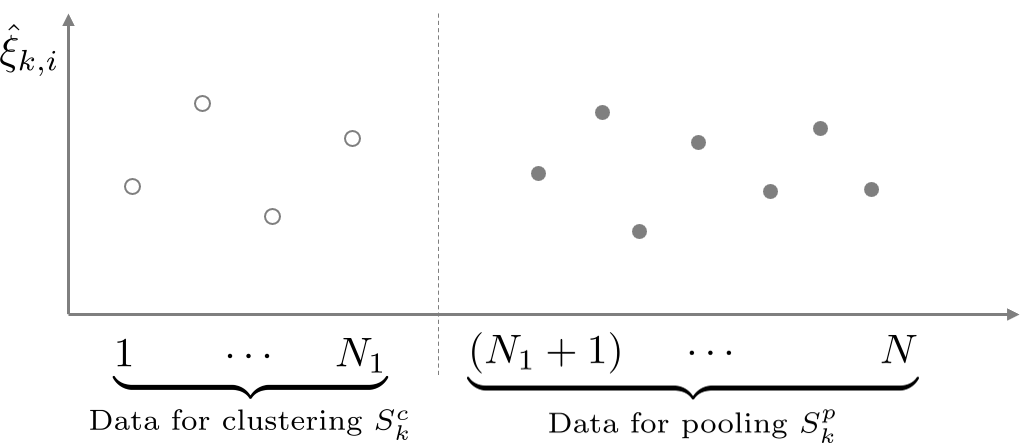}
    \caption{Data allocation for clustering and pooling for problem $k$}
    \label{fig:data_sepa}
\end{figure}

In real-world operations management settings, the underlying cluster structure may be unclear to the decision maker. To estimate the cluster structure, as demonstrated in Figure \ref{fig:data_sepa}, one may allocate $N_1$ data points of the data set $S_k$ for each $k\in \mathcal{K}$, denoted by $S_k^c =\{\hat{\xi}_{kj}: j = 1,2\ldots, N_1\}$, to identify the cluster structure with some clustering algorithm,
and then, utilize the rest of $(N-N_1)$ data points, denoted by $S_k^p = \{\hat{\xi}_{kj}: j = N_1+1, N_1+2, \ldots, N\}$, to implement the data pooling approach for subproblems within each cluster, respectively. 
In this section, we implement a simple clustering algorithm that splits all products  $\mathcal{K}$ into two clusters with an estimated cluster boundary. Specifically, for any $k \in \mathcal{K}$, we first compute the sample mean of each data set $S_k^c$, denoted by $\hat{\mu}_k^c$. We then obtain the cluster boundary by computing $\frac{1}{K}\sum_{k=1}^{K}\hat{\mu}_k^c$. Finally, we group the problem with sample mean $\hat{\mu}_k^c$ that is smaller (larger) than the cluster boundary into cluster $\hat{C}_1~(\hat{C}_2)$. Consequently, the estimated cluster structure is denoted by $(\hat{C}_1, \hat{C}_2)$. 

Given the estimated cluster structure $(\hat{C}_1, \hat{C}_2)$, one can implement the data pooling approach with the data set $S^p = \{S_1^p, S_2^p,\ldots, S_K^p\}$. With slight abuse of notation, let $\hat{\mu}_k^p = \frac{1}{N-N_1} \sum_{i=1}^{N-N_1} \hat{\xi}_{k,N_1+i}$, $k \in \mathcal{K}$, denote the sample mean obtained with the data for pooling $S^p_k$. Therefore, the decision for problem $k$ in the cluster $\hat{C}_i$ is,
\begin{equation}
x_k(\alpha_i, \mu_{i,0}, \hat{\mu}_k^p) = \frac{N-N_1}{N-N_1+\alpha_i} \hat{\mu}_k^p +\frac{\alpha_i}{N-N_1+\alpha_i}\mu_{i,0}, \quad k \in \hat{C}_i, \quad i = 1, 2,
\end{equation}
where $\mu_{i,0}$ and $\alpha_i$ denote the anchor mean and shrinkage parameter for the cluster $\hat{C}_i$, respectively. Consequently, the expected out-of-sample cost for all problems in the cluster $\hat{C}_i$ is given by
\begin{equation}\label{eqn_DataDrivenShrinkageCluster}
    Z^{p}(\alpha_i,\mu_{i,0}, \hat{C}_i) = \sum_{k \in \hat{C}_i} \mathbb{E}_{S_k^p}[\mathbb{E}_{\xi_k}[ c(x_k(\alpha_i, \mu_{i,0}, \hat{\mu}_k^p), \xi_k)]],
\end{equation}
and, the expected out-of-sample cost for all $K$ subproblems is given by $\sum_{i=1}^2 Z^{p}(\alpha_i, \mu_{i,0}, \hat{C}_i)$. Note that the cost function $Z^{p}$ is similar to the cost function $Z$ defined in \eqref{eqn_OutOfSampleCost}, except that outer expectation is taken over the data set $S_k^p$ rather than $S_k$, as a portion of available data is allocated for clustering.
Similar to the analysis in Section \ref{sec_KnownClusterMSE}, we define the shrinkage parameter and the anchor mean that can yield 
 the lowest possible cost that one can achieve with the estimated cluster structure. Specifically, suppose the true distributions of random variables are known, one can evaluate the out-of-sample cost, and identify the optimal parameters that minimize the total out-of-sample cost. Thus, given the estimated cluster structure $(\hat{C}_1, \hat{C}_2)$, the optimal a priori anchor means and shrinkage parameters are given by,
\begin{align}
\mu_{i, 0}^\mathrm{AP}(\hat{C}_i) & = \frac{1}{|\hat{C}_i|}\sum_{k \in \hat{C}_i}\mu_k,\quad i = 1,2, \label{eqn_APAnchorUnknwonCluster}\\
\alpha_i^\mathrm{AP}(\hat{C}_i) &= \frac{\sum_{k\in \hat{C}_i}\sigma_k^2}{\sum_{k\in \hat{C}_i}(\mu_k - \mu_{i, 0}^\mathrm{AP}(\hat{C}_i))^2},\quad i = 1,2. \label{eqn_APShirnkageUnknwonCluster}
\end{align}
Different from the anchor and the shrinkage parameter defined in Equations \eqref{eqn_APAnchorKnwonCluster} and \eqref{eqn_APShirnkageKnwonCluster} with the given cluster structure, we emphasize the dependence of the anchor mean and the shrinkage parameter on the estimated cluster structure in \eqref{eqn_APAnchorUnknwonCluster} - \eqref{eqn_APShirnkageUnknwonCluster}. Note that, by definition, we have $\alpha_i^\mathrm{AP}(C_i) = \alpha_i^\mathrm{AP}$ and $\mu_{i,0}^\mathrm{AP}(C_i)= \mu_{i,0}^\mathrm{AP}$.

To measure the performance of the data pooling approaches in the small-data large-scale regime, we define the {\em average expected out-of-sample cost} for direct data pooling, denoted by $\overline{Z}(\alpha,\mu_0)$, and that of cluster-based data pooling, denoted by $\overline{Z}^c(\bm{\alpha},\bm{\mu},\bm{C})$, under MSE. 
\begin{align*}
    \overline{Z}(\alpha,\mu_0) & = \lim_{K \to \infty}\frac{1}{K}\sum_{k = 1}^{K} (x_k(\alpha,\mu_{0},\hat{\mu}_k) - \mu_k)^2, \\
\overline{Z}^c(\bm{\alpha},\bm{\mu},\bm{C}) &= \lim_{K \to \infty}\frac{1}{K}\sum_{i \in \{1,2\}} \sum_{k \in C_i}(x_k(\alpha_i,\mu_{i,0},\hat{\mu}^p_k) - \mu_k)^2.
\end{align*}
The average out-of-sample cost $\overline{Z}(\alpha,\mu_0)$ is incurred with the choice of shrinkage parameter $\alpha$ and anchor mean $\mu_0$, which can be specified a priori or obtained via a data-driven procedure. In addition to the shrinkage parameters ${\bm \alpha}=(\alpha_1, \alpha_2)$ and the anchor means ${\bm \mu} = (\mu_1, \mu_2)$, the average out-of-sample cost $\overline{Z}^c(\bm{\alpha},\bm{\mu},\bm{C})$ is also contingent on the cluster structure $\bm{C} = (C_1, C_2)$, which is either given or estimated by a clustering algorithm.


To investigate the benefit of the cluster-based data pooling approach when the cluster structure is unknown,  we assume the random variables $\xi_k$, $k\in \mathcal{K},$ follow Gaussian distributions and their means are uniformly distributed over two disjoint intervals. 
\begin{assumption} \label{asmp_2}
 The random variable $\xi_k,\mbox{ }k \in \mathcal{K}$, satisfies the following,
\begin{itemize}
    \item[(i)] the true mean of $\xi_k$, denoted as $\mu_k$, is uniformly distributed over the intervals $[a,\frac{a+b}{2}] \cup [\frac{a+b}{2}+d(b-a), b+d(b-a)]$, where $a, b, d$ are constants such that $b>a\geq0$ and $d>0$;
    \item[(ii)] the random variable $\xi_k \sim N(\mu_k,\sigma_k^2)$ where $\sigma_k < \infty$.
\end{itemize}
\end{assumption}



Assumption \ref{asmp_2} (i) assumes that the true means are randomly sampled from two disjoint intervals, and the separateness of two clusters is indicated by the normalized distance $d$. Assumption \ref{asmp_2} (ii) assumes the random variables follow Gaussian distributions. Without loss of generality, we denote the set of problems by $C_1$ ($C_2$) if the mean of the corresponding random variable is drawn from the interval $[a,\frac{a+b}{2}]$ ($[\frac{a+b}{2}+d(b-a), b+d(b-a)]$). 
When the underlying cluster structure, $(C_1, C_2)$, is given, suggested by the analysis in Section \ref{sec_KnownClusterMSE}, implementing the cluster-based data pooling approach with optimal anchor and shrinkage parameters defined in   \eqref{eqn_APAnchorKnwonCluster} and \eqref{eqn_APShirnkageKnwonCluster} 
 provides a lower bound of the out-of-sample cost, which is, $\sum_{i=1}^2 Z^{p}(\alpha_i^\mathrm{AP}, \mu_{i,0}^\mathrm{AP}, C_i)$.

\begin{proposition} \label{prop_BenefitUnknownClusterN1}
Let $\Delta = \lim_{K\to\infty} \frac{1}{K}Z(\alpha^\mathrm{AP}, \mu_0^\mathrm{AP}, \mathcal{K}) - \frac{1}{K}\sum_{i=1}^2 Z^{p}(\alpha_{i}^\mathrm{AP}, \mu_{i,0}^\mathrm{AP}, C_i)$. When Assumption \ref{asmp_2} holds, if $\frac{\overline{\sigma}}{\sqrt{N_1}} \ge \frac{b-a}{6}$ and $d\ge \sqrt{\frac{1}{12- N_1(b-a)^2/(4\overline{\sigma}^2)}-\frac{1}{12}}-\frac{1}{2}$, where $\overline{\sigma} = \sqrt{\frac{1}{K}\sum_{k=1}^{K}\sigma_k^2}$, we have $\Delta > 0$ and $\Delta$ increases as $d$ increases.
\end{proposition}

Proposition \ref{prop_BenefitUnknownClusterN1} demonstrates the potential benefit of the cluster-based data pooling approach that sacrifices $N_1$ data points to identify the cluster structure, and utilizes the rest of $(N-N_1)$ data points to implement the data pooling approach to respective clusters. Specifically, when $d$ is large enough, and the sample variance is relatively large, the cluster-based data pooling approach can result in smaller expected out-of-sample costs, even at the expense of utilizing a few data points for clustering analysis. However, such expected out-of-sample cost $\sum_{i=1}^2 Z^{p}(\alpha_i^\mathrm{AP}, \mu_{i,0}^\mathrm{AP}, C_i)$ is achieved by exploiting the true cluster structure information and distribution information of each subproblem, which are unavailable in practice. In the following, we characterize the condition under which the cluster-based data pooling approach can outperform the data pooling approach when the cluster structure and distribution information are unknown.

\begin{theorem}\label{thm_DataDrivenClusterBenefitThreshold}
 Given $N$ data points for each subproblem, $N_1 < N-1$ data points are used to estimate the cluster structure $(\hat{C}_1, \hat{C}_2)$.  For $i=1, 2$, let
\begin{align*}
    \hat{\mu}_{i,0}(\hat{C}_i) = \frac{1}{|\hat{C}_i|}\sum_{k\in \hat{C}_i} \hat{\mu}_k^p,\quad
    \hat{\alpha}_i(\hat{C}_i) = \frac{\sum_{k \in \hat{C}_i}\frac{1}{N-N_1-1}\sum_{j = N_1+1}^{N}(\hat{\xi}_{k,j}-\hat{\mu}_k^p)^2}{\sum_{k\in \hat{C}_i}(\hat{\mu}_{i,0}-\hat{\mu}_k^p)^2-\frac{1}{N-N_1}\sum_{k \in \hat{C}_i}\frac{1}{N-N_1 - 1}\sum_{j = N_1+1}^{N}(\hat{\xi}_{k,j}-\hat{\mu}_k^p)^2}.    
\end{align*} 
When Assumption \ref{asmp_2} holds, $\frac{\overline{\sigma}}{\sqrt{N_1}} \ge \frac{b-a}{6}$, and $d\ge \sqrt{\frac{1}{12- N_1(b-a)^2/(4\overline{\sigma}^2)}-\frac{1}{12}}-\frac{1}{2}$, where $\overline{\sigma} = \sqrt{\frac{1}{K}\sum_{k=1}^{K}\sigma_k^2}$, and the cost function of the $k$-th subproblem is the mean squared error, that is, $c_k = (x_k-\xi_k)^2, k\in \mathcal{K}$,
  there exists a threshold $d^*$ such that when $d \ge d^*$, we have,
 \begin{equation}
\mathbb{E}_{S}\Big[\overline{Z}(\hat{\alpha},\hat{\mu}_0) - \overline{Z}^c(\hat{\bm{\alpha}}(\hat{\bm{C}}),\hat{\bm{\mu}}(\hat{\bm{C}}),\hat{\bm{C}})\Big] > 0,
 \end{equation}
 where $\hat{\bm{\alpha}}(\hat{\bm{C}}) = (\hat{\alpha}_1(\hat{C}_1),\hat{\alpha}_2(\hat{C}_2))$ and $\hat{\bm{\mu}}(\hat{\bm{C}}) = (\hat{\mu}_{1,0}(\hat{C}_1),\hat{\mu}_{2,0}(\hat{C}_2))$.
\end{theorem}

Theorem \ref{thm_DataDrivenClusterBenefitThreshold} theoretically demonstrates the benefit of the proposed cluster-based data pooling approach in a data-driven setting when the distance between the underlying clusters is sufficiently large. To unfold the intuition behind Theorem \ref{thm_DataDrivenClusterBenefitThreshold}, we decompose the difference between the expected out-of-sample cost of the cluster-based data pooling approach with the estimated clusters and that of the direct data pooling approach into the following four components,
\begin{align*}
     & \mathbb{E}_{S}[\overline{Z}(\hat{\alpha},\hat{\mu}_0)] - \mathbb{E}_{S}\bigg[\overline{Z}^c(\hat{\bm{\alpha}}(\hat{\bm{C}}),\hat{\bm{\mu}}(\hat{\bm{C}}),\hat{\bm{C}})\bigg] \\
      = & \underbrace{\vphantom{\mathbb{E}_{S}[\overline{Z}(\alpha^{\mathrm{AP}},\mu_0^{\mathrm{AP}})] -  E_{S}\bigg[\overline{Z}^c(\bm{\alpha^{\mathrm{AP}}}(\bm{C}),\bm{\mu^{\mathrm{AP}}}(\bm{C}),\bm{C})\bigg]} \mathbb{E}_{S}[\overline{Z}(\hat{\alpha},\hat{\mu}_0)] - \mathbb{E}_{S}[\overline{Z}(\alpha^{\mathrm{AP}},\mu_0^{\mathrm{AP}})]}_{\text{I: Loss of data pooling with $\hat{\mu}$ and $\hat{\alpha}$ (Proposition \ref{prop_DataDrivenPooling})}}  + \underbrace{\mathbb{E}_{S}[\overline{Z}(\alpha^{\mathrm{AP}},\mu_0^{\mathrm{AP}})] -  \mathbb{E}_{S}\bigg[\overline{Z}^c(\bm{\alpha}^{\mathrm{AP}}(\bm{C}),\bm{\mu}^{\mathrm{AP}}(\bm{C}),\bm{C})\bigg]}_{\text{II: Benefit of exploiting given cluster structure (Proposition \ref{prop_BenefitUnknownClusterN1})}}  \\
     &+ \underbrace{\mathbb{E}_{S}\bigg[\overline{Z}^c(\bm{\alpha}^{\mathrm{AP}}(\bm{C}),\bm{\mu}^{\mathrm{AP}}(\bm{C}),\bm{C})\bigg]  - \mathbb{E}_{S}\bigg[\overline{Z}^c(\bm{\alpha}^{\mathrm{AP}}(\hat{\bm{C}}),\bm{\mu}^{\mathrm{AP}}(\hat{\bm{C}}),\hat{\bm{C}})\bigg]}_{\text{III: Loss due to inaccurate estimation of cluster structure (Lemma \ref{lma_CostDifferenceBound})}} \\
     &+ \underbrace{\mathbb{E}_{S}\bigg[\overline{Z}^c(\bm{\alpha}^{\mathrm{AP}}(\hat{\bm{C}}),\bm{\mu}^{\mathrm{AP}}(\hat{\bm{C}}),\hat{\bm{C}})\bigg] - \mathbb{E}_{S}\bigg[\overline{Z}^c(\hat{\bm{\alpha}}(\hat{\bm{ C}}),\hat{\bm{\mu}}(\hat{\bm{C}}),\hat{\bm{C}})\bigg]}_{\text{IV: Loss of cluster-based data pooling with $\hat{\mu}_{i,0}$ and $\hat{\alpha}_{i}$ (Lemma \ref{lemma_1})}}.
\end{align*}

As $K \to \infty$, the data-driven anchor and shrinkage parameters are close to the optimal a priori ones, and thus, the gap between the associated expected out-of-sample cost and that of the oracle (i.e., Parts I and IV) converges to zero (see Proposition \ref{prop_DataDrivenPooling} and Lemma \ref{lemma_1}). Therefore, whether the cluster-based data pooling approach can generate additional benefits over direct data pooling depends on the asymptotic behaviors of the benefit of exploiting the clustering structure (i.e., Part II) and the loss due to inaccurate clustering outcomes (i.e., Part III). For Part II, in Proposition \ref{prop_BenefitUnknownClusterN1}, we have shown that the benefit of the cluster-based data pooling approach is more significant as the distance between two clusters, $d$, increases when the cluster structure is given. In addition, for Part III, the optimality gap due to the inaccurate estimation of the cluster structure decreases exponentially in the distance $d$ (Lemma \ref{lma_CostDifferenceBound}).  Thus, when $d$ is sufficiently large, the benefit of leveraging the cluster structure dominates the loss caused by the inaccurate estimation of cluster structure, and thus, the cluster-based data pooling approach can outperform the direct data pooling approach, even when the cluster structure is unknown and needs to be estimated.

\begin{figure}[htbp!]
\centering
\subfigure[The impact of the number of data points for clustering]{
\begin{minipage}[t]{0.5\linewidth}
\centering
\includegraphics[height=2.5in]{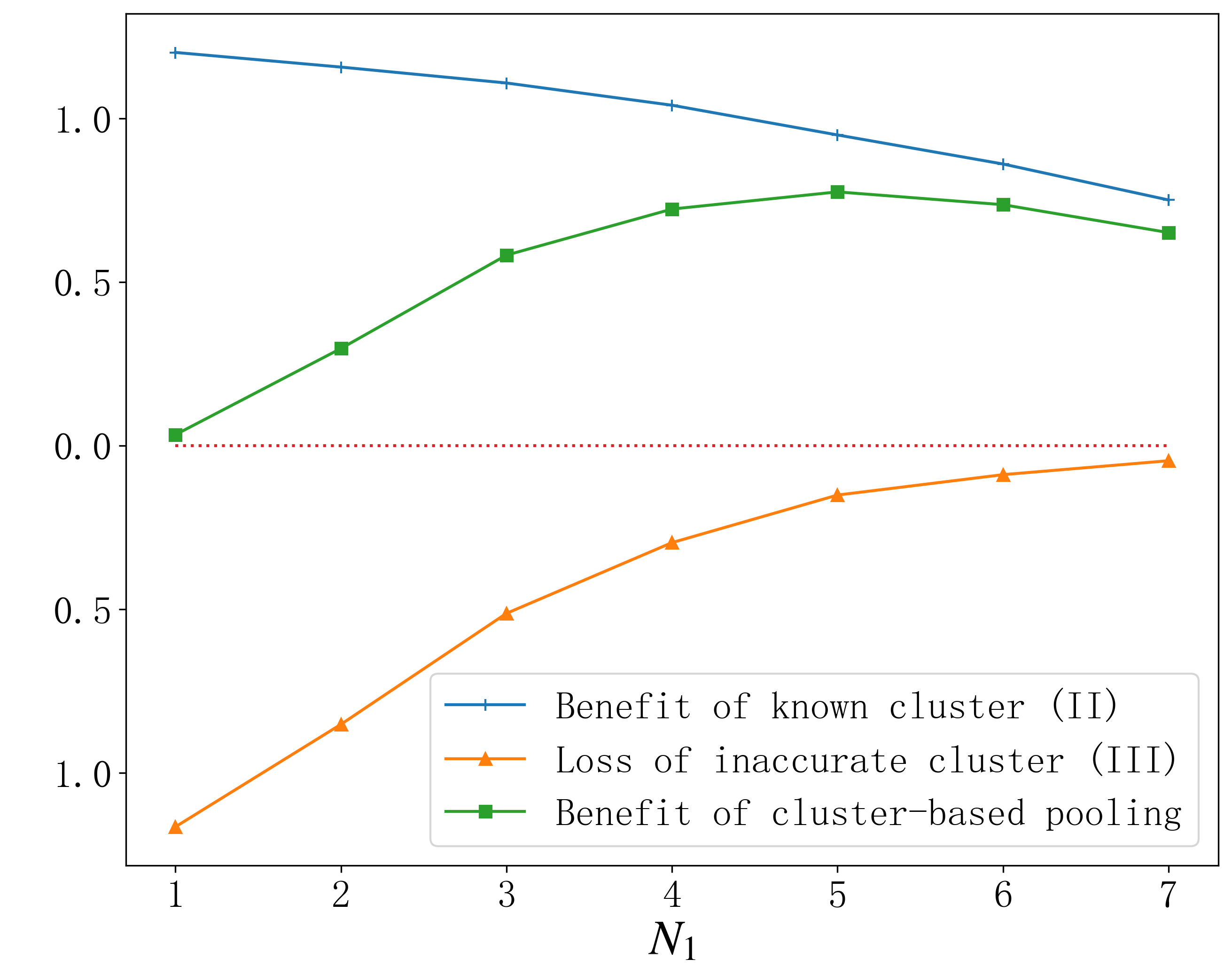}
\end{minipage}%
}%
\subfigure[The impact of distance between clusters]{
\begin{minipage}[t]{0.5\linewidth}
\includegraphics[height=2.5in]{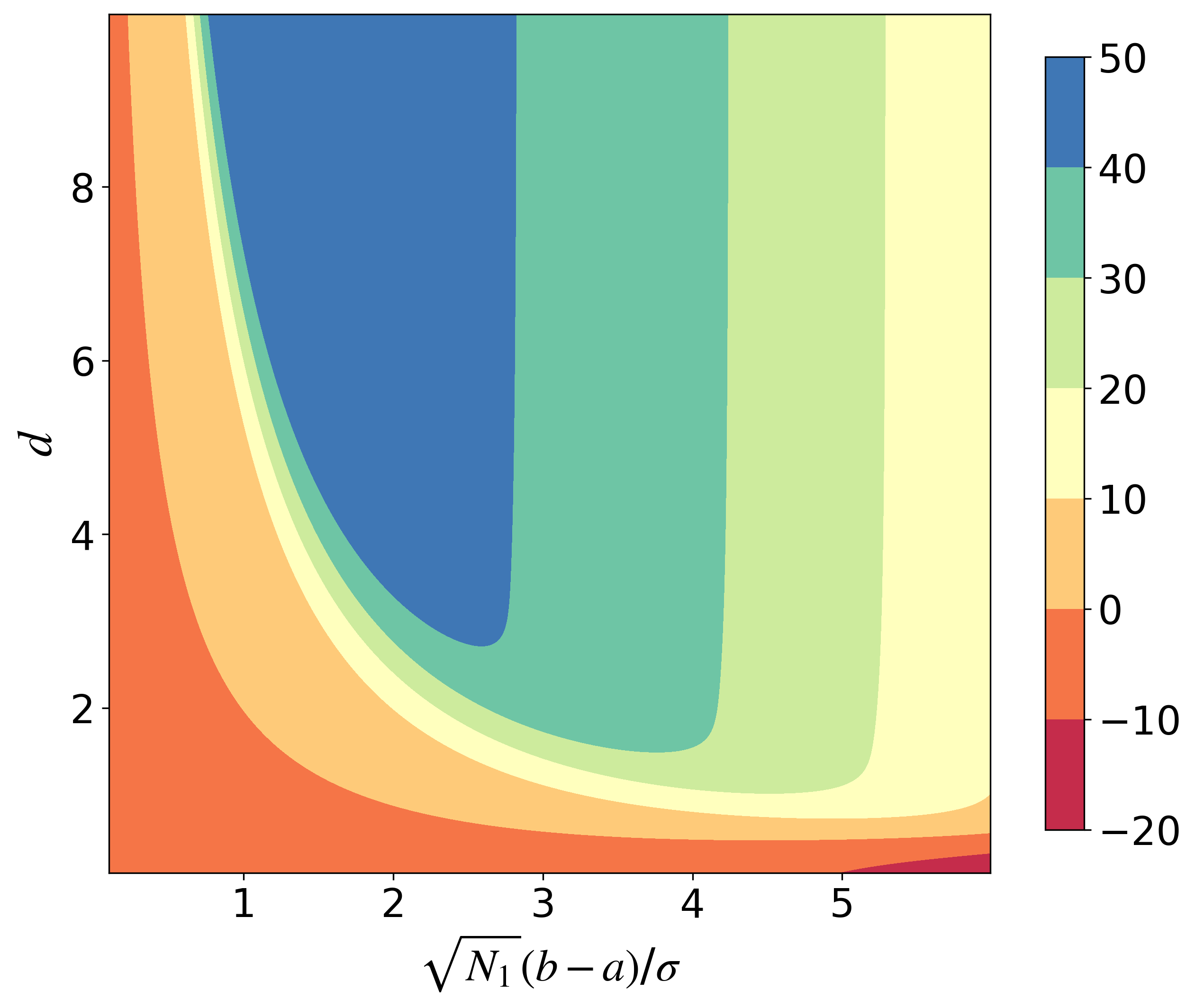}
\end{minipage}%
}%

\centering
\caption{Benefit of cluster-based data pooling}
\label{fig:MSE_whole_picture}
\parbox[t]{0.95\linewidth}{\footnotesize Notes. In panel (a), we demonstrate the potential benefit of cluster-based data pooling over direct data pooling with different numbers of data points for clustering. The true mean $\mu_k$ is randomly drawn from the two uniform distributions $ \mathrm{U}(10, 15)$ and $\mathrm{U}(25, 30)$, and the random variable of problem $k$, $\xi_k \sim \mathrm{N}(\mu_k,5^2)$. The number of observations for each problem is $N=10$. The number of data points for clustering $N_1$ ranges from 1 to 7.  The reported values are the average of 1000 simulated instances.
For panel (b), we provide a full characterization of the theoretical benefit of cluster-based pooling over direct pooling as $K \to \infty$. The true mean of the problem $\mu_k$ is randomly drawn from the two uniform distributions $ \mathrm{U}(a, \frac{a+b}{2})$ and $\mathrm{U}(\frac{a+b}{2} + d(b-a), b+d(b-a))$, and the random variable for problem $k$, $\xi_k \sim \mathrm{N}(\mu_k, \sigma^2)$. The distance between the two clusters $d$ ranges from 0 to 10. 
}
\end{figure}

 As demonstrated in Figure \ref{fig:MSE_whole_picture} (a), how to allocate the data points for clustering and pooling takes a critical role in determining the interplay between the four components in the above equation. In particular, as $N_1$ increases, the benefit of exploiting the given cluster structure, i.e., Part II, decreases. It is because an increase in $N_1$ implies fewer data points reserved for data pooling, and thus, less accurate estimates of shrinkage parameters and the true mean with the data for pooling (i.e., higher variance of $\hat{\mu}_k^p$; see Equation \eqref{prop_3.3_final_eq} in the proof of Proposition \ref{prop_BenefitUnknownClusterN1}). On the other hand, as $N_1$ increases, one can obtain more accurate estimates of true means with the data for clustering, and thus, alleviate the loss due to inaccurate cluster structure estimation, indicated by Part III. In addition, the number of data points for clustering, $N_1$, doesn't influence the value of the first and fourth components when the number of problems grows. Therefore, the benefit of the cluster-based data pooling approach over direct data pooling first increases in $N_1$ and then decreases in $N_1$, demonstrating a non-monotone relationship with respect to the number of data points allocated for clustering.

Our discussion in Section \ref{sec_DataPooling_MSE} suggests that applying clustering analysis to all problems can help reduce the dispersion of the true means, and thus, improve the performance of the data pooling approach. Surprisingly, we observe that the cluster-based data pooling approach does not always outperform the direct data pooling approach that shrinks the solutions of all problems using the same shrinkage parameter. The following theorem formally characterizes the conditions in which the cluster-based data pooling approach cannot generate additional benefits compared with the direct data pooling approach in the asymptotic regime. 
\begin{theorem}\label{thm_DataDrivenClusterWorse} Let $\tilde{y} = \frac{\sqrt{N_1}(b-a)}{\sigma}$ and 
define the function
 \begin{align*}
    &L(y) \\
    &= \Bigg[\frac{y^2(4+3 \mathrm{Erf}\left[\frac{y}{2 \sqrt{2}}\right]^2)}{48}-\frac{2
e^{-\frac{y^2}{4}}}{\pi }-\frac{e^{-\frac{y^2}{8}} \sqrt{\frac{2}{\pi }} \mathrm{Erf}\left[\frac{y}{2 \sqrt{2} }\right] \sqrt{N_1} \left(-4 +y^2\right)}{y}+\frac{2  \mathrm{Erf}\left[\frac{-y}{2
\sqrt{2}}\right]^2 \left(-2 +y^2 N_1\right)}{y^2}
    \Bigg]^{-1} \\
&-\frac{12}{y^2} - 1.
\end{align*}
When Assumption \ref{asmp_2} holds, $\sigma_k=\sigma, k\in\mathcal{K}$, and $d=0$, if and only if (i) $\tilde{y} > 6$ or (ii) $0 \leq \tilde{y} \leq  6$ and $L(\tilde{y}) < 0$, we have,
\begin{equation}
\mathbb{E}_{S}[\overline{Z}(\alpha^\mathrm{AP},\mu_0^\mathrm{AP})] - \mathbb{E}_{S}\bigg[\overline{Z}^c(\hat{\bm{\alpha}}^\mathrm{AP}(\hat{\bm{C}}),\hat{\bm{\mu}}^\mathrm{AP}(\hat{\bm{C}}),\hat{\bm{C}})\bigg] < 0
\end{equation}
\end{theorem}

Theorem \ref{thm_DataDrivenClusterWorse} shows that when Assumption \ref{asmp_2} holds and $d = 0$,\footnote{We conduct an extensive numerical study and find that when $\tilde{y} \in [0,6]$, we have, $L(\tilde{y}) < 0$.} the cluster-based data pooling approach does not generate benefit over the direct data pooling approach as the former sacrifices $N_1$ data points for clustering rather than pooling. In other words, when the underlying distributions are relatively concentrated, the cluster-based data pooling approach that requires a certain amount of data points to estimate the cluster structure may not outperform the direct data pooling approach. In addition, we provide a complete characterization of the benefit of the cluster-based data pooling approach over the direct data pooling approach under MSE in Figure \ref{fig:MSE_whole_picture} (b). Given the ratio $ \sqrt{N_1(b-a)} / \sigma$, the advantage of cluster-based data pooling is more evident as the distance between two clusters of problems $d$ increases. When $d$ is small, however, the cluster-based data pooling approach may not generate additional benefits over the direct data pooling approach.

{\bf Remark 1.} One may further enhance the performance of cluster-based data pooling by conducting clustering analysis recursively on each estimated cluster. As suggested by Proposition \ref{prop_GivenClusterBenefit}, implementing cluster-based data pooling with the optimal a priori shrinkage parameters and anchor distribution can always improve over direct data pooling. In addition, as indicated by Theorem \ref{thm_DataDrivenPooling}, when the number of problems grows, the average out-of-sample cost obtained with data-driven parameters is close to that of the optimal a priori ones when implementing cluster-based data pooling. This suggests that if the number of problems within each cluster is sufficient, then further segmenting the clusters, which does not necessitate sacrificing new data points, may result in improved performance of cluster-based data pooling. Thus, when identifying the cluster structure, one may conduct a bisect clustering algorithm \citep{SteinbachKarypisKumar2000, MurugesanZhang2011}, a case of divisive clustering, which does not require specifying the number of clusters in advance \citep{Hand2007}, and recurse on the clustering procedure as long as there is a sufficient amount of problems within each cluster. In Section~\ref{sec_numerical_realdata}, we employ the bisect clustering concept on a real data set with a newsvendor cost function and use cross-validation to determine the threshold for the minimum number of problems per cluster.

\subsection{Connection with the Existing Approaches in the Literature}\label{SS_ConnectionOtherApproaches}
In addition to the direct data pooling approach proposed by \cite{GuptaKallus2022}, the cluster-based data pooling approach is also related to the Data Aggregation with Clustering (DAC) approach introduced by \cite{CohenZhangJiao2022}. The DAC approach aims to improve demand prediction accuracy and aggregates data based on the cluster structure of the coefficients of covariates. One can apply the DAC approach to our problem setting under MSE by considering a single-dimensional constant covariate.


To illustrate the differences and relationships between the cluster-based data pooling approach, the DAC approach, and the direct data pooling approach, we first present three basic data-driven approaches under MSE given the historical data, $\{S_k,k \in \mathcal{K}\}$. (i) {\em No aggregation}: This decoupling approach treats each problem separately and applies the SAA approach to each problem solely based on respective data. That is, $x_k = \hat{\mu}_k, k \in \mathcal{K}$. (ii) {\em Na\"ive aggregation}: one may aggregate the data of all problems and assign the grand mean to the decision of each problem. That is, $x_k = \frac{1}{NK} \sum_{j=1}^K \sum_{i=1}^N \hat{\xi}_{j, i} =  \frac{1}{K} \sum_{j=1}^K \hat{\mu}_j$, where $\hat{\mu}_j$ is the sample mean estimate of problem $j$. (iii) {\em Cluster-based aggregation}: the third approach segments all problems into clusters and derives the decisions using the sample mean of the aggregated data in each cluster. That is, $x_k = \frac{1}{|\hat{C}_i|}\sum_{j \in \hat{C}_i}^{K}\hat{\mu}_j, k \in \hat{C}_i$ where $(\hat{C}_1, \hat{C}_2)$ denotes the cluster structure. 
\begin{figure}[htbp!]
    \centering
     \includegraphics[width=1\linewidth]{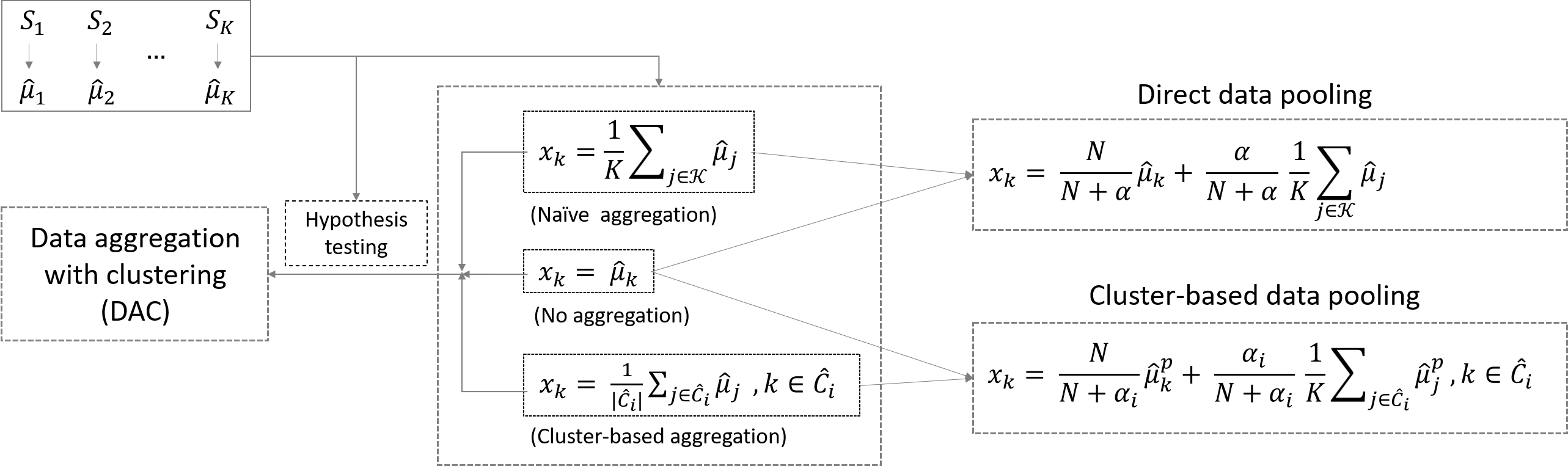}
    \caption{Data-driven approaches with limited data under MSE}
    \label{fig:DAC}
\end{figure}

As shown in Figure \ref{fig:DAC}, the DAC approach decides which basic data-driven approaches to adopt via hypothesis testing. Specifically, the null hypothesis is that the coefficients of covariates from two problems are the same. If the percentage of the hypothesis tests that are accepted at confidence level $(1-\theta)$ is above (below) a threshold, $R_U$ ($R_L$), then one should implement the na\"{i}ve aggregation (decoupling) approach. Otherwise, the cluster-based aggregation should be employed. The direct data pooling approach aims to strike a balance between the direct aggregation and decoupling approaches via the shrinkage parameter. The cluster-based data pooling approach flexibly utilizes the data for clustering and pooling, and thus, may inherit the advantages of three basic approaches. The data of each problem are divided into two subsets for clustering and pooling, respectively. It not only leverages the (estimated) cluster structure of all problems to aggregate problems with high degrees of similarity but also connects the cluster-based aggregation and decoupling approaches with shrinkage parameters, which are tuned for respective clusters. 

\begin{figure}[htbp!]
    \centering
     \includegraphics[width=0.6\linewidth]{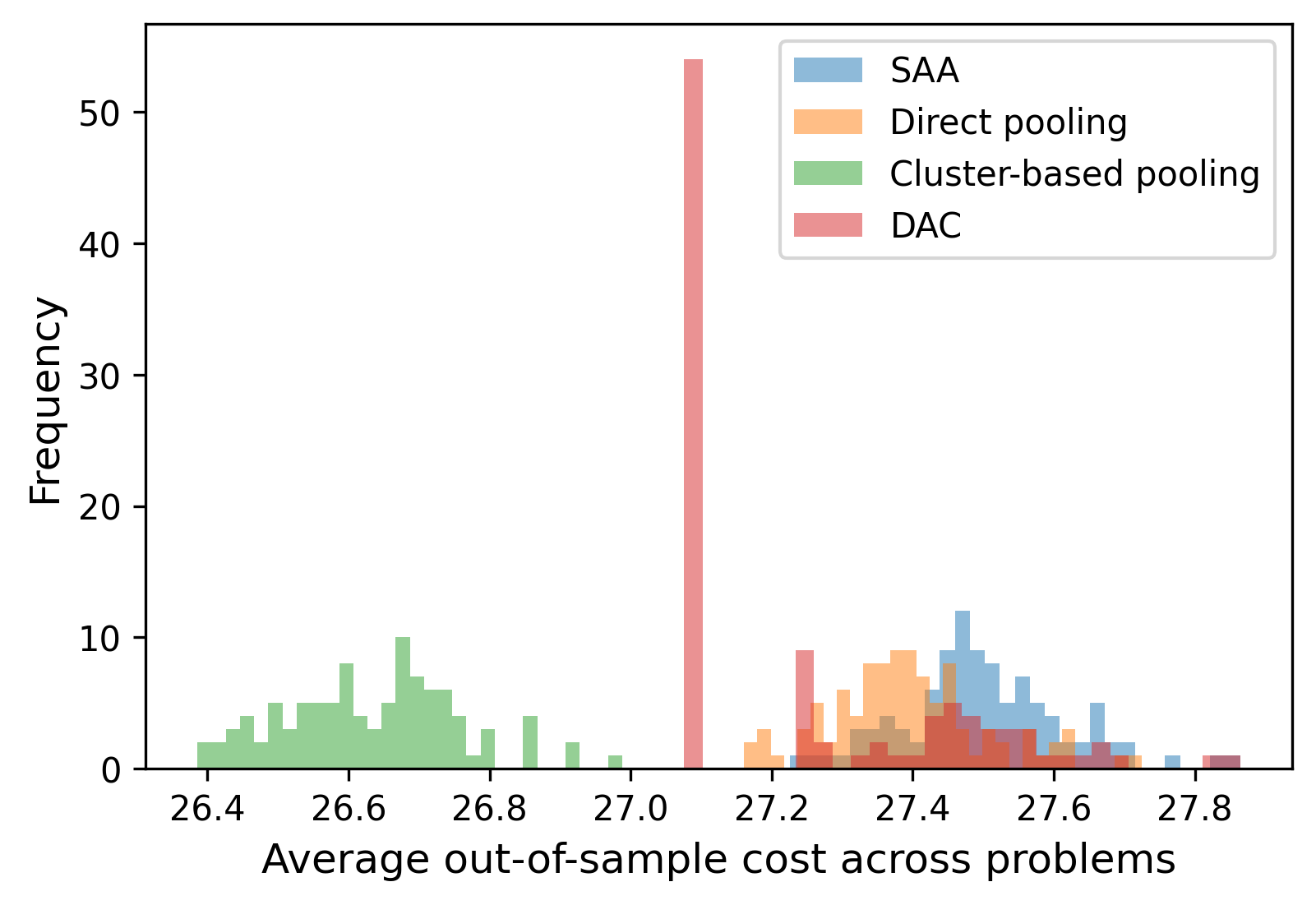}
    \caption{Histogram of average out-of-sample 
    costs  with 1000 randomly generated instances}
    \label{Fig_AlgorithmComparisionMSE}
    
\end{figure}

To investigate the performance of the proposed approach as well as those state-of-art ones, we conduct a numerical experiment with the setup described in Figure~\ref{fig:MSE_whole_picture} (a). The number of observations for clustering, $N_1$, is set to 5 for the cluster-based data pooling approach. The confidence level $\theta$ is set to 0.05 and the thresholds $R_U$ and $R_L$ are set to 0.9 and 0.4, respectively, for the DAC approach.\footnote{If $R_L$ is set to 0.6, following the configuration in \cite{CohenZhangJiao2022}, then the ratio of accepting the hypothesis in our experiments does not exceed the threshold $R_L$, and thus, the DAC approach is equivalent to the SAA approach.} We randomly generate 100 instances, and report the histogram of the average out-of-sample cost of each problem using the decisions obtained by different approaches in Figure \ref{Fig_AlgorithmComparisionMSE}. As shown in Figure \ref{Fig_AlgorithmComparisionMSE}, the cluster-based data pooling approach outperforms the benchmark approaches, meanwhile, the SAA approach results in the worst performance. As suggested by the discussion in Section \ref{sec_mse_unknown}, the benefit of clustering over the direct data pooling approach is significant, especially when the distance between two clusters is moderate or large. 

The DAC approach exhibits robust performance but occasionally results in a similar performance as that of the decoupling approach. It is because, in most cases, it employs the cluster-based aggregation approach that assigns the same decision for all problems within each cluster. On the other hand, in a few cases, it derives the decisions of each problem separately, which is equivalent to the SAA approach. In either case, the DAC approach results in higher out-of-sample costs than that of the cluster-based data pooling approach. The intuition is that the latter contains a larger solution space than that of the former. Consider a special case of two clusters, the DAC approach implemented by the oracle recovers the underlying cluster structure, and sets the average of true means of problems in each cluster (i.e., $\mu_{i,0}^\mathrm{AP}$) as the decisions for all problems within that cluster. When the cluster-based data pooling approach is implemented by an oracle and the shrinkage parameter $\alpha_i$ for cluster $C_i$ is set to $\infty$, the corresponding Shrunken-SAA decisions give the solutions obtained by the DAC approach. In other words, the cluster-based data pooling approach implemented by the oracle searches over a larger solution space and thus should result in a lower out-of-sample cost than that of the DAC approach. This may explain why the cluster-based data pooling approach outperforms the DAC approach in a data-driven setting under MSE.

\section{Cluster-based Data Pooling for General Cost} \label{sec_ClusterGeneral}
In this section, we present the cluster-based data pooling algorithm under general cost functions. However, different from the analysis under MSE, it is challenging to derive the closed forms for shrinkage parameters for general cost functions, which can provide implications for computing the data-driven ones. Therefore, once the cluster structure is estimated via a clustering algorithm, we adopt the Shrunken-SAA algorithm proposed by \cite{GuptaKallus2022} to determine the shrinkage parameters for respective clusters.

\begin{algorithm}[!ht]
  \footnotesize
  \caption{The Cluster-based Shrunken-SAA Algorithm}
  \label{alg_CSSA}
  \begin{algorithmic}[1]
    \Require 
     Data ${S}_k^c = \{ \hat{\xi}_{k,1},\dots, \hat{\xi}_{k,N_1}\}$, ${S}_k^p = \{ \hat{\xi}_{k,N_1+1},\dots, \hat{\xi}_{k,N_1+\hat{N}_k}\}, k \in \mathcal{K}$, the statistics $\mathcal{M}$ and two anchor distributions $\{h_i, i=1, 2\}$
     \State{$(\hat{C}_1,\hat{C}_2) \leftarrow \text{CLUST}(\mathcal{K},\{{S}_k^c, k \in \mathcal{K} \},\mathcal{M} )$} // Identify the cluster structure with $\mathcal{S}^c$ using clustering algorithm $\text{CLUST}$
     \For{$i = 1, 2$}
     \State{Fix a finite grid $\mathcal{A} \subseteq [0,\infty)$}
    \For{$\alpha \in \mathcal{A},k \in \hat{C}_i,j = 1,\cdots, \hat{N}_k$}
       \State{$x_{k,-j}(\alpha,h_i, S_{k,-j}^p) \leftarrow \argmin_{x_k \in \mathcal{X}_k} \sum_{l = 1}^{\hat{N}_k} \mathds{1}_{l \neq j} \cdot c_k(x_k,\hat{\xi}_{k,N_1+l}) + \alpha \mathbb{E}_{\xi_k \sim h_i}\big[c_k(x_k,\xi_k)\big]$ } 
       \State{// Compute leave-one-out (LOO) solution}
    \EndFor
    \State{$\alpha_{h_i,\hat{C}_i}^\mathrm{CS-SAA} \leftarrow \argmin_{\alpha \in \mathcal{A}} \sum_{k \in \hat{C}_i}\sum_{j =  1}^{\hat{N}_k} c_k\big(x_{k,-j}(\alpha,h_i,S_{k,-j}^p),\hat{\xi}_{k,N_1+j}\big)$}// Modified LOO cross-validation
    \For{all $k \in \hat{C}_i$}
        \State{$x_{k}^\mathrm{CS-SAA} \leftarrow \argmin_{x_k \in \mathcal{X}_k} \sum_{j= 1}^{\hat{N}_k} c_k(x_k,\hat{\xi}_{k,N_1+j}) + \alpha_{h_i,\hat{C}_i}^\mathrm{CS-SAA} \mathbb{E}_{\xi_k \sim h_i}\big[c_k(x_k,\xi_k)\big]$ }
        \State{// Compute cluster-based Shrunken SAA solution}
    \EndFor

     \EndFor
    
    \Ensure $(\hat{C}_1, \hat{C}_2), (x_1^\mathrm{CS-SAA},\dots,x_K^\mathrm{CS-SAA})$
  \end{algorithmic}
\end{algorithm}

Algorithm \ref{alg_CSSA} formally states the cluster-based Shrunken-SAA algorithm that combines a clustering algorithm and the Shrunken-SAA algorithm under general cost. In the first phase, a generic clustering algorithm, denoted by CLUST, is adopted to identify the cluster structure using the clustering data set $S^c$ and the statistics $\mathcal{M}$. Specifically, the distance between problem instances is measured by the difference in the statistics of the respective data sets. The statistics $\mathcal{M}$ indicates which type of statistics is obtained based on the data set, such as the sample mean or the sample quantile. In the following analysis, we adopt a specific clustering algorithm, which is described in Algorithm \ref{alg_Cluster}, that partitions the problems by identifying the boundary between two clusters.\footnote{One can definitely adopt a more sophisticated clustering algorithm to estimate the cluster structure. We adopt this simple clustering algorithm that enables us to provide the theoretical guarantees of the data pooling approach. In addition, we also demonstrate the effectiveness of such clustering algorithms via numerical experiments.} Specifically, we first compute the statistics $\mathcal{M}$ for each problem, and then, compute the boundary of the clusters by computing the mean of the statistics. Finally, the problems are divided into two clusters by comparing their statistics with the boundary of the clusters. If the statistics $\mathcal{M}$ is sample mean, this clustering algorithm resembles the one that we implemented under MSE in Section \ref{sec_mse_unknown}.  

 In the second phase, the Shrunken-SAA algorithm is adopted to determine the shrinkage parameters given the anchor distribution $h_i$ via the modified leave-one-out (LOO) validation for each estimated cluster. The choice of $h_i$ is flexible. One can select a fixed distribution or determine it via data-driven procedures. For example, one can assume the anchor $h_i$ follows a specific family of distributions, e.g., Gaussian distribution, and estimate the parameters by aggregating the data, or one simply treats the empirical distribution of the aggregate data as the data-driven anchor $h_i$. The key difference between the proposed cluster-based Shrunken-SAA algorithm and the one in \cite{GuptaKallus2022} is that we allocate a portion of data points for clustering and estimate the cluster structure before implementing the Shrunken-SAA algorithm. As the discussion in Section \ref{sec_DataPooling_MSE} may suggest, one may leverage the estimated cluster structure to enhance the benefits of the data pooling approach.
\begin{algorithm}[!ht]
  \footnotesize
  \caption{CLUST($\mathcal{C}$,$S^c$,$\mathcal{M}$)}
  \label{alg_Cluster}
  \begin{algorithmic}[1]
    \Require A set of problems $\mathcal{C}$, Data for clustering  $S^c = \{S_k^c, k \in \mathcal{C}\}$, statistics $\mathcal{M}$
     \State{$\hat{\mathcal{C}_1}$ $\leftarrow \emptyset$, $\hat{\mathcal{C}_2}$ $\leftarrow \emptyset $} // Initialize the clusters
    \For{$k \in \mathcal{K}$}
    \State{$\hat{s}_k = \mathcal{M}(S_k^c)$} // Compute the statistics of each problem
    \EndFor
      \State{$C_b \leftarrow \frac{1}{|\mathcal{C}|}\sum_{k \in \mathcal{C}} \hat{s}_k$} // Compute the cluster boundary
     \For{$k \in \mathcal{C}$}
      \If{$\hat{s}_k \le C_b$}
      \State{$\hat{\mathcal{C}_1} \leftarrow \hat{\mathcal{C}_1} \cup k$}
      \Else
    \State{$\hat{\mathcal{C}_2} \leftarrow \hat{\mathcal{C}_2} \cup k$}
     \EndIf
     \EndFor
    \Ensure $(\hat{\mathcal{C}_1},\hat{\mathcal{C}_2})$
  \end{algorithmic}
\end{algorithm}

To measure the performance of the cluster-based Shrunken-SAA algorithm, we define the oracle's shrinkage parameter for the cluster-based Shrunken-SAA approach given the anchors $h_i$. Such shrinkage parameters are determined by validating with the true distribution information given the realized data for pooling. 
\begin{equation}
    \alpha_{h_i, C_i}^\mathrm{OR}(S^p)\in \argmin_{\alpha \ge 0} Z(\alpha, h_i, {C_i}|S^p) = \sum_{k \in C_i}  \mathbb{E}_{\xi_k}[c_k(x_k(\alpha,h_i,S^p_k),\xi_k)], i = 1, 2.  \label{oracle_alpha} 
\end{equation}
The oracle's shrinkage parameter $\alpha_{h_i,C_i}^\mathrm{OR}(S^p)$ depends on the observed $S^p$, and thus, is random.  By definition, the cost $\sum_{i\in\{1,2\}} Z( \alpha_{h_i,C_i}^\mathrm{OR}(S^p), h_i, {C_i}|S^p)$ provides the lowest possible cost with anchor $h_i$ given the cluster structure $(C_1, C_2)$. Note that $\alpha_{h_i,C_i}^\mathrm{OR}(S^p)$ is different from the shrinkage parameter $\alpha_{h_i}^\mathrm{AP}$, where the former is the sample-path-wise optimal shrinkage parameter, and the latter is the optimal a priori one (recall Equation \eqref{eqn_OptimalShrinkageCluster}). However, computing $\alpha_{h_i,C_i}^\mathrm{OR}(S^p)$ requires the cluster structure $(C_1, C_2)$ and distribution information $\{\mathbb{P}_k, k\in\mathcal{K}\}$, which are not available in practice. With slight abuse of notation, we omit $S^p$ in the arguments of the shrinkage parameter $\alpha_{h_i, C_i}^\mathrm{OR}(S^p)$ and the out-of-sample cost $Z(\alpha_{h_i, C_i}^\mathrm{OR}, h_i, {C_i}|S^p)$ in the following discussion.

We measure the suboptimality of cluster-based data pooling with data-driven shrinkage parameters $(\hat{\alpha}_1, \hat{\alpha}_2)$ based on the estimated cluster structure $(\hat{C}_1, \hat{C}_2)$ given the anchors $(h_1, h_2)$ as follows,
\begin{align*}
    \text{SubOpt}_{h_1,h_2}(\hat{\alpha}_1, \hat{\alpha}_2, \hat{C}_1, \hat{C}_2) = \frac{1}{K}\bigg(\sum_{i\in \{1,2\}}Z(\hat{\alpha}_i, h_i, {\hat{C}_i}) - \sum_{i\in \{1,2\}}Z(\alpha_{h_i,C_i}^\mathrm{OR}, h_i, C_i) \bigg).
\end{align*}
As indicated by the above equation, there are two sources of suboptimality. One is that the estimated cluster structure could be inaccurate, especially obtained with a limited number of data points. The other is that the shrinkage parameters obtained via the data-driven procedure differ from the oracle parameters.  In this section, we assume the following conditions before characterizing the suboptimality of cluster-based data pooling under general cost.

\begin{assumption}\label{asmp_4} The probability distribution $\mathbb{P}_k$, the anchor distribution $h_i$, and the cost function $c_k$ satisfy the following conditions.
\begin{itemize}
    \item[(i)] {\bf (Compact support)} There exists a compact set $\Xi \subseteq R$ such that, for each $k \in \mathcal{K}$, $\xi_k \sim \mathbb{P}_k$ is a real random variable whose support is contained in $\Xi$ and the support of $h$ is contained in $\Xi$ w.p. 1. In addition, there exists a constant $\Pi_1$, such that $\max\ \Xi - \min\ \Xi \le \Pi_1$.
    \item[(ii)] {\bf (Equicontinuity)}  For each $k \in \mathcal{K}$, $\{c_k(x_k,\xi_k): x\in \mathcal{X}_k\}$ is equicontinuous in $\xi_k$ for all $\xi_k \in \Xi$. Namely, for any $\epsilon > 0$, $\xi_k \in \Xi$, there exists $\delta > 0$ such that if $|\xi_k - \xi^{'}_k| \le \delta$, then $|c_k(x_k,\xi_k) - c_k(x_k,\xi^{'}_k)| \le \epsilon$ for all $x \in \mathcal{X}_k$.
    \item[(iii)] {\bf (Lipschitz, strongly-convex cost)} There exist constants $L,\gamma$ such that $c_k(x_k,\xi)$ are  $L-Lipschitz$ and $\gamma-strongly\ convex$  over $\mathcal{X}_k$. Moreover, $\mathcal{X}_k$ is non-empty and convex for $k \in \mathcal{K}$ and $\xi \in \Xi$.
    \item[(iv)] {\bf (Bounded optimal cost)} For any dataset $S$, there exists a constant $\Pi_2$ such that  $\sup_{h: P(h \in \Xi) =1}|c_k(x_k(\infty, h, S),\xi)| \le \Pi_2$ for any $k \in \mathcal{K}$ and $\xi \in \Xi$.\footnote{With a slight abuse of notation, $x_k(\alpha, h, S_k)$ denotes the Shrunken-SAA solution with the shrinkage parameter $\alpha$, the anchor distribution $h$ and the dataset $S_k$ under the general cost. With $\alpha$ set to $\infty$, the Shrunken-SAA solution completely depends on the choice of the anchor distribution $h$.}
    \item[(v)] {\bf (Randomizing amount of data)} The number of data points for clustering $N_1$ is fixed and the number of data points for pooling for problem $k$, $\hat{N}_k \sim Poisson(\lambda)$.
\end{itemize}
\end{assumption}

 Assumption \ref{asmp_4} (i) assumes that the random variables of all problems are supported on a compact set. Assumption \ref{asmp_4} (ii) and (iii) implies that the cost function $c_k(x_k,\xi_k)$ satisfies equicontinuity, strong convexity, and Lipschitz continuity. Assumption \ref{asmp_4}  (iv) assumes that the optimal cost associated with any policy is bounded. Assumption \ref{asmp_4} (v) essentially assumes that the total number of observations of each problem is random and greater than $N_1$ so that one can use these $N_1$ data points to group the problems into clusters. The above assumptions are commonly seen in the literature \citep[e.g.,][]{GuptaKallus2022}.

In the following, we characterize the suboptimality of the cluster-based Shrunken-SAA algorithm.

\begin{theorem}\label{thm_GeneralCost} When Assumption \ref{asmp_2} (i) and  Assumption \ref{asmp_4} hold, and the cluster structure $(\hat{C}_1, \hat{C}_2)$ is estimated by implementing Algorithm \ref{alg_Cluster} with the sample mean-based distance metric, given the fixed anchors $h_i, i= 1, 2$, there exists a universal constant $A$, such that for any $\delta \in (0,1)$, with probability at least $1-\delta$ such that,
\begin{align*}
    & \mathrm{SubOpt}_{h_1,h_2}(\alpha_{h_1,\hat{C}_1}^\mathrm{OR},\alpha_{h_2,\hat{C}_2}^\mathrm{OR}) \\
    \le & \underbrace{A \frac{\log^2(3/\delta) \log^{3/2}(K)}{\sqrt{K}}+ \sqrt{2}\Pi_2\frac{\log^{1/2}(3/\delta)}{\sqrt{K}}}_{\emph{Optimality gap of data pooling}} 
    + \underbrace{4\Pi_2\exp\Big(\frac{-Kd^2}{8(d+1)^2}\Big) + 2\Pi_2\exp\Big(\frac{-N_1d^2(b-a)^2}{32\Pi_1^2}\Big).}_{\emph{Loss due to estimated cluster structure}} 
\end{align*}
\end{theorem}

Theorem \ref{thm_GeneralCost} provides a theoretical guarantee on the performance of the cluster-based data pooling approach under the general cost function. When the number of problems is sufficiently large, the optimality gap decreases exponentially with respect to the distance between two clusters. In particular, the first two terms indicate the optimality gap associated with the data pooling process. As the number of problems grows, the estimated shrinkage parameter via the data pooling approach is more accurate, and the optimality gap diminishes. The last two terms in the upper bound represent the optimality gap caused by the inaccurate estimation of the cluster structure. In particular, the third term relates to the loss due to clustering with an inaccurate cluster boundary that may reduce the degree of similarity among problems within clusters. When the number of problems grows and the distance between clusters increases, one can obtain a more appropriate cluster boundary to group the problems. The fourth term is associated with the loss due to the probability of grouping a problem into a misclassified cluster and shrinking the SAA solution with the associated anchor and shrinkage parameter. Intuitively, the misclassification probability and the corresponding loss decrease exponentially in the distance between two clusters. In a nutshell, Theorem \ref{thm_GeneralCost} suggests that the cluster-based data pooling approach can yield close-to-optimal decisions under general cost when the number of problems grows and the distance between clusters increases. 

\section{Numerical Experiments}\label{sec_Numerical}
In this section, we demonstrate the effectiveness of cluster-based data pooling with general cost functions via numerical experiments. Specifically, we consider the newsvendor loss function, i.e., $c_k(x_k,\xi_k) = c_h (x_k - \xi_k)^+ + c_b (\xi_k - x_k)^+$ where $c_h$ denotes the per-unit holding cost and $c_b$ denotes the per-unit lost-sale penalty. Thus, the critical ratio for deriving the optimal newsvendor solution is computed as,  $s = \frac{c_b}{c_b + c_h}$. 
In Section \ref{sec_numerical_synthetic}, we conduct extensive numerical experiments with the synthetic data to demonstrate the advantage of cluster-based data pooling over direct data pooling. We also investigate the strengths and limitations of different distance metrics, which are critical for the cluster-based data pooling approach, and discuss the conditions under which one distance metric is preferred over the other. In Section \ref{sec_numerical_realdata}, we implement the cluster-based data pooling approach to real data collected from a European pharmacy chain to validate the performance of the proposed approaches. Since the number of clusters is not necessarily given in practice, we also devise a bisect clustering algorithm to automatically determine the number of clusters. 


\subsection{Impacts of distance metrics on the benefit of clustering} \label{sec_numerical_synthetic}
We set the number of newsvendor problems $K = 1000$, and the means of the random demands associated with each newsvendor problem are uniformly distributed, i.e., $\mu_k \sim \mathrm{U}(70,120)$. For each newsvendor problem, indexed by $k$, the associated random demand follows a Gaussian distribution, $\xi_k \sim \mathrm{N}(\mu_k, (v_k\mu_k)^2)$, where the coefficient of variation (CV), $v_k$, is sampled from a Gaussian distribution, i.e., $v_k \sim \mathrm{N}(0.2, v^2)$.\footnote{Since the coefficient of variation has to be positive, we can truncate the Gaussian distribution at zero in the numerical experiments. That is, $v_k \sim \mathrm{max}\{\mathrm{N}(0.2, v^2)$, 0\}.} The parameter $v$ captures the degree of heterogeneity of the CVs of random demands for different problems. For example, if $v = 0$, we have, $v_k = 0.2$ for any $k \in \mathcal{K}$, which implies all problems share the same CV. For each problem, the number of observed data points is $N=10$. 

To identify the similarity structure among problems, we need to choose the distance metric to measure the difference between problem instances. The first metric we consider is the difference between sample means of data sets of two problems, commonly used in clustering analysis with loss functions such as MSE. Since we consider the newsvendor loss function, the intuition may suggest that we group the problems with similar optimal solutions into the same cluster. Therefore, the second metric is the difference between the critical quantiles of the empirical distributions of historical data points of the two problems. For brevity, we refrains to the setting with two clusters of problems in this section, and focus on investigating the performance of the cluster-based data pooling approach with different metrics. In the following section, we will discuss the procedure to determine the number of clusters in the numerical analysis with real data. 

\begin{figure}[htbp!]
\centering
\subfigure[$s = 0.95,v = 0$]{
\begin{minipage}[t]{0.33\linewidth}
\centering
\includegraphics[width=2.2in,height=2.4in]{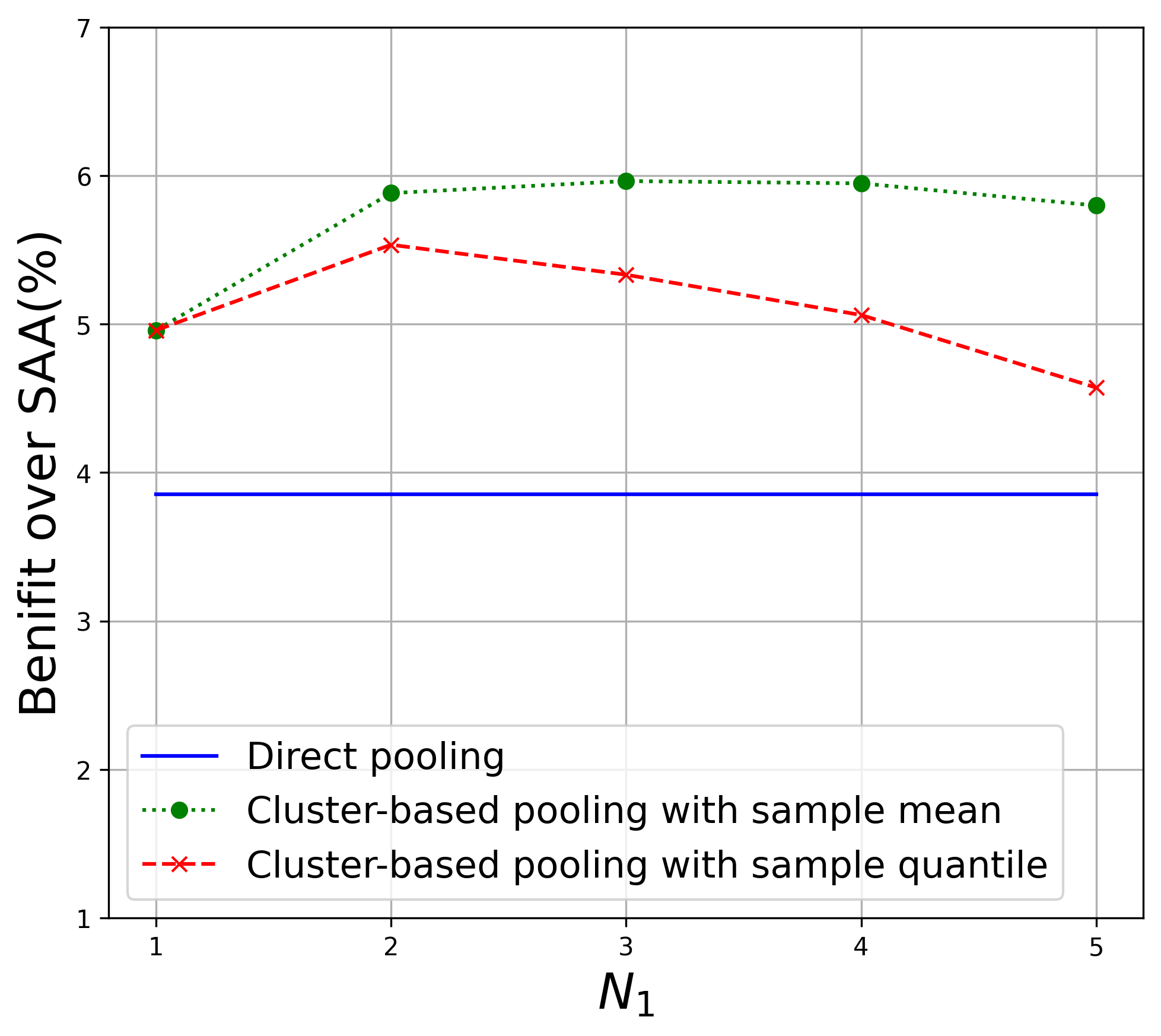}
\end{minipage}%
}%
\subfigure[$s = 0.95,v = 0.025$]{
\begin{minipage}[t]{0.33\linewidth}
\centering
\includegraphics[width=2.2in,height=2.4in]{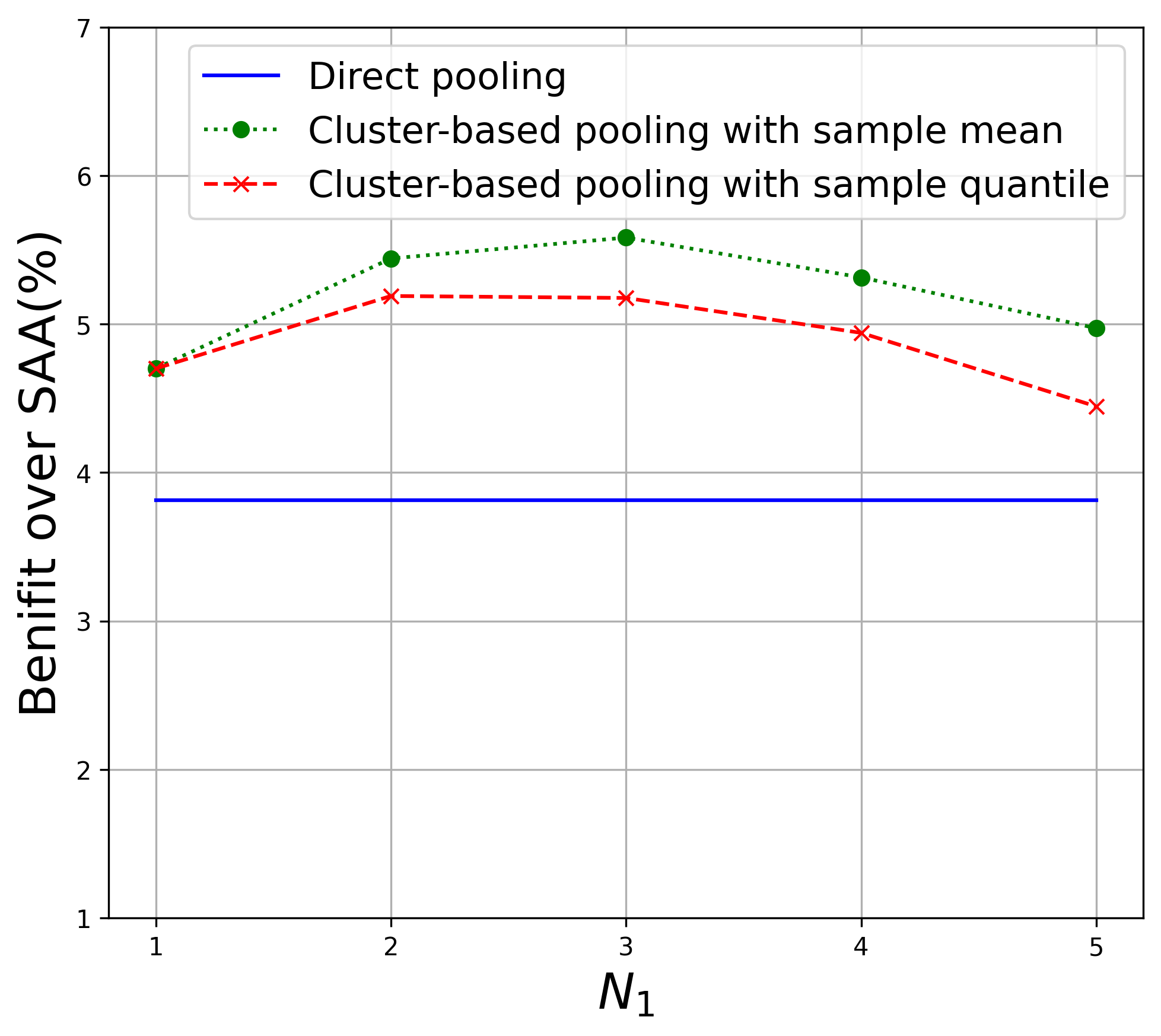}
\end{minipage}%
}%
\subfigure[$s = 0.95,v = 0.05$]{
\begin{minipage}[t]{0.33\linewidth}
\centering
\includegraphics[width=2.2in,height=2.4in]{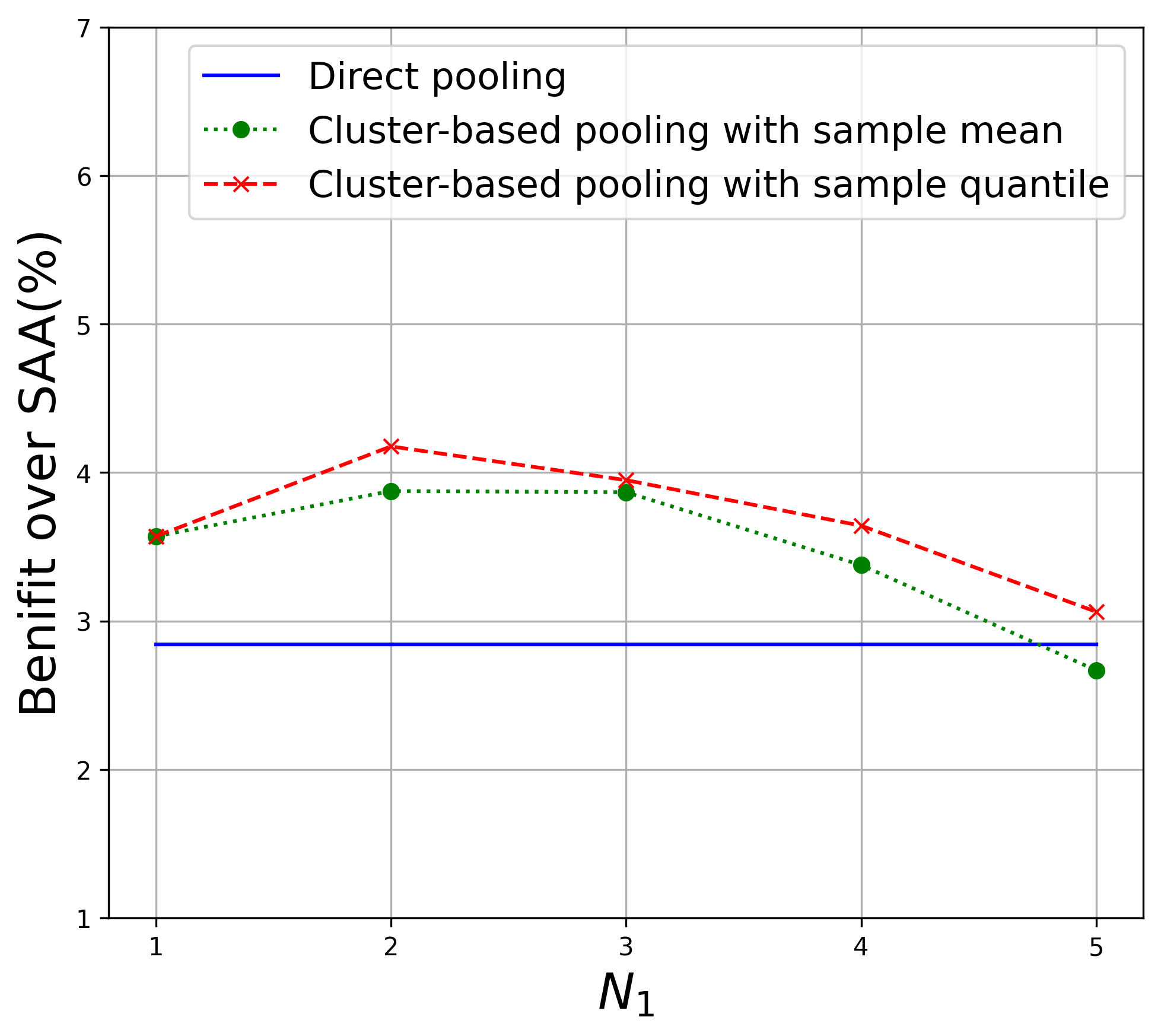}
\end{minipage}
}%

\subfigure[$s = 0.97,v = 0$]{
\begin{minipage}[t]{0.33\linewidth}
\centering
\includegraphics[width=2.2in,height=2.4in]{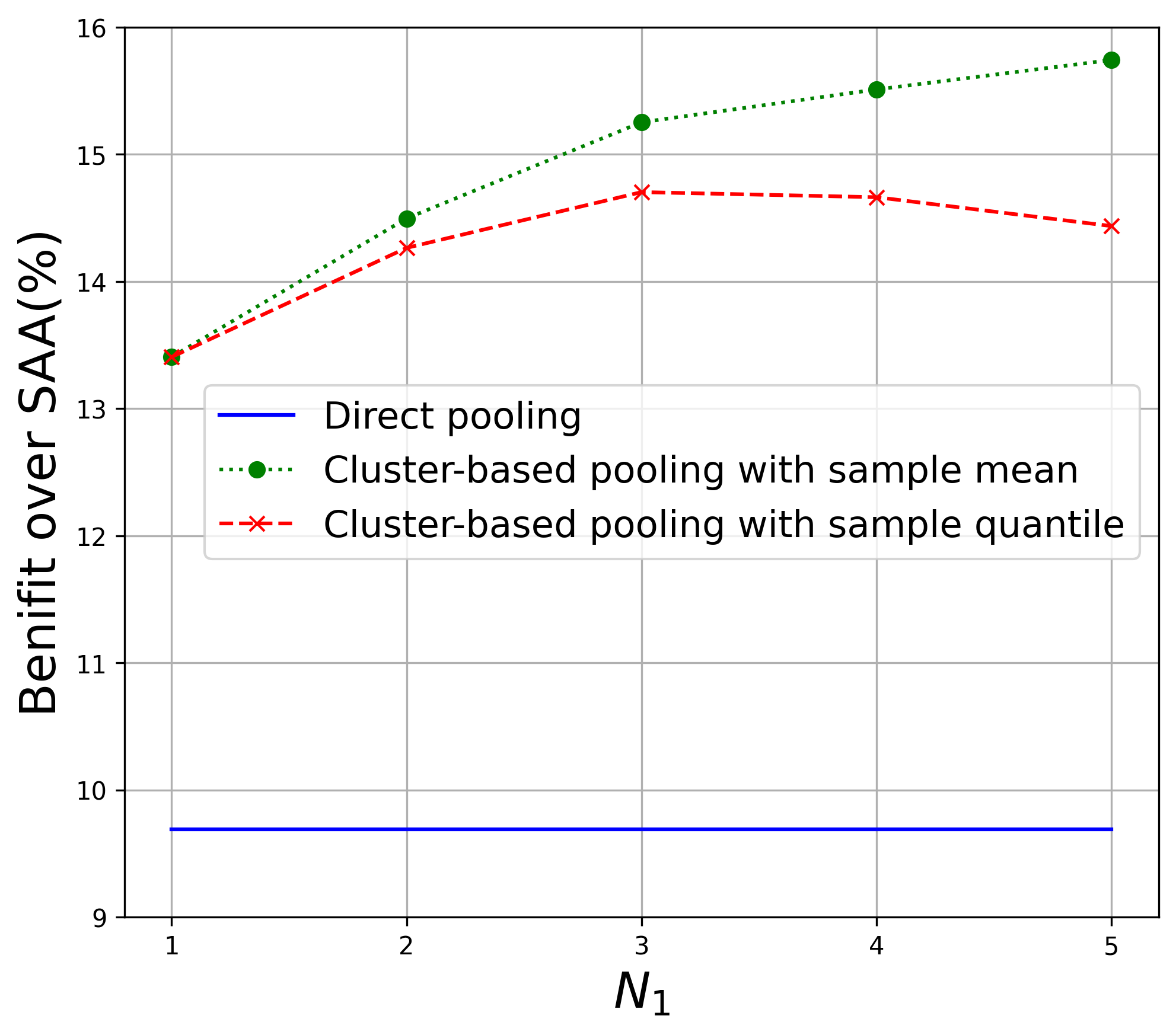}
\end{minipage}
}%
\subfigure[$s = 0.97,v = 0.025$]{
\begin{minipage}[t]{0.33\linewidth}
\centering
\includegraphics[width=2.2in,height=2.4in]{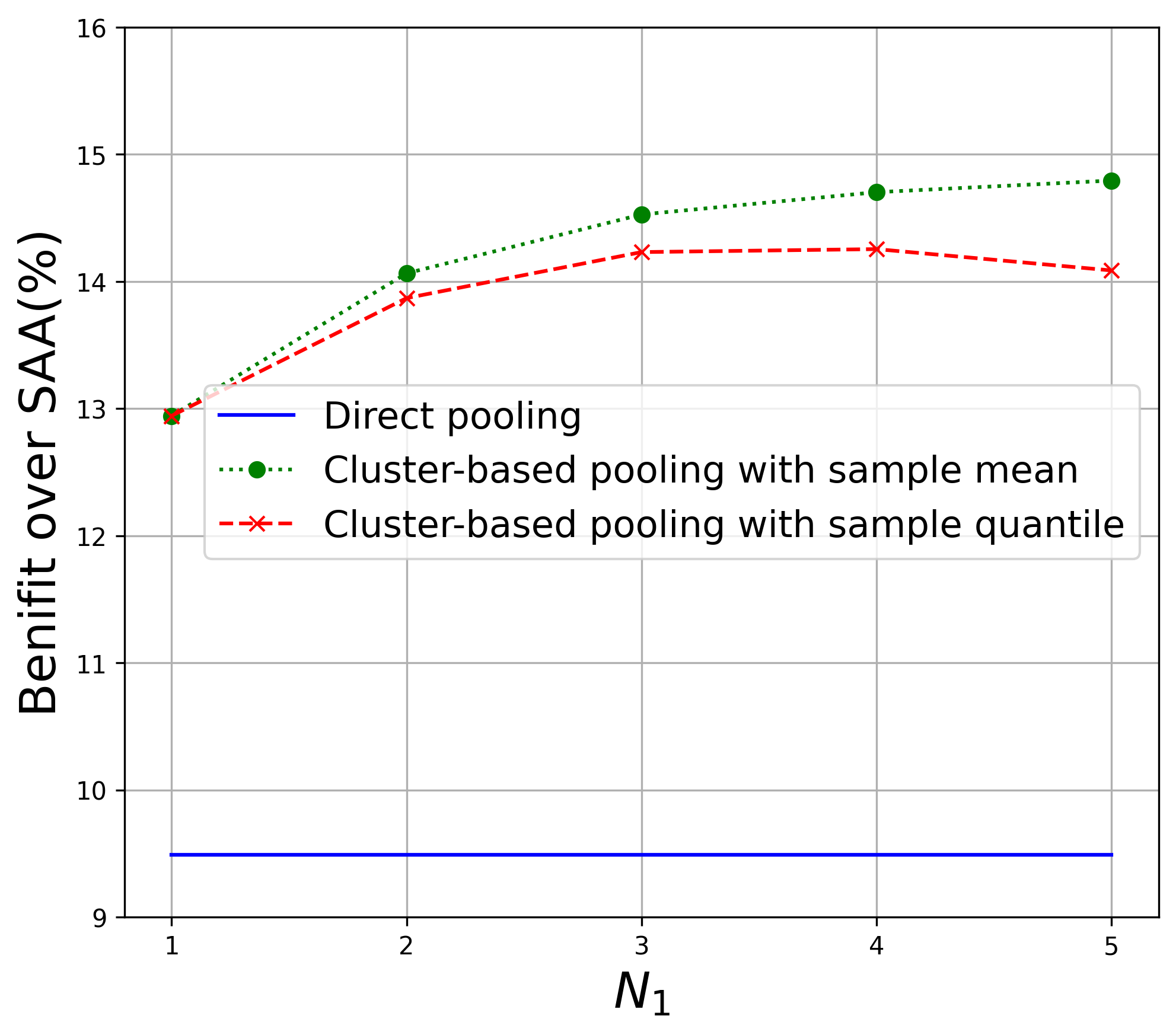}
\end{minipage}
}%
\subfigure[$s = 0.97,v = 0.05$]{
\begin{minipage}[t]{0.33\linewidth}
\centering
\includegraphics[width=2.2in,height=2.4in]{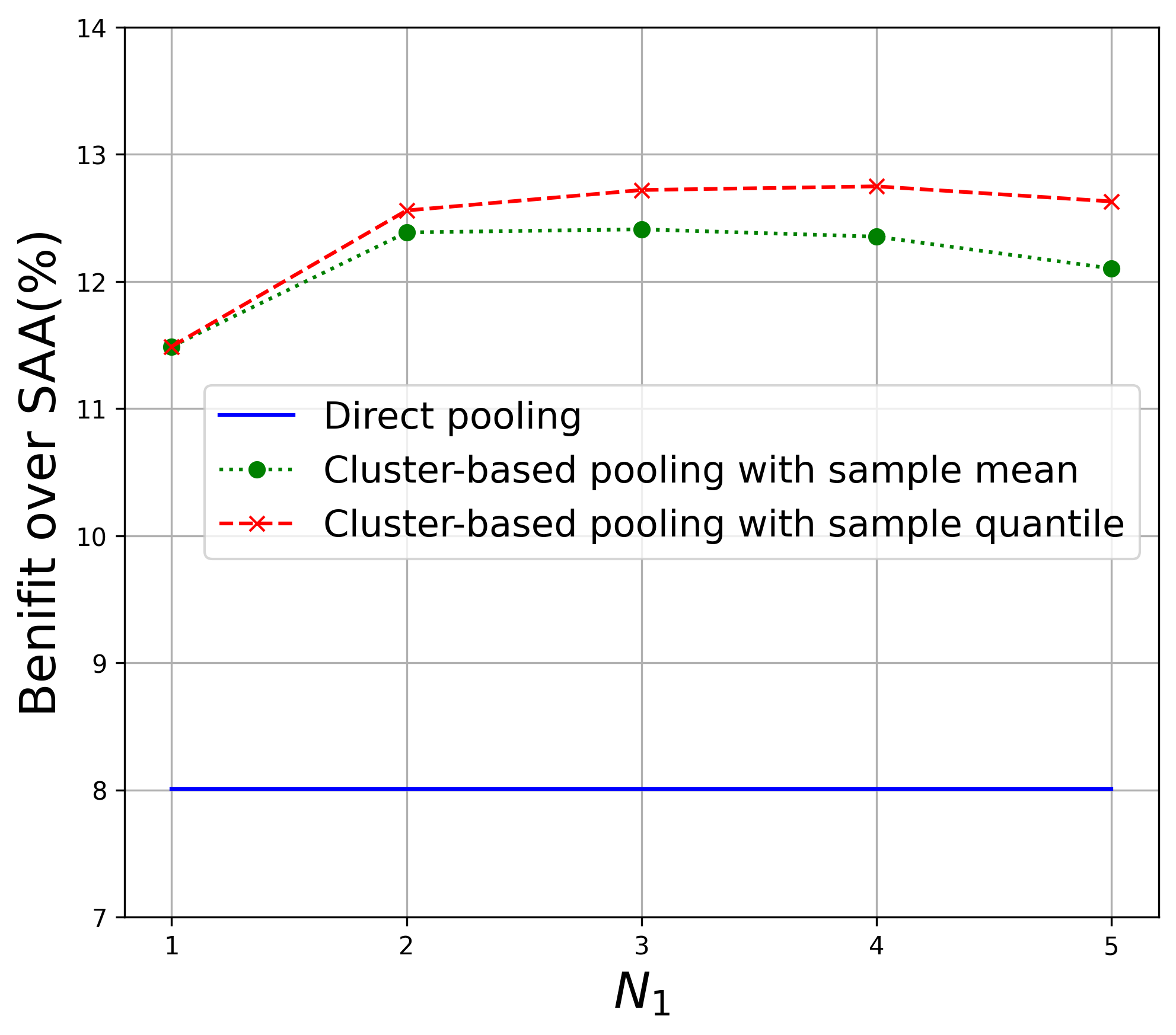}
\end{minipage}
}
\centering
\caption{Performance of data pooling approaches measured by $(Z_\mathrm{SAA}-Z_p)/Z_\mathrm{SAA} \times 100\%$.}
\label{fig_data_driven}
\parbox[t]{0.95\linewidth}{\footnotesize Notes. 
The above panels demonstrate the relative advantages of data pooling approaches over SAA. 
The mean of the random variable associated with problem $k$ is sampled from a uniform distribution, $\mu_k \sim \mathrm{U}(70,120)$. The random variable for problem $k$ is sampled from a Gaussian distribution $\xi_k \sim \mathrm{N}(\mu_k,(v_k\mu_k)^2)$ where the CV $v_k \sim \mathrm{N}(0.2,v^2)$. The degree of heterogeneity of CVs, $v$, is set to 0, 0.025, and 0.05, respectively. The critical ratio is $s \in \{0.95, 0.97\}$. The number of problems is $K= 1000$. Each reported relative advantage is the average of 100 instances. }
\end{figure}

We implement the cluster-based data pooling approach with two distance metrics based on Algorithms \ref{alg_CSSA} and \ref{alg_Cluster}. We also implement direct data pooling as a benchmark approach. We adopt the data-driven anchors to implement the data pooling approaches. In particular, the anchor distribution $h~(h_i)$ is the empirical distribution of the aggregated data of all problems (within cluster $C_i$). The discussion about the data-driven anchor can be found in Section \ref{sec_ClusterGeneral}. We compare the cluster-based data pooling approach with two distance metrics, the direct data pooling approach, and the SAA approach in the following numerical analysis. The degree of heterogeneity of CVs, $v$, is set to 0, 0.025, and 0.05, respectively, and the number of data points allocated for clustering analysis $N_1$ ranges from 1 to 5.  We use the decoupling approach that solves each problem with Sample Average Approximation (SAA) as the benchmark and calculate the ``Relative advantage over SAA (\%)'' according to the following equation. 
\begin{equation}
    \notag
    \text{Relative advantage over SAA (\%)} = \frac{Z_\mathrm{SAA} - Z_p}{Z_\mathrm{SAA}} \times 100\%,
\end{equation}
where $Z_p$ denotes the total cost obtained with the data pooling approaches. The performances of data pooling approaches are presented in Figure \ref{fig_data_driven}.

We first observe that cluster-based data pooling, in general, outperforms the direct data pooling approach, which illustrates the benefit of clustering before aggregating the data. In addition, the advantage of data pooling approaches over the SAA approach first increases and then decreases as the number of data points allocated for clustering analysis, $N_1$, increases. This is consistent with our observation under the MSE loss function (recall Figure \ref{fig:MSE_whole_picture} (a)). This suggests that if one would like to make the most use of the cluster-based data pooling approach, it is important to strike the right balance between identifying the appropriate cluster structure and shrinking the SAA solutions with a sufficient amount of aggregated data. Intuitively, the performance of direct pooling, which uses all data points for pooling, is not affected by the choice of $N_1$. 

\begin{figure}[!ht]
\centering
\subfigure[$s = 0.95,v = 0$]{
\begin{minipage}[t]{0.3\linewidth}
\centering
\includegraphics[width=2in]{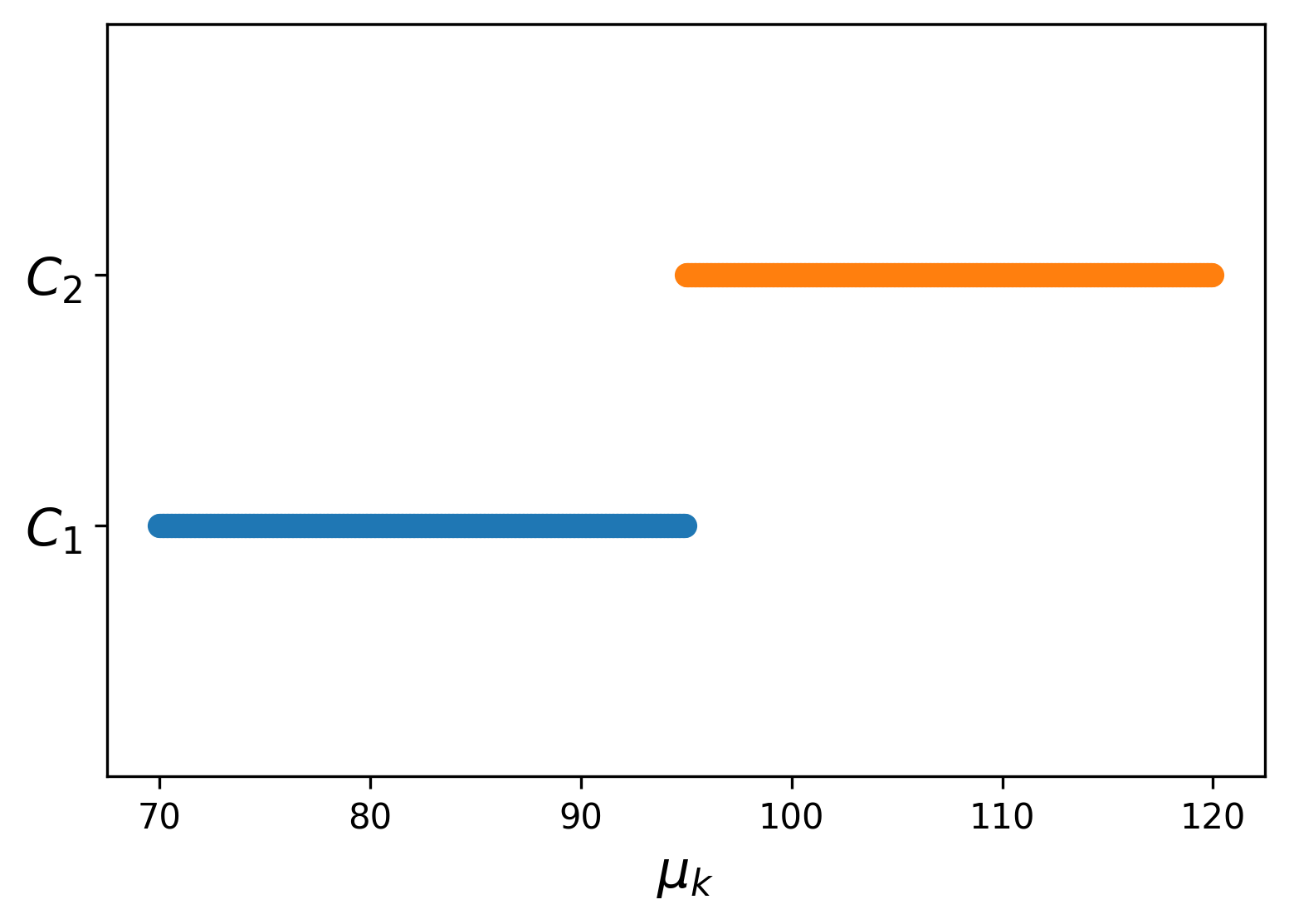}
\end{minipage}%
}%
\subfigure[$s = 0.95,v = 0.025$]{
\begin{minipage}[t]{0.3\linewidth}
\centering
\includegraphics[width=2in]{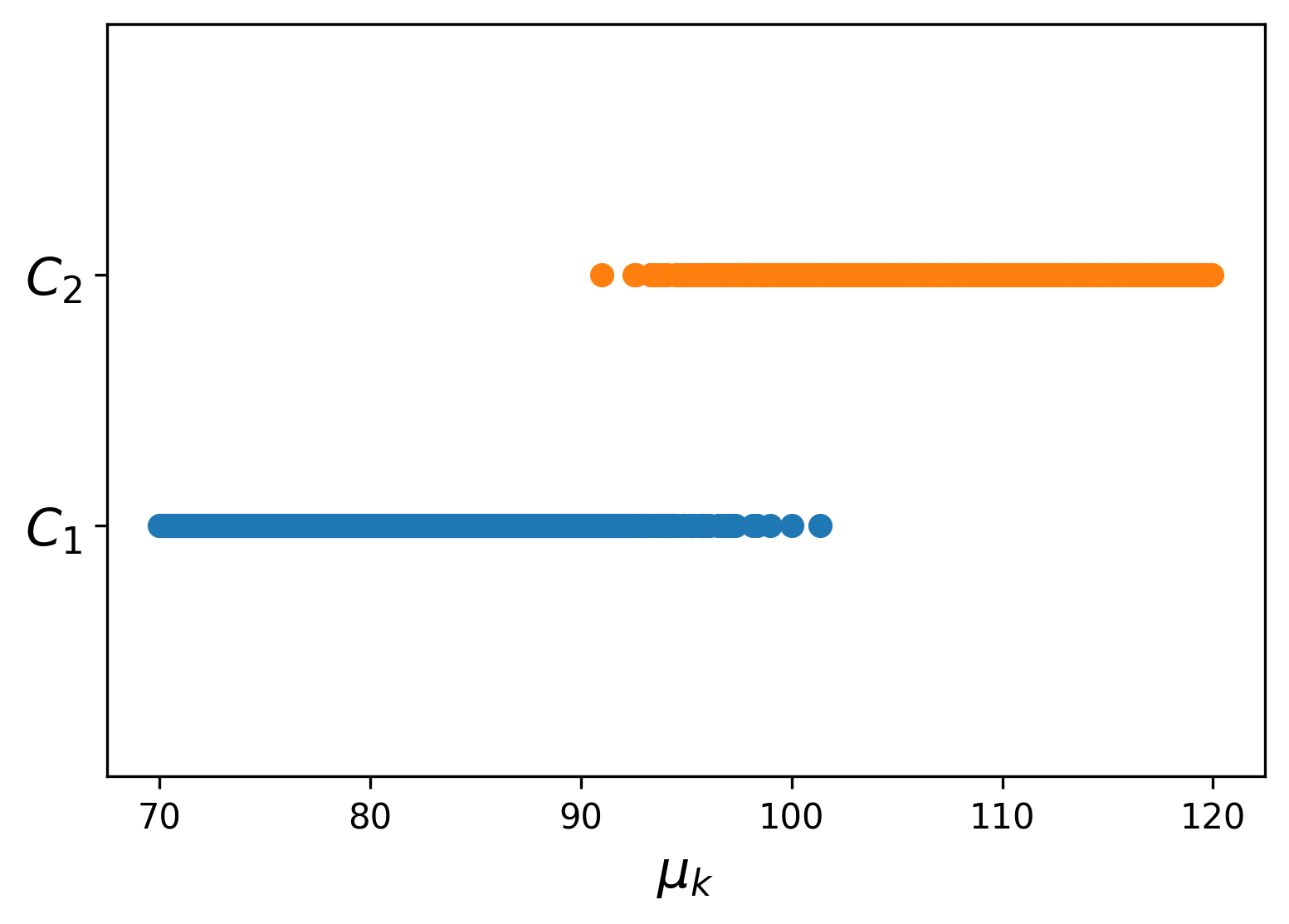}
\end{minipage}%
}%
\subfigure[$s = 0.95,v = 0.05$]{
\begin{minipage}[t]{0.3\linewidth}
\centering
\includegraphics[width=2in]{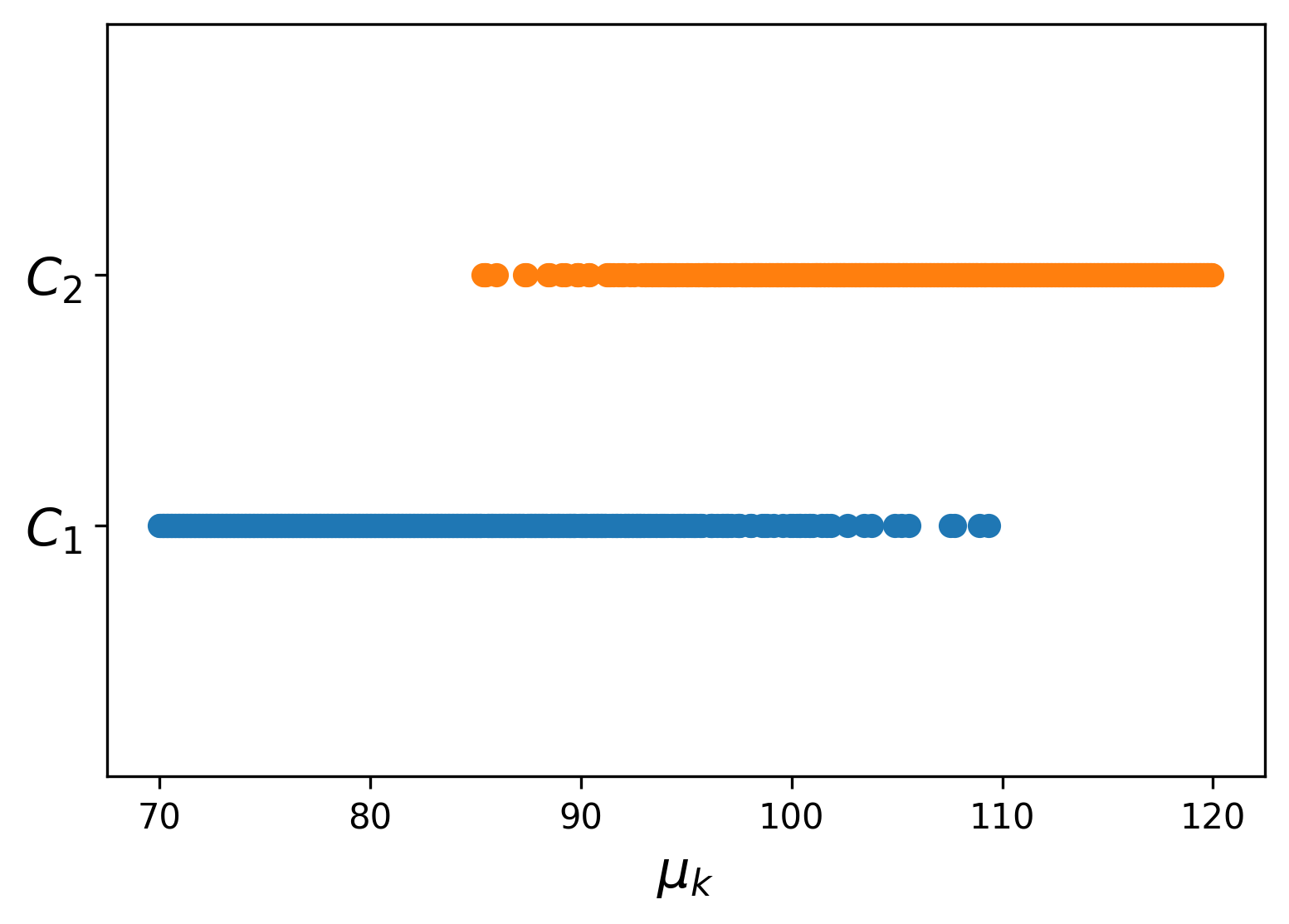}
\end{minipage}
}%

\subfigure[$s = 0.97,v = 0$]{
\begin{minipage}[t]{0.3\linewidth}
\centering
\includegraphics[width=2in]{sigma_0.png}
\end{minipage}
}%
\subfigure[$s = 0.97,v = 0.025$]{
\begin{minipage}[t]{0.3\linewidth}
\centering
\includegraphics[width=2in]{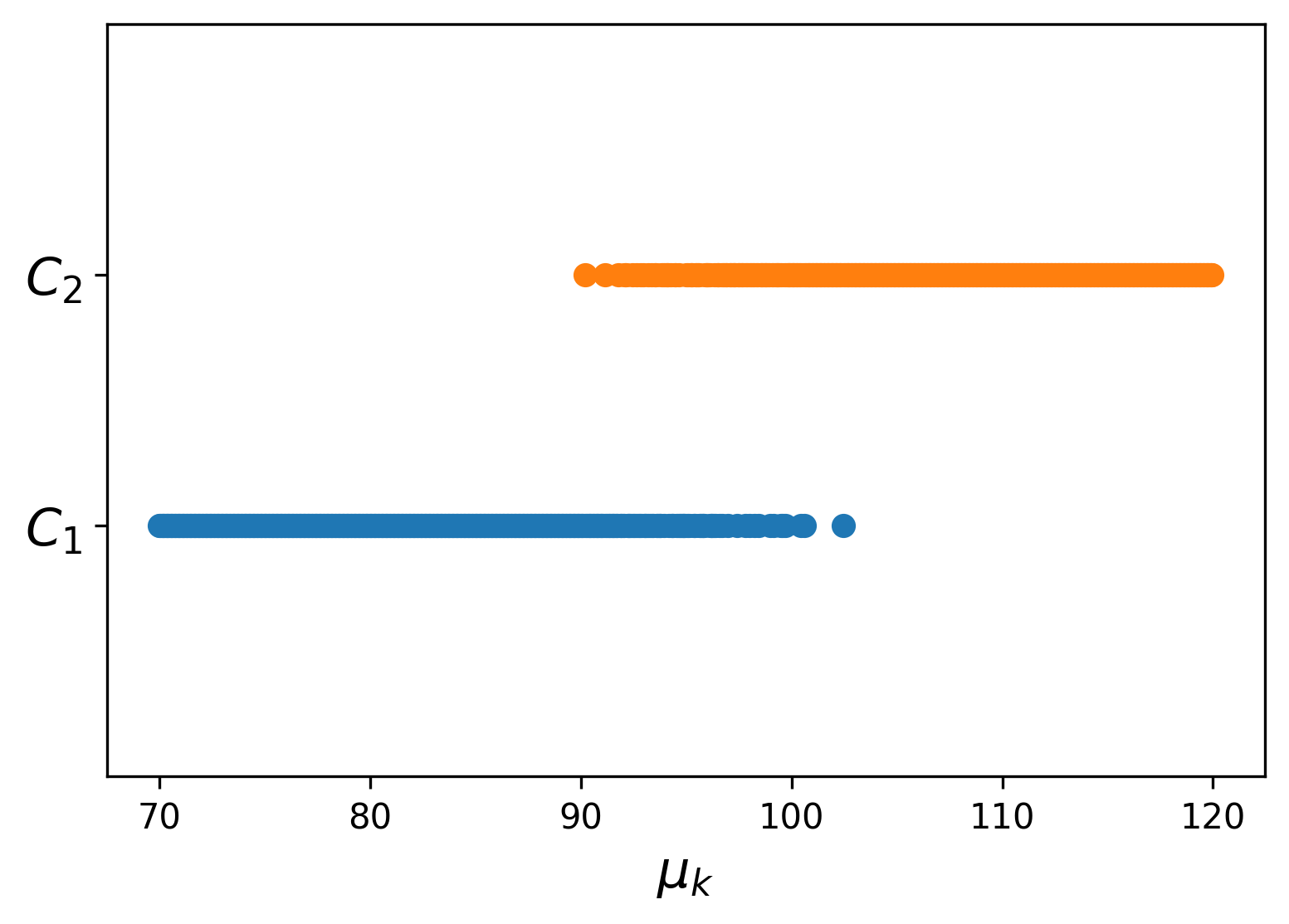}
\end{minipage}
}%
\subfigure[$s = 0.97,v = 0.05$]{
\begin{minipage}[t]{0.3\linewidth}
\centering
\includegraphics[width=2in]{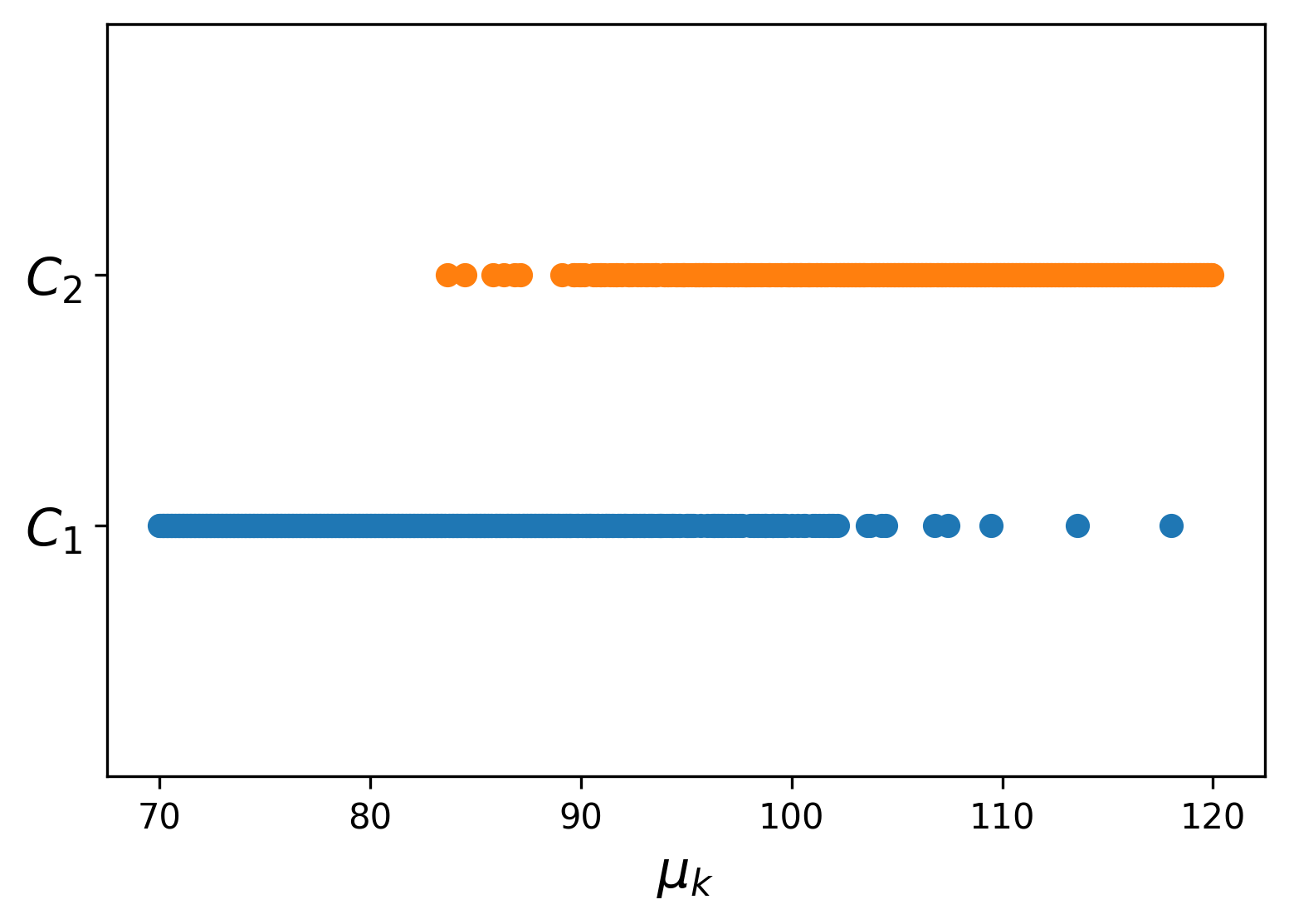}
\end{minipage}
}%
\centering
\caption{The oracle's cluster structures identified by the critical quantile-based metric}
\label{fig_optimal_cluster_diff_var}
\parbox[t]{0.95\linewidth}{\footnotesize Notes. The above panels demonstrate the oracle clusters identified with the critical quantile-based distance metric. The mean of the random variable associated with problem $k$ is sampled from a uniform distribution, $\mu_k \sim \mathrm{U}(70,120)$. The random variable for problem $k$ is sampled from a Gaussian distribution $\xi_k \sim \mathrm{N}(\mu_k,(\mu_k*v_k)^2)$ where the coefficient of variation $v_k \sim \mathrm{N}(0.2,v^2)$. The number of problems is $K=1000$. The degree of heterogeneity $v$ is set to 0, 0.025, and 0.05, respectively. The critical ratio is $s \in \{0.95, 0.97\}$.}
\end{figure}

With the newsvendor loss function, the intuition may suggest that the critical quantile-based metric should yield a more appropriate cluster structure than the sample mean-based one. Interestingly, we find that the cluster-based data pooling approach with sample mean-based metric outperforms the one associated with the critical quantile-based metric when $v=0$ or $v=0.25$, that is, the CVs of different problems are similar. 
To understand the impacts of the heterogeneity of CVs on the clustering outcomes with different distance metrics and the resulting data-pooling performance, we highlight the cluster structures identified by the oracle equipped with the true distribution information in Figure \ref{fig_optimal_cluster_diff_var}. In particular, for the sample mean-based metric, regardless of the distribution of CVs, one cluster contains the set of problems with true means $\mu_k \in  (70,95)$, and the other contains the set of problems with true means $\mu_k \in (95,120)$. On the other hand, for the critical quantile-based metric, the oracle's cluster structure depends on the value of $v$. As demonstrated in Figure \ref{fig_optimal_cluster_diff_var}, when $v = 0$, all problems share the same coefficient of variation, it is straightforward to observe that the oracle's cluster structure is the same as that of the sample mean-based metric for both critical ratios. However, as the degree of heterogeneity $v$ or critical ratio increases, there is more discrepancy between the cluster structures identified with the two metrics.

Therefore, when the parameter $v$ is small, the clustering analysis with the two distance metrics results in similar clustering outcomes. More importantly, the sample mean provides a minimum-variance unbiased estimate of the true mean, while the sample critical quantile estimate, especially using the empirical distribution of limited data points, can be biased \citep[see,][]{Parrish1990}.\footnote{We also conduct numerical experiments to investigate the bias and the variance of the sample mean and sample quantile estimates, and verify that the former results in unbiased estimate with small variance. We refer readers to Appendix \ref{appx_Numerical} for more details.}  In other words, even under the newsvendor loss function, the clustering algorithm with the sample mean-based metric can provide a more accurate estimate of the oracle's cluster structure. This may explain why the cluster-based data pooling approach with sample mean-based metric outperforms the one with critical quantile when the parameter $v$ is relatively small. However, as the parameter $v$ increases, the gap between the cluster-based data pooling approach with two metrics shrinks, and, their ranks may even swap when $v$ is relatively large. This indicates that when the CVs of random demands are dispersed, the cluster structure obtained with the sample mean-based metric is significantly different from the optimal structure under the newsvendor loss, and thus, results in an inferior performance relative to the one obtained with critical quantile-based metric. This suggests that the implementation of cluster-based data pooling approaches requires a careful choice of distance metric, which could depend on the heterogeneity of CVs of random variables for all problems.

\subsection{Implementation of the cluster-based Shrunken-SAA algorithm with real data} \label{sec_numerical_realdata}
In this subsection, we implement the proposed data pooling approaches as well as the benchmark approaches with the real data, which consists of daily sales at the store level for a European pharmacy chain with locations across 7 countries. We follow the data preprocessing procedure in \cite{GuptaKallus2022}.\footnote{For the detailed procedure for data preprocessing, readers can refer to https://github.com/vgupta1/JS\_SAA} The only difference is that we treat the historical sales data as continuous, while \cite{GuptaKallus2022} discretize the historical sales data with the maximum and minimum values inferred from the entire data set. 

{\bf Validation strategy.} After preprocessing the real data set, we have the historical sales data for 629 days from 1105 stores. Individual stores may be closed during these 629 days due to local holidays. Thus, almost all 1105 stores have at least one day with missing data and $16\%$ of stores have $20\%$ of days with missing data. We implement the repeated random subsampling validation to evaluate the proposed approaches. In particular, for each store, we randomly sample sales data for 10 days as the training set and reserve the remaining sales data as the test set. We repeat this procedure 100 times to mitigate randomness that arises in the sampling process. Due to the possibility of missing data, the number of effective training data points for store $k$, denoted by $\hat{N}_k$, might be smaller than 10. 

\begin{figure}[htbp!]
    \centering
     \includegraphics[width=0.6\linewidth]{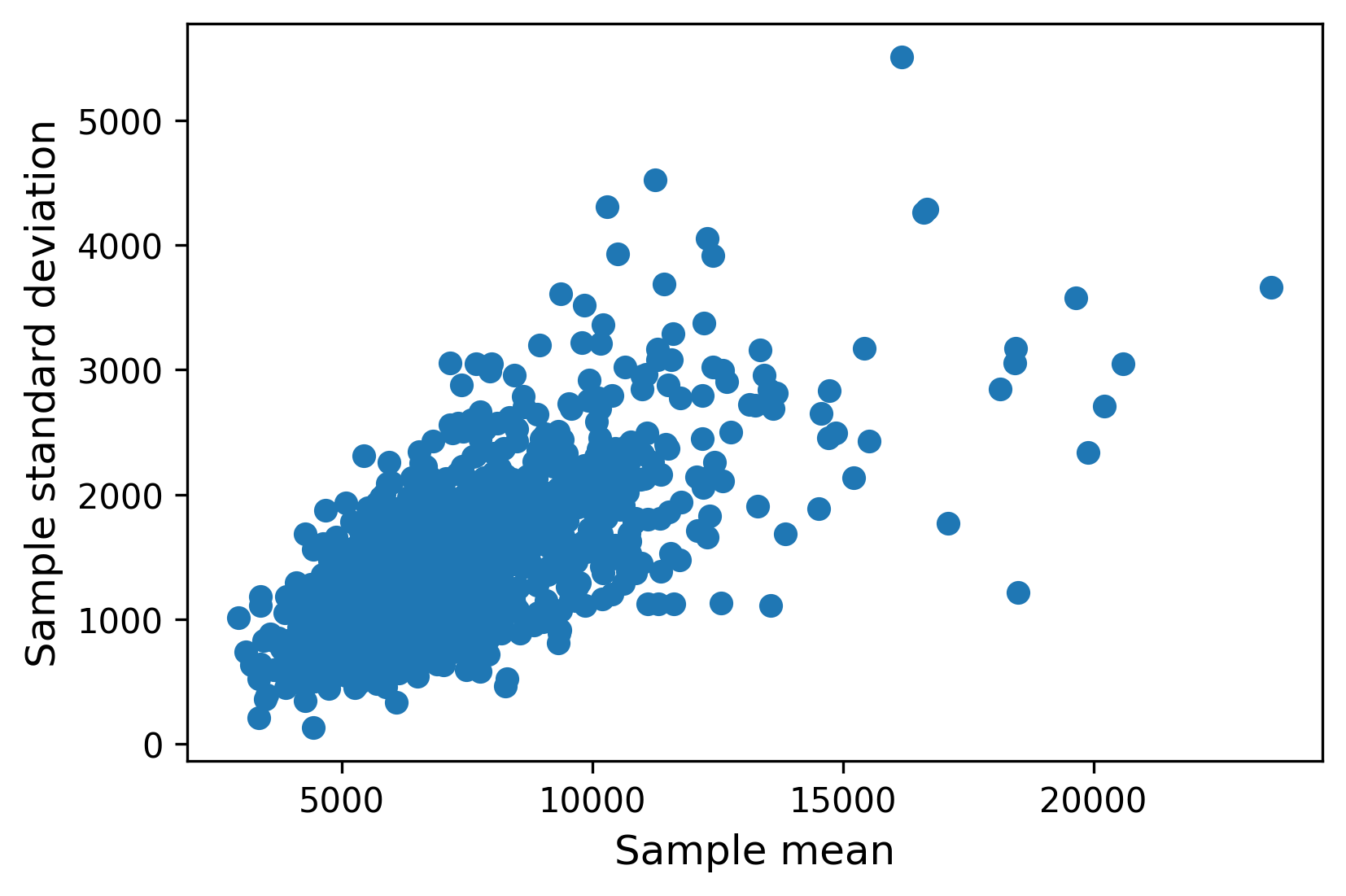}
    \caption{Sample mean and sample standard deviation of demands at store levels}
    \label{fig:sample_mean_stde}
    \parbox[t]{0.95\linewidth}{\footnotesize Notes. The figure presents the pairs of sample mean and sample standard deviation for store-level sales data. Specifically, we randomly select 1000 stores, and sample 10 historical sales data, and compute the sample mean and the sample deviation. }
\end{figure}

{\bf Distance metrics}. Based on the discussion in Section \ref{sec_numerical_synthetic}, the sample mean-based metric is preferred when CVs of random demands are similar even under the newsvendor loss function. However, in practice, the distribution of CVs is not known. One possible approach is to test whether the linear relationship between the sample mean and the sample standard deviation is significant. We present the pairs of the sample mean and sample standard deviation computed based on 10-day sales data in Figure \ref{fig:sample_mean_stde}, and our statistical analysis suggests that the linear relationship between the sample mean and the sample standard deviation is significant at a confidence level of 95\% (see Appendix \ref{appx_Numerical}). This suggests that the cluster-based data pooling approach with the sample mean-based metric may generate a higher benefit of clustering before pooling data for decision-making compared to the sample quantile-based metric. 

{\bf Bisect clustering}. Different from Section \ref{sec_numerical_realdata}, the number of clusters is not known when analyzing with the real data. We adopt the bisect clustering algorithm \citep[e.g.,][]{SteinbachKarypisKumar2000, MurugesanZhang2011, KallusMao2023}, a case of divisive clustering, that identifies a hierarchical cluster structure.  The bisect clustering algorithm is described in Algorithm \ref{alg_BisectCluster}. In particular, a base clustering algorithm, denoted by $\mathrm{CLUST}$, is implemented recursively to group problems into two sub-clusters. For example, the clustering algorithm $\mathrm{CLUST}$  simply groups problems into two sub-clusters based on a boundary of the sub-clusters (e.g., Algorithm \ref{alg_Cluster}), or employs the K-means clustering algorithm with the number of clusters set to 2. Therefore, the resulting cluster structure can be represented by a binary tree. 

As the clustering algorithm proceeds, one may obtain a more sophisticated cluster structure. However, the average number of problems within each cluster would also decrease, which may affect the effectiveness of the data pooling procedure. Therefore, as suggested by Remark 1, we monitor the number of problems within each cluster during the implementation of the bisect clustering algorithm. Specifically, when the number of problems under consideration is below the minimum number of problems, denoted by $k^\ast$, the clustering algorithm terminates segmenting that set of problems into smaller clusters.

\begin{algorithm}[!ht]
  \footnotesize
  \caption{Bisect clustering}
  \label{alg_BisectCluster}
  \begin{algorithmic}[1]
    \Require A set of problems $\mathcal{K}$, Data for clustering, $\{S_k^c, k \in \mathcal{K}\}$, the minimum number of problems for data pooling $k^\ast$, and statistics $\mathcal{M}$.
    \Function{BisectCluster}{$\mathcal{C}$, $k^\ast$, $\mathcal{M}$}

       \State {$S^c \leftarrow \{S_k^c, k \in \mathcal{C}\}$}
       \State $(\hat{\mathcal{C}}_1, \hat{\mathcal{C}}_2) \leftarrow \Call{CLUST}{\mathcal{C}, S^c,\mathcal{M}}$
    
       \If{ $\min\{|\hat{\mathcal{C}}_1|, |\hat{\mathcal{C}}_2|\} < k^\ast$ }
       \State \Return $\{\mathcal{C}\}$
       \Else
       \State ${\mathcal{C}}_1 \leftarrow \Call{BisectCluster}{\hat{\mathcal{C}}_1, k^\ast, \mathcal{M}}$, $\mathcal{C}_2 \leftarrow \Call{BisectCluster}{\hat{\mathcal{C}}_2, k^\ast, \mathcal{M}}$, 
       \State \Return $\mathcal{C}_1 \cup \mathcal{C}_2$
       \EndIf
    \EndFunction
    \Ensure $\Call{BisectCluster}{\mathcal{K}}$
  \end{algorithmic}
\end{algorithm}

\begin{figure}[htbp!]
    \centering
     \includegraphics[width=0.5\linewidth]{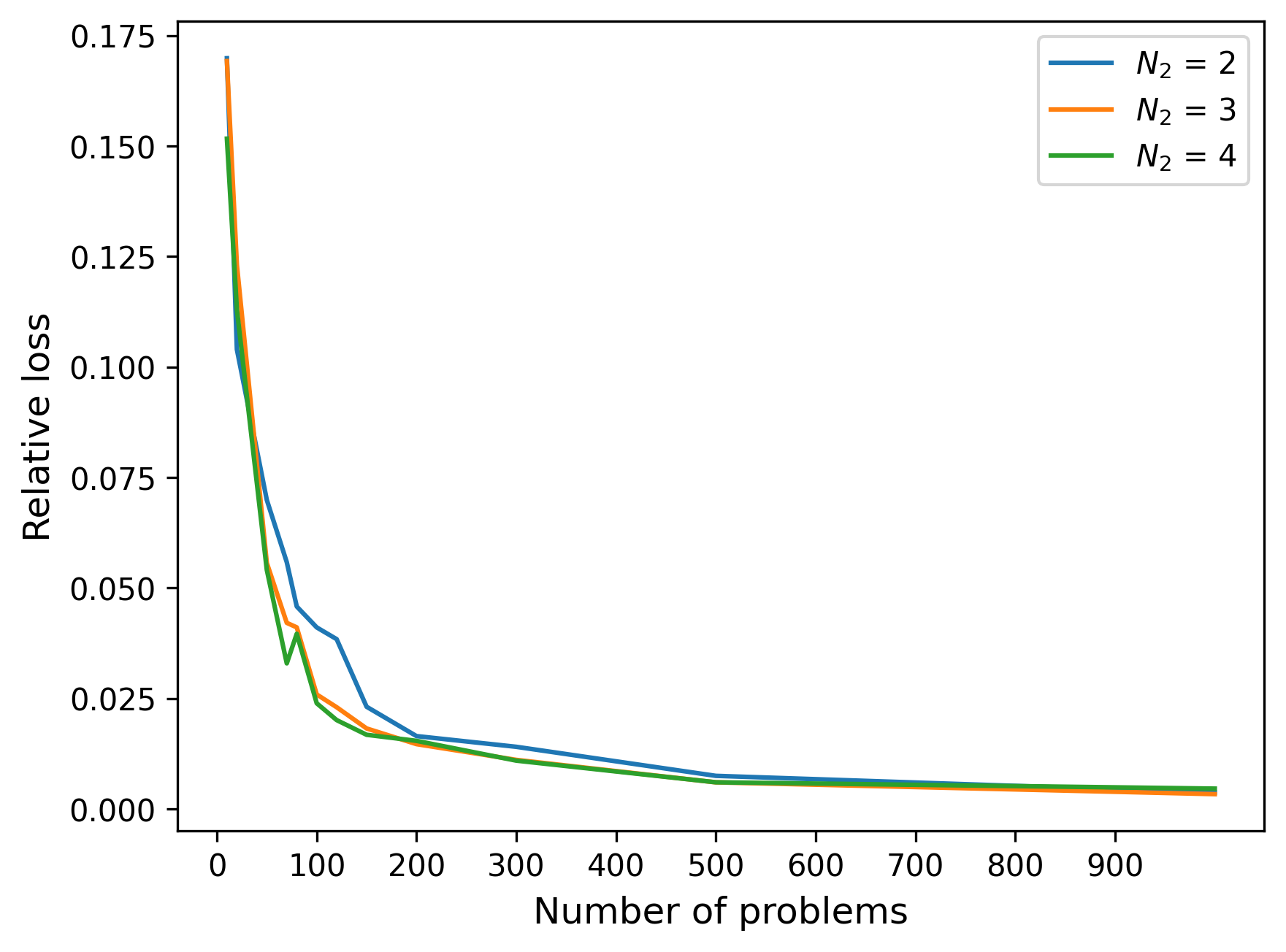}
    \caption{Relative loss of direct data pooling with different numbers of problems}
    \label{fig:ratio_diff_numbers}
    \parbox[t]{0.95\linewidth}{\footnotesize Notes. The figure presents the relative loss of direct data pooling compared with the lower bound cost obtained with shrinkage parameter tuned with the validation data set. The size of the validation set, denoted by $N_2$, is set to 2, 3, and 4 respectively.   }

\end{figure}

{\bf The minimum number of problems for data pooling.} We determine the critical hyperparameter, the minimum number of problems for implementing the data pooling procedure, using cross-validation. For each store $k$, we randomly sample $N_2$ data points from $\hat{N}_k$ historical sales data points as the validation set, while the remaining ones constitute the training set. Given the number of problems $K$, we can apply the direct data pooling approach with the training set and compute the out-of-sample cost using the validation set. To evaluate the performance of the data pooling approach with $K$ problems, we compute a lower bound cost with respect to the validation set. Specifically, one can obtain the oracle's shrinkage parameter $\alpha_{K}^{\ast}$ using Equation \eqref{oracle_alpha} with the validation set. Intuitively, the out-of-sample cost associated with the shrinkage parameter $\alpha_{K}^{\ast}$ provides a lower bound for the cost obtained with the decisions made by the data-driven data pooling approach. Thus, we can compute the gap between the cost with the data-driven data pooling approach and that of the lower bound benchmark as
\begin{equation*}
     \text{Relative Loss } (\%) = \frac{Z_p(\hat{\alpha}_K) - Z_p(\alpha_K^\ast)}{Z_p(\alpha_K^\ast)}  \times 100\%.
\end{equation*}
As depicted in Figure \ref{fig:ratio_diff_numbers}, for any given size of validate set, denoted as $N_2$, the gap between the cost obtained with the data-driven shrinkage parameter and the lower bound cost decreases steeply when the number of problems is relatively small, e.g., $K \leq 100$. This implies that increasing the number of problems in this range can significantly improve the performance of the data pooling approach. However, when the number of problems is greater than 150, the curve of the relative loss becomes flat. In other words, including more problems yields diminishing marginal benefits for direct data pooling. This phenomenon is also observed in \cite{GuptaKallus2022}. Therefore, we choose $k^\ast=150$ as the minimum number of problems when implementing the bisect clustering algorithm.   

\begin{figure}[!ht]
\centering
\subfigure[$s = 0.95$]{
\begin{minipage}[t]{0.5\linewidth}
\centering
\includegraphics[width=3.2in,height=2.8in]{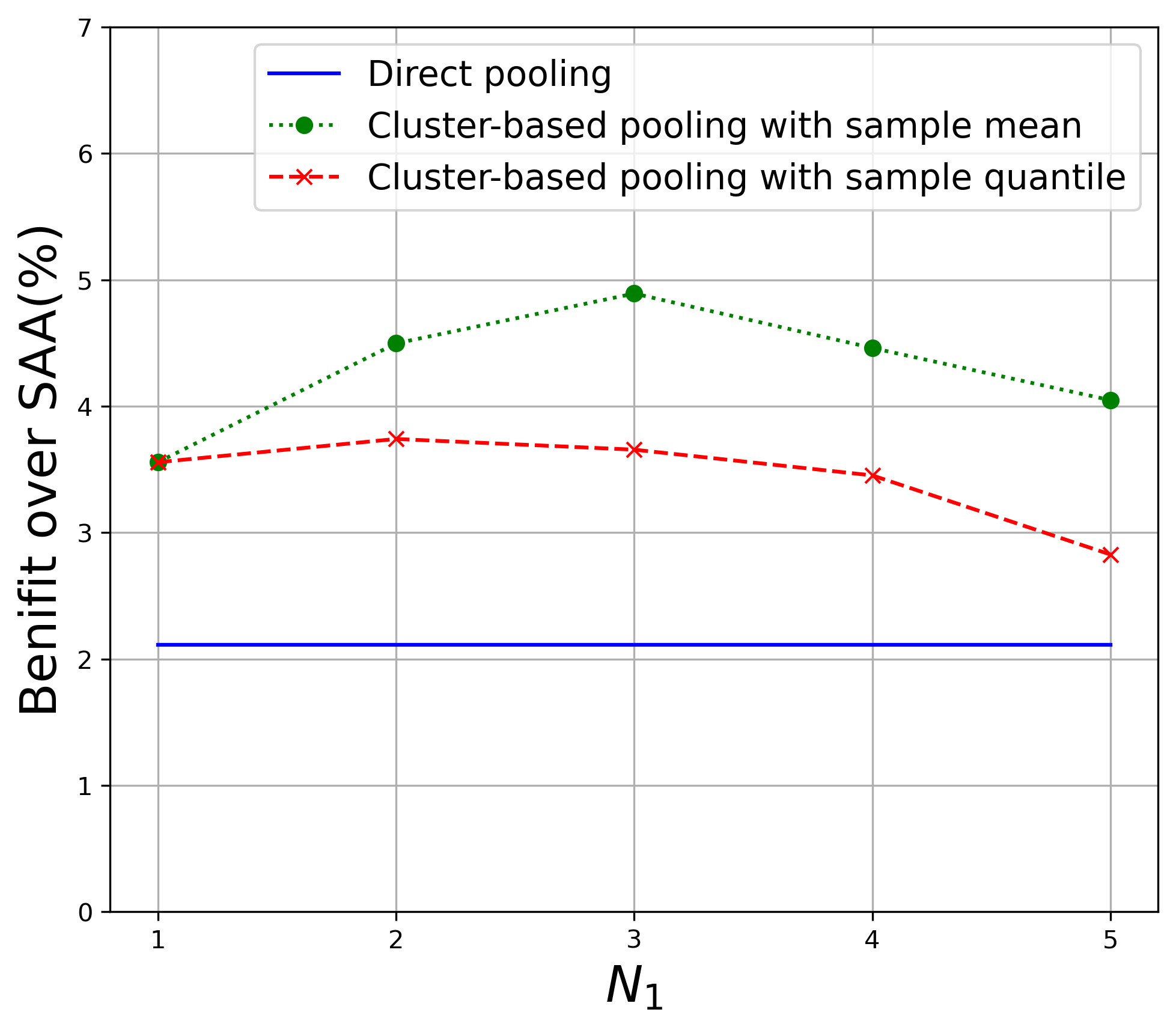}
\end{minipage}
}%
\subfigure[$s = 0.97$]{
\begin{minipage}[t]{0.5\linewidth}
\centering
\includegraphics[width=3.2in,height=2.8in]{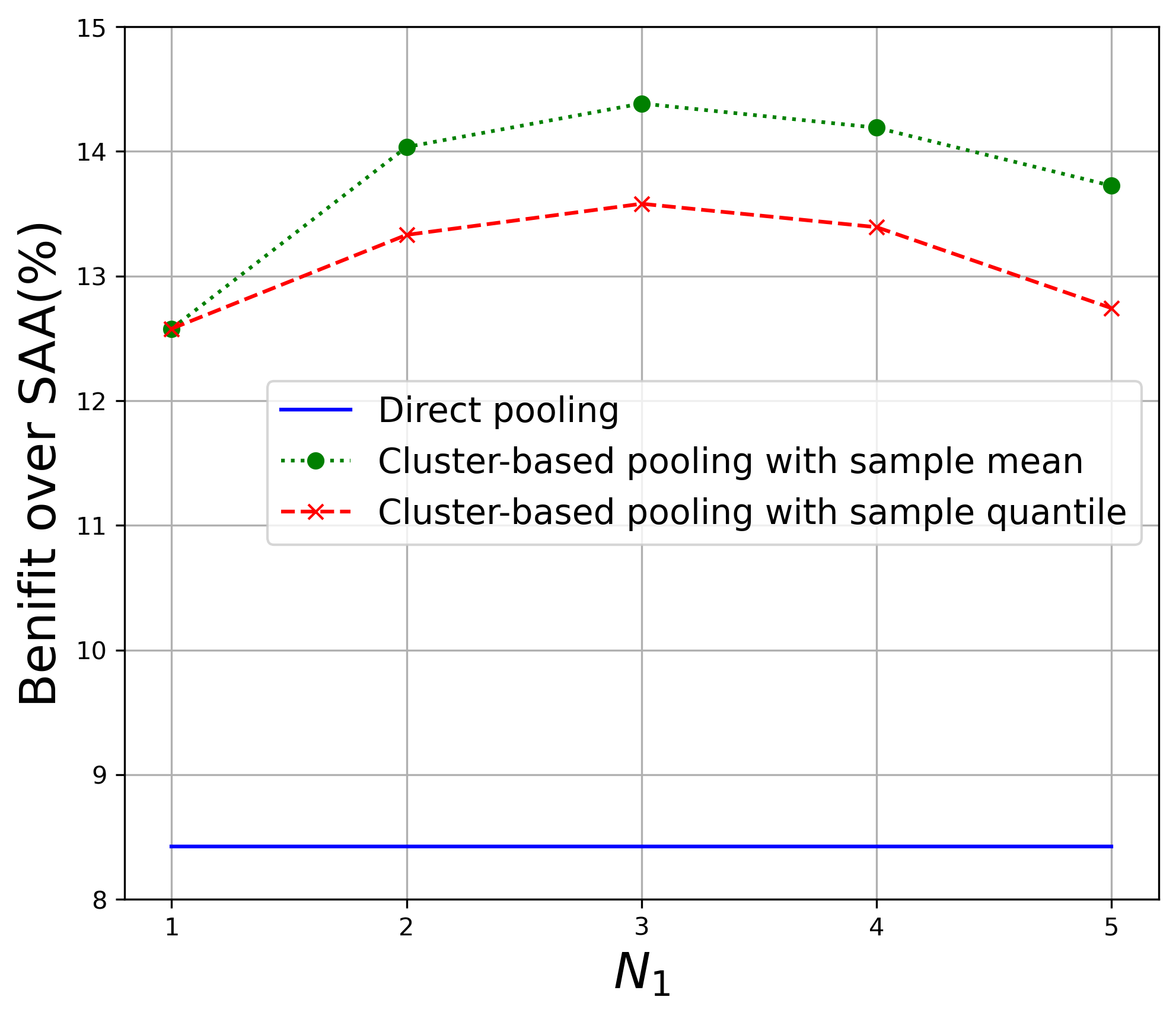}
\end{minipage}%
}%
\centering
\caption{Performance of data pooling approaches measured by $(Z_\mathrm{SAA}-Z_p)/Z_\mathrm{SAA} \times 100\%$.}
\label{fig_real_data_driven}
\parbox[t]{0.95\linewidth}{\footnotesize Notes. The above panels demonstrate the relative advantages of data pooling approaches over SAA with real data. 
The number of problems is $K=1105$. Each reported relative advantage is the average of 100 instances. The newsvendor critical ratio is set to 0.95 and 0.97, respectively.}
\end{figure}

{\bf Benefit of clustering}. Given the tuned minimum number of problems for each cluster, we illustrate the benefits of the data pooling approaches over the SAA approach with real data in Figure \ref{fig_real_data_driven}. With a limited number of observations, e.g., $N=10$, both direct data pooling and cluster-based data pooling approaches demonstrate significant advantages over the SAA approach, of which performance is guaranteed only with a sufficient amount of data \citep{KimPasupathyHenderson2015}. In addition, the advantages of the data pooling approaches are more evident when the newsvendor ratio $s$ is larger. The intuition is that when the critical ratio $s$ is relatively high, the SAA approach requires a greater amount of data to accurately estimate the critical quantile especially to achieve a satisfactory performance, a phenomenon which is also observed in \cite{BesbesMouchtaki2023}. 
More importantly, the cluster-based data pooling approach outperforms the direct data pooling approach with different newsvendor critical ratios and numbers of data points for clustering. This demonstrates the uniform benefit of clustering before pooling the data for decision-making. Finally, as we discussed previously, the CVs of the random demands in the real data turn out to be similar. Thus, combined with the observation in Section \ref{sec_numerical_synthetic}, cluster-based data pooling with sample mean-based metric outperforms the one using critical quantile-based metric. In other words, the choice of distance metrics impacts the benefit of clustering for data pooling approaches. A careful investigation of the distributional pattern across all problems is desired to guide the effective implementation of the cluster-based data pooling approach.

\begin{figure}[htbp!]
    \centering
     \includegraphics[width=0.55\linewidth]{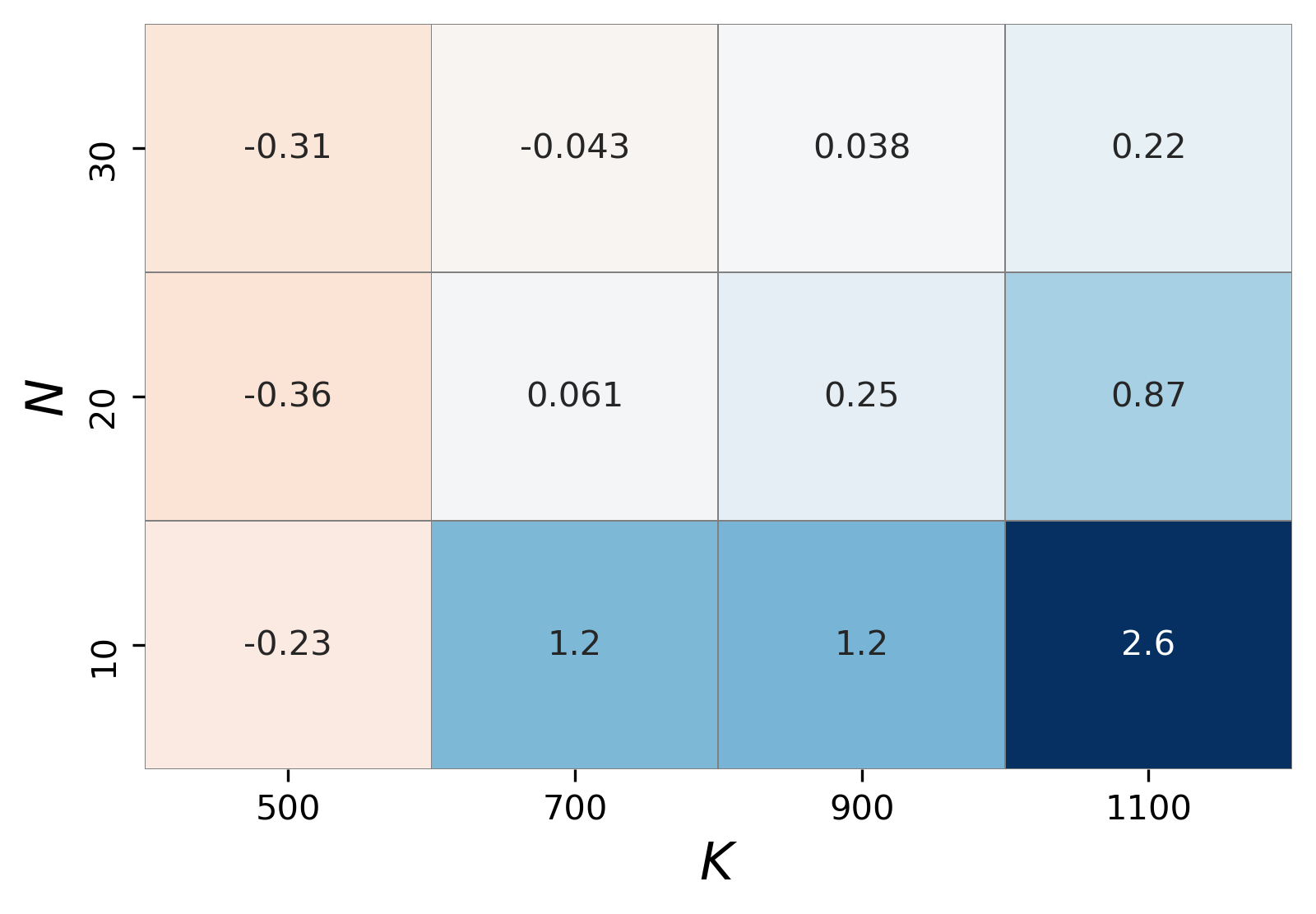}
    \caption{The relative advantages of cluster-based pooling over direct pooling with real data}
    \label{fig:vary_K_N}
    \parbox[t]{0.95\linewidth}{\footnotesize Notes. The above heatmap demonstrates the relative advantages of the cluster-based pooling approach over direct pooling with real data. The total number of problems $K$ is set to 500, 700, 900, and 1100. The number of data points $N$ is set to 10, 20, 30. The number of data points for clustering is $N_1=2$. Each reported relative advantage is the average of 100 instances. The newsvendor critical ratio $s$ is set to 0.95. The clustering distance metric is the sample mean.}
\end{figure}

Finally, we compare the performance of the cluster-based data pooling with that of the direct data pooling under different numbers of problems $K$ and numbers of observations $N$. Specifically, $K$ varies from 500 to 1100 and $N$ varies from 10 to 30. Given a pair of $K$ and $N$, we randomly sample $K$ stores from the data set and sample $N$ data points for each store. The number of data points allocated for clustering $N_1=2$, and the sample mean is adopted as the distance metric for the cluster-based data pooling approach.\footnote{The results are robust when the choice of $N_1$ is within 30\% of N .} We report the average difference between the costs associated with the cluster-based and direct data pooling approaches from 100 randomly generated instances. As demonstrated in Figure \ref{fig:vary_K_N},  when the number of problems is relatively large and the sample size is relatively small, the cluster-based data pooling approach outperforms direct data pooling. However, when the number of problems is relatively small, cluster-based data pooling does not provide additional benefits over the direct one. Because the effective implementation of data pooling requires involving a certain amount of problems. With a relatively small number of problems, after conducting clustering analysis, the number of problems within each cluster is limited, and thus, it may affect the effectiveness of data pooling. Furthermore, when the number of observations increases, as suggested by Equations \eqref{eqn_ShrunkenSAAMSE} and \eqref{eqn_ClusterShrunkenSAAMSE}, the anchor distribution takes a less critical role in shaping the Shrunken-SAA solutions, and the benefit of clustering, which hinges on the flexible choices of anchors, becomes marginal. In a nutshell, leveraging the cluster structure among problems is more beneficial under the small-data large-scale regime.


\section{Conclusion}\label{sec_conclusion}
We propose a novel cluster-based data pooling approach that can flexibly exploit the cluster structure of problems when implementing the Shrunken-SAA algorithm within each cluster for decision-making with limited data. This approach not only carefully determines the extent to which the limited data of each problem is pooled with the data from other problems, but also leverages the similarity structure of problems to enhance the efficacy of data pooling. We theoretically characterize the conditions under which harnessing cluster structure is beneficial for data pooling under MSE. With the given cluster structure, cluster-based data pooling consistently generates additional benefits over the direct one that neglects such structure. When the cluster structure is unknown, clustering analysis, which may sacrifice a few data points, is still beneficial, as long as the distance between clusters is substantial. We further provide a theoretical guarantee on the performance of cluster-based data pooling under the general cost function. Empirical evidence highlights the impacts of the choice of the distance metrics measuring the problem instances and validates the superior performance of the proposed approach over the existing ones, especially in the large-scale small-data regime. 

In this study, we utilize the observed data for clustering analysis. One may utilize the covariate information associated with each problem to identify the cluster structure, which may further enhance the performance of data aggregation approaches. In addition, we study an offline learning setting where the data are given in advance. One may also consider an online learning setting where the decision maker dynamically categorizes the problems based on the collected data and determines the operational decisions for each problem.  

\clearpage
\clearpage
\bibliographystyle{informs2014} 
\bibliography{main} 

\ECSwitch


\ECHead{Appendices}
\section{Proofs of Formal Results} \label{appx_A}

The proof of Proposition \ref{prop_DataDrivenPooling} requires the following lemmas.

\begin{lemma}[Hoeffding's inequality \citep{Hoeffding1994}]\label{lma_hoffding_inequality}
Let $Y_1, ..., Y_n$ be independent random variables such that $a_i\le Y_i\le b_i$. Let $W_n = \sum_{i=1}^n Y_i$. Then for any $t>0$, we have
\begin{equation}
    \notag
    P\Big(\Big|W_n/n - \mathbb{E}[W_n]/n\Big| \ge t\Big) \le 2\exp\Big(-2n^2t^2/\Big(\sum_{i = 1}^{n}(b_i-a_i)^2\Big)\Big).
\end{equation}
\end{lemma}

\begin{lemma}[Concentration of sub-Gaussian random variables \citep{Wainwright2019}]\label{lma_SubGaussinConcentration}
Let $Y_1, ..., Y_n$ be independent sub-Gaussian random variables such that $\mathbb{E}[\exp(s(Y_1-\mathbb{E}[Y_1]))] \le \exp(cs^2/2)$ for a constant $c>0$. Let $W_n = \sum_{i=1}^n Y_i$. Then for any $t>0$, we have
\begin{equation}
    \notag
    P\Big(\Big|W_n/n - \mathbb{E}[W_n]/n\Big| \ge t\Big) \le 2\exp(-nt^2/(2c)).
\end{equation}
\end{lemma}

\begin{lemma}[Concentration of sub-Exponential random variables \citep{Wainwright2019}]\label{lma_SubExponentialConcentration}
 Let $Y_1, ..., Y_n$ be independent sub-exponential random variables with parameters $(v_i,b_i)$. Let $W_n = \sum_{i=1}^n Y_i$, we have,
\begin{align*}
    P\Big(\Big|W_n/n-\mathbb{E}[W_n]/n\Big| \ge t\Big) \le \left\{
    \begin{aligned}
     &2\exp(-nt^2/(2v^2), && \text{for}  \quad 0 \le t \le \frac{v^2}{b}, \\
     &2\exp(-nt^2/(2b)), && \text{for}   \quad t > \frac{v^2}{b},
    \end{aligned}
    \right.
\end{align*}
where $b = \max_{i} b_i,\mbox{ }v^2 = \sum_i \frac{v_i^2}{n}$.
\end{lemma}

{\bf \noindent Proof of Proposition \ref{prop_DataDrivenPooling}.}
We first prove Proposition \ref{prop_DataDrivenPooling} when Assumption \ref{asmp_1} (i) holds. 

For the part (i) of Proposition \ref{prop_DataDrivenPooling}, by Hoeffding's inequality, it is easy to verify that $\hat{\mu}_0 \to_p \mu_0^{\mathrm{AP}}$ as $|C| \to \infty$. In order to show that $\hat{\alpha} \to_p \alpha^\mathrm{AP}$ as $|C|\to\infty$, we will prove that, as $|C| \to \infty$, 
\begin{align}
    \frac{1}{|C|}\sum_{k \in C}\frac{1}{N-1}\sum_{j = 1}^{N}(\hat{\xi}_{kj}-\hat{\mu}_k)^2 &\to_p \frac{1}{|C|}\sum_{k \in C}\sigma_k^2,\label{eqn_NoiseVarianceConverge}\\ \frac{1}{|C|}\sum_{k \in C}(\hat{\mu}_k - \hat{\mu}_0)^2 &\to_p  \frac{1}{|C|}\sum_{k \in C} \mathbb{E}[(\hat{\mu}_k - \mu_0^{\mathrm{AP}})^2].\label{eqn_MeanVarianceConverge}
\end{align}
Suppose Equations \eqref{eqn_NoiseVarianceConverge} and \eqref{eqn_MeanVarianceConverge} holds. In addition, by definition, we have $\mathbb{E}[(\mu_0^{\mathrm{AP}} - \hat{\mu}_k)^2] = \frac{\sigma_k^2}{N} + (\mu_0^{\mathrm{AP}}-\mu_k)^2$. By Slutsky's Theorem and the fact that convergence in probability is equivalent to convergence in distribution when the limit is a constant, we can conclude that $\hat\alpha \to_p \alpha^\mathrm{AP}$. 

To show Equation \eqref{eqn_NoiseVarianceConverge}, we define random variables $Y_k = \frac{1}{N-1}\sum_{j =1}^{N}(\hat{\xi}_{kj}-\hat{\mu}_k)^2$, $k=1,2,\ldots,K$. By definition, we have, $\mathbb{E}[Y_k] = \sigma_k^2$. By Assumption \ref{asmp_1} (i), it is easy to verify that $0\le Y_k \le 8a^2_\mathrm{max}$. By Hoeffding's inequality, for any given $t > 0$, we have,
\begin{equation}
    \notag
    P\Bigg(\Bigg|\frac{1}{|C|}\sum_{k \in C} Y_k -\frac{1}{|C|}\sum_{k \in C}\sigma_k^2  \Bigg| >t  \Bigg) \le 2 \exp\Bigg(-\frac{|C|t^2}{32a^4_\mathrm{max}}\Bigg) \to 0 \quad \text{as}\quad |C| \to \infty.
\end{equation}
Therefore, as $|C| \to \infty$,
\begin{equation}
    \frac{1}{|C|}\sum_{k \in C} Y_k = \frac{1}{|C|}\sum_{k \in C}\frac{1}{N-1}\sum_{j = 1}^{N}(\hat{\xi}_{kj}-\hat{\mu}_k)^2 \to_p \frac{1}{|C|}\sum_{k \in C}\sigma_k^2.
\end{equation}
To show Equation \eqref{eqn_MeanVarianceConverge}, we  decompose the difference between $\frac{1}{|C|}\sum_{k \in C}(\hat{\mu}_k - \hat{\mu}_0)^2$  and $\frac{1}{|C|}\sum_{k \in C}\mathbb{E}[(\hat{\mu}_k - \mu_0^\mathrm{AP})^2]$ into two parts. Specifically,
\begin{align}
     &\bigg | \frac{1}{|C|}\sum_{k \in C}\Big((\hat{\mu}_k - \hat{\mu}_0)^2-\mathbb{E}[(\mu_0^\mathrm{AP} - \hat{\mu}_k)^2] \Big) \bigg|\label{eqn_VarianceMeanDecompose_1}\\
     = &\bigg | \frac{1}{|C|}\sum_{k \in C}\Big((\hat{\mu}_k - \hat{\mu}_0)^2-(\hat{\mu}_k - \mu_0^{\mathrm{AP}})^2 + (\hat{\mu}_k - \mu_0^{\mathrm{AP}})^2-\mathbb{E}[(\mu_0^{\mathrm{AP}} - \hat{\mu}_k)^2] \Big) \bigg| \label{eqn_VarianceMeanDecompose_2}\\
     \le&\underbrace{\bigg|\frac{1}{|C|}\sum_{k \in C}\Big((\hat{\mu}_k - \hat{\mu}_0)^2-(\hat{\mu}_k - \mu_0^{\mathrm{AP}})^2 \Big) \Bigg|}_{\mathrm{A}} +\underbrace{\Bigg| \frac{1}{|C|}\sum_{k \in C}\Big((\hat{\mu}_k - \mu_0^{\mathrm{AP}})^2-\mathbb{E}[(\mu_0^{\mathrm{AP}} - \hat{\mu}_k)^2] \Big) \bigg|}_{\mathrm{B}}.\label{eqn_VarianceMeanDecompose_3}
\end{align}
Next, we prove the bounds for Parts A and B, respectively. 

For Part A, 
\begin{equation}\label{eqn_PartAConverge}
\bigg|\frac{1}{|C|}\sum_{k \in C}\big((\hat{\mu}_k - \hat{\mu}_0)^2-(\hat{\mu}_k - \mu_0^{\mathrm{AP}})^2 \big) \bigg| \le \frac{1}{|C|}\sum_{k \in C} \Big|\hat{\mu}_0 -\mu_0^{\mathrm{AP}} \Big|\cdot \Big|2\hat{\mu}_k - \hat{\mu}_0 -\mu_0^{\mathrm{AP}} \Big| \le 4a_\mathrm{max} \Big|\hat{\mu}_0 -\mu_0^{\mathrm{AP}} \Big|.
\end{equation}
Since $|\hat{\mu}_0 - \mu_0^{\mathrm{AP}}| \to_p 0 $ as $|C| \to \infty$, we have, $\big|\frac{1}{|C|}\sum_{k \in C}\big((\hat{\mu}_k - \hat{\mu}_0)^2-(\hat{\mu}_k - \mu_0^{\mathrm{AP}})^2 \big) \big| \to_p 0$.

For Part B, we define random variables $\tilde{Y}_k = (\hat{\mu}_k -\mu_0^{\mathrm{AP}})^2$, $k=1,2,\ldots,K$. By Assumption \ref{asmp_1} (i), we have, $0 \leq \tilde{Y}_k \leq 4a_\mathrm{max}^2$. By Hoeffding's inequality in Lemma \ref{lma_hoffding_inequality},
\begin{equation}
    P\Bigg( \Bigg | \frac{1}{|C|}\sum_{k \in C}\Big(\tilde{Y}_k - \mathbb{E}[\tilde{Y}_k] \Big) \Bigg| > t \Bigg) \le 2\exp\Big(-\frac{|C|t^2}{8a^2_\mathrm{max}}\Big) \to 0 \quad \text{as} \quad |C|\to \infty.
\end{equation}
Thus, we have, $\Big| \frac{1}{|C|}\sum_{k \in C}\big((\hat{\mu}_k - \mu_0^{\mathrm{AP}})^2-\mathbb{E}[(\hat{\mu}_k - \mu_0^{\mathrm{AP}})^2] \big) \Big| \to_p 0$. Combined with Equation \eqref{eqn_PartAConverge}, we can conclude that Equation \eqref{eqn_MeanVarianceConverge} holds. This completes the proof of Part (i) when Assumption \ref{asmp_1} (i) holds.

For Part (ii) of Proposition \ref{prop_DataDrivenPooling}, we observe that, 
\begin{align*}
     &\frac{1}{|C|}\sum_{k \in C}  (x_k(\hat{\alpha},\hat{\mu}_0,\hat{\mu}_k)-\mu_k)^2  - \frac{1}{|C|}\sum_{k \in C}  (x_k(\alpha^\mathrm{AP},\mu_0^{\mathrm{AP}},\hat{\mu}_k)-\mu_k)^2 \\
      \le & \frac{1}{|C|}\sum_{k \in C}  \Big|(x_k(\hat{\alpha},\hat{\mu}_0,\hat{\mu}_k)-\mu_k)^2 -  (x_k(\alpha^\mathrm{AP},\mu_0^{\mathrm{AP}},\hat{\mu}_k)-\mu_k)^2  \Big|\\
       = & \frac{1}{|C|}\sum_{k \in C}  \Big|\Big(\hat{\mu}_k-\mu_k+\frac{\hat{\alpha}}{N+\hat{\alpha}}(\hat{\mu}_0-\hat{\mu}_k)\Big)^2 - \Big(\hat{\mu}_k-\mu_k+\frac{\alpha^\mathrm{AP}}{N+\alpha^\mathrm{AP}}(\mu_0^{\mathrm{AP}}-\hat{\mu}_k)\Big)^2\Big| \\
       =& \frac{1}{|C|}\sum_{k \in C}  \Big|\frac{\hat{\alpha}}{N+\hat{\alpha}}(\hat{\mu}_0-\hat{\mu}_k) - \frac{\alpha^\mathrm{AP}}{N+\alpha^\mathrm{AP}}(\mu_0^{\mathrm{AP}}-\hat{\mu}_k) \Big| \cdot \Big|2(\hat{\mu}_k-\mu_k)+\frac{\hat{\alpha}}{N+\hat{\alpha}}(\hat{\mu}_0-\hat{\mu}_k)+ \frac{\alpha^\mathrm{AP}}{N+\alpha^\mathrm{AP}}(\mu_0^{\mathrm{AP}}-\hat{\mu}_k) \Big| \\
     \le &   \frac{8a_\mathrm{max}}{|C|}\sum_{k \in C} \Big|\frac{\hat{\alpha}}{N+\alpha_{\hat{\mu}_o}}\hat{\mu}_0 - \frac{\alpha^\mathrm{AP}}{N+\alpha^\mathrm{AP}}\mu_0^{\mathrm{AP}}+\hat{\mu}_k(\frac{\alpha^\mathrm{AP}}{N+\alpha^\mathrm{AP}}-\frac{\hat{\alpha}}{N+\hat{\alpha}}) \Big| \\
      \le &  8a_\mathrm{max} \Big|\frac{\hat{\alpha}}{N+\hat{\alpha}}\hat{\mu}_0 - \frac{\alpha^\mathrm{AP}}{N+\alpha^\mathrm{AP}}\mu_0^{\mathrm{AP}} \Big| + 8a^2_\mathrm{max} \Big| \frac{\alpha^\mathrm{AP}}{N+\alpha^\mathrm{AP}}-\frac{\hat{\alpha}}{N+\hat{\alpha}} \Big|.
\end{align*}
Since $|\xi_k| \le a_\mathrm{max}$ by assumption, we have, $\hat{\mu}_k, \hat{\mu}_0, \mu_0  \le a_\mathrm{max}$. In addition, $\frac{\alpha^\mathrm{AP}}{N+\alpha^\mathrm{AP}} < 1 $ and $\frac{\hat{\alpha}}{N+\hat{\alpha}}<1$ by definition. Therefore, the second inequality holds. The third inequality holds due to the triangle inequality. 

In addition,
since  $\hat{\alpha} \to_p \alpha^\mathrm{AP}$, by Continuous Mapping Theorem, we have $\frac{\hat{\alpha}}{N+\hat{\alpha}} \to_p \frac{\alpha^\mathrm{AP}}{N+\alpha^\mathrm{AP}}$. Combined with the fact that $\hat{\mu}_0 \to_p \mu_0^{\mathrm{AP}}$, by the Slutsky’s Theorem, we have,  
\begin{equation}
     8a_\mathrm{max} \Big|\frac{\hat{\alpha}}{N+\hat{\alpha}}\hat{\mu}_0 - \frac{\alpha^\mathrm{AP}}{N+\alpha^\mathrm{AP}}\mu_0^{\mathrm{AP}} \Big| + 8a^2_\mathrm{max} \Big| \frac{\alpha^\mathrm{AP}}{N+\alpha^\mathrm{AP}}-\frac{\hat{\alpha}}{N+\hat{\alpha}} \Big| \to_p 0.
\end{equation}
Therefore, we can conclude that,
\begin{equation}
\frac{1}{|C|}\sum_{k \in C}  (x_k(\hat{\alpha},\hat{\mu}_0,\hat{\mu}_k)-\mu_k)^2  \to_p \frac{1}{|C|}\sum_{k \in C}  (x_k(\alpha^\mathrm{AP},\mu_0^{\mathrm{AP}},\hat{\mu}_k)-\mu_k)^2.
\end{equation}
This completes the proof for Proposition \ref{prop_DataDrivenPooling} when Assumption \ref{asmp_1} (i) holds.

Next, we prove Proposition \ref{prop_DataDrivenPooling} when Assumption \ref{asmp_1} (ii) holds. Since $\xi_{ki}$ is normally distributed, $\hat{\mu}_k$ follows a sub-Gaussian distribution \citep[see.,  Definition 2.1 in][]{Wainwright2019}. By Lemma \ref{lma_SubGaussinConcentration}, we have, $\hat{\mu}_0 \to_p \mu_0^{\mathrm{AP}}$.

In order to prove $\hat{\alpha} \to_p \alpha^\mathrm{AP}$, we need to show Equations \eqref{eqn_NoiseVarianceConverge} and \eqref{eqn_MeanVarianceConverge} when Assumption \ref{asmp_1} (ii) holds. Define random variables $U_k = \frac{1}{N-1}\sum_{j =1}^{N}(\hat{\xi}_{kj}-\hat{\mu}_k)^2$, $k = 1,2,\ldots,K$. Since $\hat{\xi}_{kj}$ follows Gaussian distribution, we have, $\frac{(N-1)U_k}{\sigma_k^2} \sim \chi^2(N-1)$. It is known that Chi-squared distribution is Sub-Exponential distribution \citep[see, Example 2.4 in][]{Wainwright2019}. By Lemma \ref{lma_SubExponentialConcentration}, we have, for any $t>0$,
\begin{align*}
    P\Big(\Big|\frac{1}{|C|}\sum_{k\in C}\Big( \frac{(N-1)U_k}{\sigma_k^2}-(N-1)\Big)\Big| \ge t\Big) \le \left\{
    \begin{aligned}
     &2\exp\Big(-\frac{|C|t^2}{8(N-1)}\Big),  &&\text{for} \quad 0 \le t \le N-1,\\
     &2\exp\Big(-\frac{|C|t^2}{8}\Big),  &&\text{for} \quad t > N-1.
    \end{aligned}
    \right.
\end{align*}
Based on the above relation, as $|C|\to\infty$, we have, $\frac{1}{|C|}\sum_{k\in C}\frac{1}{N-1}\sum_{j = 1}^{N}(\hat{\xi}_{kj}-\hat{\mu}_k)^2 \to_p \frac{1}{|C|}\sum_{k\in C} \sigma_k^2$. That is, Equation \eqref{eqn_NoiseVarianceConverge} holds.

Similar to the proof for Assumption \ref{asmp_1} (i), based on Equations \eqref{eqn_VarianceMeanDecompose_1} - \eqref{eqn_VarianceMeanDecompose_3}, we have, 
\begin{align}
\begin{split}
&\bigg | \frac{1}{|C|}\sum_{k \in C}\Big((\hat{\mu}_k - \hat{\mu}_0)^2-\mathbb{E}[(\mu_0^{\mathrm{AP}} - \hat{\mu}_k)^2] \Big) \bigg|\\
&\le \underbrace{\bigg|\frac{1}{|C|}\sum_{k \in C}\Big((\hat{\mu}_k - \hat{\mu}_0)^2-(\hat{\mu}_k - \mu_0^{\mathrm{AP}})^2 \Big) \Bigg|}_{\mathrm{A}} +\underbrace{\Bigg| \frac{1}{|C|}\sum_{k \in C}\Big((\hat{\mu}_k - \mu_0^{\mathrm{AP}})^2-\mathbb{E}[(\mu_0^{\mathrm{AP}} - \hat{\mu}_k)^2] \Big) \bigg|}_{\mathrm{B}}. \label{eqn_VarianceMeanDecompose_4}
\end{split}
\end{align}
For Part A of the right-hand side of Equation \eqref{eqn_VarianceMeanDecompose_4}, by definitions of $\hat{\mu}_0$ and $\mu_0^{\mathrm{AP}}$, we have, 
\begin{align*}
     \bigg | \frac{1}{|C|}\sum_{k \in C}\Big((\hat{\mu}_k - \hat{\mu}_0)^2-(\hat{\mu}_k - \mu_0^{\mathrm{AP}})^2 \Big) \bigg| 
     = \Big| (\hat{\mu}_0-\mu_0^{\mathrm{AP}})^2\bigg|.
\end{align*}
Based on the above relation, since $\hat{\mu}_0 \to_p \mu_0^{\mathrm{AP}}$ as $|C|\to \infty$, we have, $\Big| (\hat{\mu}_0-\mu_0^{\mathrm{AP}})^2\Big| \to_p 0$, and consequently, $ \big| \frac{1}{|C|}\sum_{k \in C}\Big((\hat{\mu}_k - \hat{\mu}_0)^2-(\hat{\mu}_k - \mu_0^{\mathrm{AP}})^2 \Big) \big| \to_p 0$.
For Part B of the right-hand side of Equation \eqref{eqn_VarianceMeanDecompose_4}, we have,
\begin{align*}
    &\bigg| \frac{1}{|C|}\sum_{k \in C}\Big((\hat{\mu}_k - \mu_0^{\mathrm{AP}})^2- \mathbb{E}[(\hat{\mu}_k - \mu_0^{\mathrm{AP}})^2 \Big)\bigg|\\
    = &\bigg | \frac{1}{|C|}\sum_{k \in C}\Big((\hat{\mu}_k - \mu_k +\mu_k- \mu_0^{\mathrm{AP}})^2-\mathbb{E}[(\hat{\mu}_k - \mu_k +\mu_k-\mu_0^{\mathrm{AP}})^2] \Big)\Bigg| \\
    = & \bigg | \frac{1}{|C|}\sum_{k \in C}\Big((\hat{\mu}_k-\mu_k)^2-\mathbb{E}[(\hat{\mu}_k-\mu_k)^2]\Big)+\frac{1}{|C|}\sum_{k \in C}2(\mu_k-\mu_0^{\mathrm{AP}})(\hat{\mu}_k-\mu_k)  \bigg|  \\
     \le & \bigg |  \frac{1}{|C|}\sum_{k \in C}\Big((\hat{\mu}_k-\mu_k)^2-\mathbb{E}[(\hat{\mu}_k-\mu_k)^2]\Big)\bigg| + 2b_\mathrm{max}\big|\hat{\mu}_0-\mu_0^{\mathrm{AP}}\big| +  \bigg|\frac{1}{|C|}\sum_{k \in C} 2\mu_k(\hat{\mu}_k-\mu_k) \bigg|\\
     \leq & \frac{\sigma_\mathrm{max}^2}{N}\bigg | \frac{1}{|C|}\sum_{k \in C}\bigg(\Big(\frac{\sqrt{N}(\hat{\mu}_k-\mu_k)}{\sigma_k}\Big)^2-\mathbb{E}\Big[\Big(\frac{\sqrt{N}(\hat{\mu}_k-\mu_k)}{\sigma_k}\Big)^2\Big]\bigg) \bigg| + 2b_\mathrm{max}\big|\hat{\mu}_0-\mu_0^{\mathrm{AP}}\big| +  \bigg|\frac{1}{|C|}\sum_{k \in C} 2\mu_k(\hat{\mu}_k-\mu_k) \bigg|.
\end{align*}
The first inequality holds due to the assumption that $|\mu_k| \le b_\mathrm{max}$. The second inequality holds due to the assumption that $\sigma_k \le \sigma_\mathrm{max}$. We observe that $\Big(\frac{\sqrt{N}(\hat{\mu}_k-\mu_k)}{\sigma_k}\Big)^2 \sim \chi^2(1)$, which follows a sub-Exponential distribution \citep[See, Example 2.4 in][]{Wainwright2019}. By Lemma \ref{lma_SubExponentialConcentration}, we have, $\big | \frac{1}{|C|}\sum_{k \in C}\big(\big(\frac{\sqrt{N}(\hat{\mu}_k-\mu_k)}{\sigma_k}\big)^2-\mathbb{E}\big[\big(\frac{\sqrt{N}(\hat{\mu}_k-\mu_k)}{\sigma_k}\big)^2\big]\big) \big| \to_p 0$, and thus, $\big| \frac{1}{|C|}\sum_{k \in C}\big((\hat{\mu}_k-\mu_k)^2-\mathbb{E}[(\hat{\mu}_k-\mu_k)^2]\big) \big| \to_p 0$, as $|C|\to \infty$. In addition, $|\mu_k| \leq b_\mathrm{max}$, thus, $\mu_k(\hat{\mu}_k-\mu_k)$ also follows a sub-Gaussian distribution. Similarly, it is easy to verify that $\Big|\frac{1}{|C|}\sum_{k \in C} 2\mu_k(\hat{\mu}_k-\mu_k) \Big| \to_p 0$. Therefore, we can conclude that, $ \frac{1}{|C|}\sum_{k \in C} (\hat{\mu}_k - \mu_0^{\mathrm{AP}})^2 \to_p  \frac{1}{|C|}\sum_{k \in C}\mathbb{E}[( \hat{\mu}_k-\mu_0^{\mathrm{AP}})^2]$. 

Since we have shown that both Parts A and B of Equation \eqref{eqn_VarianceMeanDecompose_4} converge to zeros in probability, we can conclude that, $\frac{1}{|C|}\sum_{k \in C}(\hat{\mu}_k - \hat{\mu}_0)^2 \to_p \frac{1}{|C|}\sum_{k \in C}\mathbb{E}[(\hat{\mu}_k - \mu_0^{\mathrm{AP}})^2]$. 
In addition, by definition, we have $\mathbb{E}[(\mu_0^{\mathrm{AP}} - \hat{\mu}_k)^2] = \frac{\sigma_k^2}{N} + (\mu_0^{\mathrm{AP}}-\mu_k)^2$. Therefore, we have shown that Equations \eqref{eqn_NoiseVarianceConverge} and \eqref{eqn_MeanVarianceConverge} hold under Assumption \ref{asmp_1} (ii). By Slutsky's Theorem and the fact that convergence in probability is equivalent to convergence in distribution, we can conclude that $\hat{\alpha} \to_p \alpha^\mathrm{AP}$.

For part (ii) of Proposition \ref{prop_DataDrivenPooling}, for ease of exposition, let $\hat{\theta} = \frac{\hat{\alpha}}{N+\hat{\alpha}}$ and $\theta^\mathrm{AP} = \frac{\alpha^\mathrm{AP}}{N+\alpha^\mathrm{AP}}$.
\begin{align*}
     & \frac{1}{|C|}\sum_{k \in C} (x_k(\hat{\alpha},\hat{\mu}_0,\hat{\mu}_k)-\mu_k)^2   -  \frac{1}{|C|}\sum_{k \in C} (x_k(\alpha^\mathrm{AP},\mu_0^{\mathrm{AP}},\hat{\mu}_k)-\mu_k)^2  \\
     =& \frac{1}{|C|}\sum_{k \in C} \Big((\hat{\mu}_k - \mu_k+\hat{\theta}(\hat{\mu}_0-\hat{\mu}_k))^2 -(\hat{\mu}_k - \mu_k+\theta^\mathrm{AP}(\mu_0^{\mathrm{AP}}-\hat{\mu}_k))^2 \Big) \\
    =  & \frac{1}{|C|}\sum_{k \in C} \Big( (\hat{\mu}_k-\mu_k)(2-\theta^\mathrm{AP}-\hat{\theta})(\hat{\theta}\hat{\mu}_0-\theta^\mathrm{AP}\mu_0^{\mathrm{AP}})-\mu_k(\hat{\mu}_k-\mu_k)(2-\theta^\mathrm{AP}-\hat{\theta})(\hat{\theta}-\theta^\mathrm{AP}) \\
    &+  (\hat{\theta}\hat{\mu}_0-\theta^\mathrm{AP}\mu_0^{\mathrm{AP}}-\mu_k(\hat{\theta}-\theta^\mathrm{AP}))(-\mu_k(\theta^\mathrm{AP}+\hat{\theta})+\hat{\theta}\hat{\mu}_0+\theta^\mathrm{AP}\mu_0^{\mathrm{AP}})-(\hat{\mu}_k-\mu_k)^2(2-\theta^\mathrm{AP}-\hat{\theta})(\hat{\theta}-\theta^\mathrm{AP}) \\
    &+\mu_k(\hat{\mu}_k-\mu_k)(\hat{\theta}-\theta^\mathrm{AP})(\theta^\mathrm{AP}+\hat{\theta})-(\hat{\mu}_k-\mu_k)(\hat{\theta}-\theta^\mathrm{AP})(\hat{\theta}\hat{\mu}_0+\theta^\mathrm{AP}\mu_0^{\mathrm{AP}})\Big) \\
    \le & 2\Big|(\hat{\mu}_0-\mu_0^{\mathrm{AP}})(\hat{\theta}\hat{\mu}_0-\theta^\mathrm{AP}\mu_0^{\mathrm{AP}})\Big| +   4|\hat{\theta}-\theta^\mathrm{AP}|\Big|\frac{1}{|C|}\sum_{k \in C} \mu_k(\hat{\mu}_k-\mu_k) \Big|+(b_\mathrm{max}+|\hat{\mu}_0|)\Big|(\hat{\mu}_0-\mu_0^{\mathrm{AP}})(\hat{\theta} - \theta^\mathrm{AP}) \Big| \\
    & +\Big|(|\hat{\theta}\hat{\mu}_0-\theta^\mathrm{AP}\mu_0^{\mathrm{AP}}|+b_\mathrm{max}|\hat{\theta}-\theta^\mathrm{AP}|)(3b_\mathrm{max}+|\hat{\mu}_0|) \Big| +2\Big|\frac{1}{|C|}\sum_{k \in C}(\hat{\mu}_k-\mu_k)^2\Big|\cdot\Big|(\hat{\theta}-\theta^\mathrm{AP}) \Big|.
\end{align*}
The last inequality holds due to the fact that $\hat{\theta} , \theta^\mathrm{AP} < 1$, and the true mean of each subproblem is bounded by $b_\mathrm{max}$ by Assumption \ref{asmp_1} (ii).

We have shown that $\Big|\frac{1}{|C|}\sum_{k \in C} \mu_k(\hat{\mu}_k-\mu_k) \Big| \to_p 0$ and $\hat{\mu}_0 \to_p \mu_0^{\mathrm{AP}}$ when Assumption \ref{asmp_1} (ii) holds. Since $\hat{\alpha} \to_p \alpha^\mathrm{AP}$, by Continuous Mapping Theorem, we have, $\hat{\theta} \to_p \theta^\mathrm{AP}$. In addition,  $\Big|\frac{1}{|C|}\sum_{k \in C}(\hat{\mu}_k-\mu_k)^2\Big| \le \frac{\sigma^2_\mathrm{max}}{N} \Big|\frac{1}{|C|}\sum_{k \in C}(\frac{\sqrt{N}(\hat{\mu}_k-\mu_k)}{\sigma_k})^2\Big|$. Furthermore, by Lemma \ref{lma_SubExponentialConcentration} , $\frac{1}{|C|}\sum_{k \in C}\big(\frac{\sqrt{N}(\hat{\mu}_k-\mu_k)}{\sigma_k}\big)^2 \to_p 1$. Thus, we have, as $|C|\to \infty$, $\Big|\frac{1}{|C|}\sum_{k \in C}(\hat{\mu}_k-\mu_k)^2\Big| \to_p \frac{\sigma^2_\mathrm{max}}{N}$, which is bounded. Thus, the right-hand side of the above equation converges to zero in probability. 
\begin{equation}
\frac{1}{|C|}\sum_{k \in C} (x_k(\hat{\alpha},\hat{\mu}_0,\hat{\mu}_k)-\mu_k)^2   \to_p  \frac{1}{|C|}\sum_{k \in C} (x_k(\alpha^\mathrm{AP},\mu_0^{\mathrm{AP}}, \hat{\mu}_k)-\mu_k)^2.
\end{equation}
Therefore, we have shown that Part (ii) of Proposition \ref{prop_DataDrivenPooling} when Assumption \ref{asmp_1} (ii) holds. This completes the proof for Proposition \ref{prop_DataDrivenPooling}.
\QED

{\bf \noindent Proof of Proposition \ref{prop_GivenClusterBenefit}.}
To connect the cost of the data pooling approach and that of the cluster-based data pooling approach, we consider a benchmark that implements the cluster-based data pooling approach with a single shrinkage parameter.  Recall that the optimal shrinkage parameters $\alpha_{\mu_{i,0}}^\mathrm{AP}, i = 1,2$, for the cluster-based data pooling approach are obtained by solving the problem defined in \eqref{eqn_OptimalShrinkageCluster}. In this benchmark, one additional constraint $\alpha_1 = \alpha_2$ is added, and thus, the corresponding expected out-of-sample cost provides an upper bound of that of the cluster-based data pooling approach. Specifically, the single shrinkage parameter can be computed by solving the following optimization problem.
\begin{equation}
    \tilde{\alpha}^\mathrm{AP} \in \argmin_{\alpha} \quad Z(\alpha,\mu_{1,0}^\mathrm{AP},C_1) + Z(\alpha,\mu_{2,0}^\mathrm{AP},C_2).
\end{equation}
Suppose the means and variances of the subproblems for each cluster are known, with direct computation, we obtain the shrinkage parameter,
\begin{equation}\label{eqn_BenchmarkCost}
    \tilde{\alpha}^\mathrm{AP} =  \frac{\sum_{k=1}^{K}\sigma_k^2}{\sum_{k\in C_1}(\mu_k-\mu_{1,0}^\mathrm{AP})^2+\sum_{k\in C_2}(\mu_k-\mu_{2,0}^\mathrm{AP})^2},
\end{equation}
and the corresponding  expected out-of-sample cost,
\begin{equation}
\sum_{i \in \{1,2\}}Z(\tilde{\alpha}^\mathrm{AP},\mu_{i,0}^\mathrm{AP},C_i)  = \sum_{k=1}^{K}\sigma_k^2+\Big(\sum_{k=1}^{K}\frac{\sigma_k^2}{N}\Big)\frac{N}{N+\tilde{\alpha}^\mathrm{AP}}.
\end{equation}
By optimality of $\alpha_{i}^\mathrm{AP}$, it is easy to verify that,
\begin{equation}\label{eqn_ClusterBenchmark}
\sum_{i \in \{1,2\}}Z(\alpha_{i}^\mathrm{AP},\mu_{i, 0}^\mathrm{AP},C_i)\le \sum_{i \in \{1,2\}}Z(\tilde{\alpha}^\mathrm{AP},\mu_{i, 0}^\mathrm{AP},C_i).
\end{equation}
By definitions of $\mu_0^\mathrm{AP}$, $\mu_{i,0}^\mathrm{AP}, i=1,2$, we have, 
\begin{equation}
    \sum_{k\in C_1}(\mu_k-\mu_{1,0}^\mathrm{AP})^2+\sum_{k\in C_2}(\mu_k-\mu_{2,0}^\mathrm{AP})^2 - \sum_{k\in \mathcal{K}}(\mu_k-\mu_{0}^\mathrm{AP})^2 = - \frac{1}{K}|C_2||C_1|(\mu_{1,0}^\mathrm{AP} - \mu_{2,0}^\mathrm{AP})^2 \le 0.
\end{equation} 
Thus, we have the following relationship between the shrinkage parameters for the data pooling approach and the benchmark.
\begin{equation}
    \tilde{\alpha}^\mathrm{AP} =  \frac{\sum_{k=1}^{K}\sigma_k^2}{\sum_{k\in C_1}(\mu_k-\mu_{1,0}^\mathrm{AP})^2+\sum_{k\in C_2}(\mu_k-\mu_{2,0}^\mathrm{AP})^2} \ge \frac{\sum_{k=1}^{K}\sigma_k^2}{\sum_{k\in \mathcal{K}}(\mu_k-\mu_{0}^\mathrm{AP})^2} = \alpha^\mathrm{AP}.
\end{equation}
Since the expected out-of-sample cost of the data pooling approach, defined in \eqref{eqn_GuptaCost}, and that of the benchmark, defined in \eqref{eqn_BenchmarkCost}, only differ in their shrinkage parameters, and such expected out-of-sample cost decreases in the shrinkage parameter, we have,
\begin{equation}\label{eqn_GuptaBenchmark}
\sum_{i \in \{1,2\}}Z(\tilde{\alpha}^\mathrm{AP},\mu_{i,0}^\mathrm{AP},C_i) \le Z(\alpha^\mathrm{AP},\mu_0^\mathrm{AP},\mathcal{K}).
\end{equation}
Combining the inequalities \eqref{eqn_ClusterBenchmark} and \eqref{eqn_GuptaBenchmark}, we have, 
\begin{equation}
\sum_{i \in \{1,2\}}Z(\alpha_{i}^\mathrm{AP},\mu_{i, 0}^\mathrm{AP},C_i) \leq 
    \sum_{i \in \{1,2\}}Z(\tilde{\alpha}^\mathrm{AP},\mu_{i,0}^\mathrm{AP} ,C_i) \le Z(\alpha^\mathrm{AP},\mu_0^\mathrm{AP},\mathcal{K}).
\end{equation}
In particular, if $\mu_{1,0}^\mathrm{AP} \neq \mu_{2,0}^\mathrm{AP}$, we have, $\frac{-|C_2||C_1|(\mu_{1,0}^\mathrm{AP} - \mu_{2,0}^\mathrm{AP})^2}{K} < 0$ and $\tilde{\alpha}^\mathrm{AP} > \alpha^{\text{AP}}$. Thus, we have,
\begin{equation}
    \sum_{i \in \{1,2\}}Z(\alpha_{i}^\mathrm{AP},\mu_{i, 0}^\mathrm{AP},C_i) \leq 
    \sum_{i \in \{1,2\}}Z(\tilde{\alpha}^\mathrm{AP},\mu_{i,0}^\mathrm{AP} ,C_i) < Z(\alpha^\mathrm{AP},\mu_0^\mathrm{AP},\mathcal{K}).
\end{equation}
This completes the proof. 
\QED

{\noindent \bf Proof of Theorem \ref{thm_DataDrivenPooling}.}
  We can prove Theorem \ref{thm_DataDrivenPooling} by applying Proposition \ref{prop_DataDrivenPooling} to each cluster $C_i$, $i=1,2$. 

\QED

{\noindent \bf Proof of Proposition \ref{prop_BenefitUnknownClusterN1}.}
Recall from Section \ref{sec_KnownClusterMSE}, to minimize the expected out-of-sample cost of the data pooling approach, $Z(\alpha^\mathrm{AP}, \mu_0^\mathrm{AP}, \mathcal{K})$, by Equation \eqref{eqn_OptAlphaPooling}, the optimal shrinkage parameter satisfies $\alpha^\mathrm{AP} = \frac{\frac{1}{K}\sum_{k =1}^K\sigma_k^2}{\frac{1}{K}\sum_{k=1}^{K}(\mu_k - \mu_0^\mathrm{AP})^2}
$ where $\mu_0^\mathrm{AP} = \frac{1}{K} \sum_{k=1}^K \mu_k$. Define  $\overline{\sigma}^2 = \frac{1}{K}\sum_{k =1}^K\sigma_k^2$.  Since
$\frac{1}{K}\sum_{k=1}^K (\mu_k -  \mu_0^\mathrm{AP})^2 \to \mathrm{Var}[\mu_k]$ as $K\to\infty$, we have,
\begin{equation}
 \alpha^\mathrm{AP} = \frac{\overline{\sigma}^2}{\frac{1}{K}\sum_{k=1}^{K}(\mu_k - \mu_0^\mathrm{AP})^2}\to \frac{12\overline{\sigma}^2}{(b-a)^2(1+3d+3d^2)}, 
\end{equation}
and consequently,
\begin{equation}
    \frac{1}{K} Z(\alpha^\mathrm{AP}, \mu_0^\mathrm{AP}, \mathcal{K}) \to \overline{\sigma}^2\Big(1+\Big(N+\frac{12\overline{\sigma}^2}{(b-a)^2(1+3d+3d^2)}\Big)^{-1}\Big).
\end{equation}

 When implementing the data pooling approach to each cluster separately with optimal shrinkage parameters, similar to Equation \eqref{eqn_APMSEKnownCluster}, we can compute the expected out-of-sample cost,
\begin{align*}\label{eqn_MSEUnknownCluster}
  \sum_{i \in \{1,2\}}Z^{p}(\alpha_i^\mathrm{AP},\mu_{i,0}^\mathrm{AP},C_i)  = \sum_{k=1}^{K}\sigma_k^2+ \sum_{k\in C_1}\frac{\sigma_k^2}{N-N_1+\alpha_1^\mathrm{AP}}+\sum_{k\in C_2}\frac{\sigma_k^2}{N-N_1+\alpha_2^\mathrm{AP}}. 
\end{align*}
If we add a constraint such that the shrinkage parameters for two clusters have to be the same, that is, $\alpha_1^\mathrm{AP} = \alpha_2^\mathrm{AP}$, recall the proof of Proposition \ref{prop_GivenClusterBenefit}, we can derive the optimal common shrinkage parameter as,  $\tilde{\alpha}^\mathrm{AP} =  \frac{\sum_{k=1}^{K}\sigma_k^2}{\sum_{k\in C_1}(\mu_k-\mu_{1,0}^\mathrm{AP})^2+\sum_{k\in C_2}(\mu_k-\mu_{2,0}^\mathrm{AP})^2}$. The following relationship holds for any given $K$, where the left-hand side provides an upper bound of the out-of-sample cost of cluster-based data pooling.
\begin{align*}
     \sum_{i \in \{1,2\}}Z^{p}(\alpha_i^\mathrm{AP},\mu_{i,0}^\mathrm{AP},C_i) \le \sum_{i \in \{1,2\}} Z^{p}(\tilde\alpha^\mathrm{AP},\mu_{i,0}^\mathrm{AP},C_i) = \sum_{k=1}^{K}\Big(\sigma_k^2+ \frac{\sigma_k^2}{N-N_1+\tilde{\alpha}^\mathrm{AP}}\Big).
\end{align*}
As $K\to\infty$, we have, $\frac{1}{|C_i|} \sum_{k\in C_i} (\mu_k -\mu_{i,0}^\mathrm{AP})^2 \to \mathrm{Var}[\mu_k|k\in C_i]$. Consequently, we have, $\tilde{\alpha}^\mathrm{AP} \to \frac{48\overline{\sigma}^2}{(b-a)^2}$. Thus, we have, 
\begin{equation}
 \lim_{K\to\infty}\frac{1}{K}\sum_{i \in \{1,2\}}Z^{p}(\tilde\alpha^\mathrm{AP},\mu_{i,0}^\mathrm{AP},C_i)  = \overline{\sigma}^2\Big(1 + \Big(N-N_1+\frac{48\overline{\sigma}^2}{(b-a)^2}\Big)^{-1}\Big).
\end{equation}
The gap between out-of-sample costs associated with direct data pooling and cluster-based data pooling can be decomposed as,
\begin{eqnarray*}
    \Delta &=& \lim_{K\to\infty} \frac{1}{K} \Big(Z(\alpha^\mathrm{AP}, \mu_0^\mathrm{AP},\mathcal{K}) - \sum_{i \in \{1,2\}}Z^{p}(\alpha_i^\mathrm{AP},\mu_{i,0}^\mathrm{AP},C_i) \Big)\\
    &=& \lim_{K\to\infty} \frac{1}{K} \Big( Z(\alpha^\mathrm{AP}, \mu_0^\mathrm{AP},\mathcal{K})  - \sum_{i \in \{1,2\}} Z^{p}(\tilde\alpha^\mathrm{AP},\mu_{i,0}^\mathrm{AP},C_i) +\sum_{i \in \{1,2\}}Z^{p}(\tilde\alpha^\mathrm{AP},\mu_{i,0}^\mathrm{AP},C_i) - \sum_{i \in \{1,2\}}Z^{p}(\alpha_i^\mathrm{AP},\mu_{i,0}^\mathrm{AP},C_i)\Big). 
\end{eqnarray*}
We first show that the difference between the first two terms is positive. 
When $\frac{\overline{\sigma}}{\sqrt{N_1}} \ge \frac{b-a}{6}$ and $d\ge \sqrt{\frac{1}{12-\frac{N_1(b-a)^2}{4\overline{\sigma}^2}}-\frac{1}{12}}-\frac{1}{2}$, we have, $\frac{48\overline{\sigma}^2}{(b-a)^2} - N_1 \geq \frac{12\overline{\sigma}^2}{(b-a)^2(1+3d+3d^2)}$, and thus,
\begin{eqnarray*}
  \Delta_1 &=&  \lim_{K\to\infty} \frac{1}{K} \Big( Z(\alpha^\mathrm{AP}, \mu_0^\mathrm{AP},\mathcal{K})  - \sum_{i \in \{1,2\}} Z^{p}(\tilde\alpha^\mathrm{AP},\mu_{i,0}^\mathrm{AP},C_i) \Big)  \\
           &=&  \overline{\sigma}^2\Big(1+\Big(N+\frac{12\overline{\sigma}^2}{(b-a)^2(1+3d+3d^2)}\Big)^{-1}\Big) - \overline{\sigma}^2\Big(1 + \Big(N-N_1+\frac{48\overline{\sigma}^2}{(b-a)^2}\Big)^{-1}\Big) \geq 0. \label{prop_3.3_final_eq}
\end{eqnarray*}
For the last two terms, we have,
\begin{align*}
    \Delta_2 = \lim_{K \to \infty} \frac{1}{K} \Big(\sum_{i \in \{1,2\}}Z^{p}(\tilde\alpha^\mathrm{AP},\mu_{i,0}^\mathrm{AP},C_i) - \sum_{i \in \{1,2\}}Z^{p}(\alpha_i^\mathrm{AP},\mu_{i,0}^\mathrm{AP},C_i)\Big) \ge 0.
\end{align*}
Finally, combining the above inequalities, we have,
$\Delta = \Delta_1+\Delta_2  \ge 0.$
Furthermore,  it is easy to verify that $\Delta$ increases in $d$ as the out-of-sample cost of direct data pooling increases in $d$. This completes the proof.
\QED

The proof of Theorem \ref{thm_DataDrivenClusterBenefitThreshold} requires the following lemmas.

Lemma \ref{lma_ClusterRatio} characterizes the percentage of problems that belong to an underlying cluster or an estimated cluster. For ease of exposition, let $R_{i,j}$ denote the percentage of problems that belong to Cluster $i$ but are grouped into Cluster $j$ using $N_1$ data points with the clustering Algorithm \ref{alg_Cluster}. Specifically, $R_{i,j} = \frac{1}{K} \sum_{k=1}^K \mathds{1}(k \in C_i, k\in \hat{C}_j), i = 1,2, j = 1,2$.

\begin{lemma} \label{lma_ClusterRatio}
Under Assumption \ref{asmp_2}, for $i \in \{1,2\}$, as $K \to \infty$, we have, 
\begin{itemize}
    \item[(i)] $\frac{|C_i|}{K} \to_p \frac{1}{2}$;
    \item[(ii)] if the cluster structure is estimated by setting the average sample mean of all problems as the clustering boundary, then,  $\frac{|\hat{C}_i|}{K} \to_p \frac{1}{2}$.
\end{itemize} 
\end{lemma}
{\bf \noindent Proof of Lemma \ref{lma_ClusterRatio}.}
For the underlying cluster structure $(C_1, C_2)$,  without loss of generality, when the sampled mean $\mu_k \in [a,\frac{a+b}{2}]$, then $k \in C_1$, and when the sampled mean $\mu_k \in [\frac{a+b}{2}+d(b-a),b+d(b-a)]$, then $k \in C_2$. Since the means are uniformly distributed over the disjoint intervals of $[a,\frac{a+b}{2}]$ and $[\frac{a+b}{2}+d(b-a),b+d(b-a)]$ of which the lengths are the same, as $K \to \infty$, $\frac{|C_i|}{K} \to_p \frac{1}{2}$ for any $i \in \{1,2\}$.

We assume that the decision maker sets the average sample mean of all problems as the clustering boundary which means $\frac{1}{K}\sum_{k \in \mathrm{k}} \hat{\mu}_k^c$. Then when $K \to \infty$, the  boundary that separates two clusters $\hat{C}_1$ and $\hat{C}_2$ goes to $\frac{a+b}{2}+\frac{d(b-a)}{2}$, that is, $\hat{\mu}_k^c < (>) \frac{a+b}{2}+\frac{d(b-a)}{2}$ implies $k \in \hat{C}_1 (\hat{C}_2)$. 
 By Assumption \ref{asmp_2}, the mean of $N_1$ data points for any $k \in \mathcal{K}$ follows the Gaussian distribution with mean $\mu_k$ and variance $\frac{\sigma_k^2}{N_1}$. Moreover, for any $k \in C_1$, there exists $k^\prime \in C_2$ such that their means are symmetric around the boundary $\frac{a+b}{2}+\frac{d(b-a)}{2}$, i.e., $\frac{a+b}{2}+\frac{d(b-a)}{2} - \mu_k = \mu_{k^\prime} - \frac{a+b}{2}+\frac{d(b-a)}{2}$, and
\begin{align*}
P(k \in \hat{C}_2) = \int_{\frac{a+b}{2}+\frac{d(b-a)}{2}}^{\infty}\frac{\sqrt{N_1}}{\sqrt{2\pi}\sigma_k} e^{-\frac{N_1(x-\mu_k)^2}{2\sigma_k^2}} d(x)= \int_{-\infty}^{\frac{a+b}{2}+\frac{d(b-a)}{2}}\frac{\sqrt{N_1}}{\sqrt{2\pi}\sigma_k} e^{-\frac{N_1(x-\mu_{k^\prime})^2}{2\sigma_k^2}} d(x) = P(k^\prime \in \hat{C}_1).
\end{align*}
The above equality holds according to the symmetric property of Gaussian distribution. Thus, the probability of grouping the problems in the cluster $C_1$ to $\hat{C}_2$ is the same as that of group problems in $C_2$ to $\hat{C}_1$, which implies that $R_{1,2} - R_{2,1} \to_p 0$ as $K \to \infty$. By definition, we have, $|C_i| = K(R_{i,i}+R_{i,j})$ and $|\hat{C}_i| = K(R_{i,i}+R_{j,i})$. Thus, $\frac{|\hat{C}_i|}{K} = \frac{|C_i|}{K} +R_{j,i} - R_{i,j}$. Because $\frac{|C_i|}{K} \to_p \frac{1}{2}$ and $R_{1,2} - R_{2,1} \to_p 0$, as $K \to \infty$, we have, $\frac{|\hat{C}_i|}{K} \to_p \frac{1}{2}$ for any $i \in \{1,2\}$.
\QED

\begin{lemma} \label{lemma_1}
When Assumption \ref{asmp_2} holds, suppose for each $k \in \mathcal{K}$ that the cost function of the $k$-th subproblem is the mean squared error, that is, $c_k = (x_k-\xi_k)^2$. Given $N$ data points for each subproblem, $N_1 < N - 1$ data points are used to estimate obtain the cluster structure $(\hat{C}_1, \hat{C}_2)$. For $i=1, 2$, let
\begin{align*}
    \hat{\mu}_{i,0}(\hat{C}_i) = \frac{1}{|\hat{C}_i|}\sum_{k\in \hat{C}_i} \hat{\mu}_k^p,\quad
    \hat{\alpha}_i(\hat{C}_i) = \frac{\sum_{k \in \hat{C}_i}\frac{1}{N-N_1-1}\sum_{j = N_1+1}^{N}(\hat{\xi}_{kj}-\hat{\mu}_k^p)^2}{\sum_{k\in \hat{C}_i}(\hat{\mu}_{i,0}(\hat{C}_i)-\hat{\mu}_k^p)^2-\frac{1}{N-N_1}\sum_{k \in \hat{C}_i}\frac{1}{N-N_1 - 1}\sum_{j = N_1+1}^{N}(\hat{\xi}_{kj}-\hat{\mu}_k^p)^2}.    \end{align*}
As $K \to \infty$, we have, for $i=1, 2$,
\begin{itemize}
    \item[(i)] $\hat{\alpha}_i(\hat{C}_i) \to_p \alpha_i^\mathrm{AP}(\hat{C}_i)$;
    

    \item[(ii)] $ \frac{1}{|\hat{C}_i|}\sum_{k \in \hat{C}_i}  (x_k(\hat{\alpha}_i(\hat{C}_i),\hat{\mu}_{i,0}(\hat{C}_i),\hat{\mu}_k^p)-\mu_k)^2  \to_p \frac{1}{|C|}\sum_{k \in C}  (x_k(\alpha_i^\mathrm{AP}(\hat{C}_i),\mu_{i,0}^{\mathrm{AP}}(\hat{C}_i),\hat{\mu}_k^p)-\mu_k)^2$
\end{itemize}
\end{lemma}
{\bf Proof of Lemma \ref{lemma_1}.} The proof of Lemma \ref{lemma_1} is similar to Part (ii) of  Proposition \ref{prop_DataDrivenPooling}.
The key difference is that $\hat{C}_i$ is a random set that depends on the dataset $S_k^c$. Since the dataset for pooling $S_k^p$ is independent of $S_k^c$, and $|\hat{C}_i| \to \infty$ as $K \to \infty$, the proof can be obtained the same as that of Part (ii) of Proposition \ref{prop_DataDrivenPooling}.
\QED

Let $\mathrm{Erf}(x) =\frac{2}{\sqrt{\pi}}\int_{0}^{x}e^{-t^2}dt$ denote the {\em error function} and $\mathrm{Erfc}(x) = 1 - \mathrm{Erf}(x)$ the {\em complementary error function}.
 The following lemma \ref{eqn_WrongClusterRatio} provides an upper bound of the percentage $R_{i,j}$ as $K \to \infty$ for $i\neq j$. 

\begin{lemma}\label{eqn_WrongClusterRatio}
When Assumption \ref{asmp_2} holds, suppose for each $k \in \mathcal{K}$ that the cost function of the $k$-th subproblem is the mean squared error, that is, $c_k = (x_k-\xi_k)^2$. Given $N$ data points for each subproblem, $N_1 < N$ data points are used to estimate obtain the cluster structure $(\hat{C}_1, \hat{C}_2)$ and we denote $\sigma_{max} = max_k\{ \sigma_k \}$. We have,
\begin{equation}
    \lim_{K\to\infty} R_{i,j} =  \lim_{K\to\infty} R_{j,i} \leq \frac{1}{4} \mathrm{Erfc} \Big[ \frac{(b-a)d\sqrt{N_1}}{2\sqrt{2}\sigma_{max}}\Big].
\end{equation}
\end{lemma}

{\bf \noindent Proof of Lemma \ref{eqn_WrongClusterRatio}.}
By Lemma \ref{lma_ClusterRatio}, we have, $R_{i,j} = R_{j,i}$. Recall from the proof of Lemma \ref{lma_ClusterRatio}, for any $k \in \mathcal{K}$, the probability that the subproblem associated with $\mu_k$ is grouped into Cluster $\hat{C}_1$ is given by,
\begin{align}
    P(k \in \hat{C}_1) = \int_{-\infty}^{\frac{a+b+d(b-a)}{2}}\frac{\sqrt{N_1}}{\sqrt{2\pi}\sigma_{k}}e^{-\frac{N_1(x-\mu_k)^2}{2\sigma_k^2}}dx = \frac{1}{2}\mathrm{Erfc}\Big[\frac{\mu_k-\frac{1}{2}(a+b+d(b-a)) }{\sqrt{2/N_1} \sigma_k} \Big].
\end{align}
Since the complementary error function is a decreasing function, the probability $P(k \in \hat{C}_1)$ is maximized when $\mu_k = \frac{a+b}{2}+d(b-a)$ for $k\in C_2$.  Thus, we have,
\begin{equation}
P(k \in \hat{C}_1|k \in C_2)  \leq  \frac{1}{2}\mathrm{Erfc}\Big[\frac{(b-a)d\sqrt{N_1}}{2\sqrt{2}\sigma_k}\Big].
\end{equation}
Therefore, we can bound the percentage of problems that belong to Cluster $C_2$ but are grouped into Cluster $\hat{C}_1$ using the sample mean $\hat{\mu}_k^c$,  
\begin{align}
    \lim_{K \to \infty} R_{2,1}  & = \int_{(b+a)/2+d(b-a)}^{b+d(b-a)} \frac{1}{b-a} P(k\in \hat{C}_1 | k \in C_2) d \mu_k\\
    & \le \int_{(b+a)/2+d(b-a)}^{b+d(b-a)}\frac{1}{b-a}\frac{1}{2}\mathrm{Erfc}\Big[ \frac{(b-a)d\sqrt{N_1}}{2\sqrt{2}\sigma_k}\Big]d \mu_k \\
    & \le \frac{1}{4}\mathrm{Erfc}\Big[ \frac{(b-a)d\sqrt{N_1}}{2\sqrt{2}\sigma_\mathrm{max}}\Big].
\end{align}
The above proof also applies to $R_{1,2}$. This completes the proof.
\QED

\begin{lemma}\label{lma_CostDifferenceBound}
When Assumption \ref{asmp_2} holds, suppose for each $k \in \mathcal{K}$ that the cost function of the $k$-th subproblem is the mean squared error, that is, $c_k = (x_k-\xi_k)^2$. Given $N$ data points for each subproblem, $N_1 < N$ data points are used to estimate the cluster structure $(\hat{C}_1, \hat{C}_2)$. There exists a random variable $\overline{G}$ and a deterministic function of $d$, $\overline{g}(d)$, such that
\begin{itemize}
    \item[(i)] $\frac{1}{K}\sum_{i \in \{1,2\}}\sum_{k \in \hat{C}_i}(x_k(\alpha_i^{\mathrm{AP}}(\hat{C}_i),\mu_i^{\mathrm{AP}}(\hat{C}_i),\hat{\mu}_k^p) -\mu_k)^2 - \frac{1}{K}\sum_{i \in \{1,2\}} \sum_{k \in C_i}(x_k(\alpha_i^{\mathrm{AP}}(C_i),\mu_i^{\mathrm{AP}}(C_i),\hat{\mu}_k^p) -\mu_k)^2 \le \overline{G}$ w.p. 1;
    \item[(ii)] $\overline{G} \to_p \overline{g}(d)$ as $K \to \infty$, and $\overline{g}(d) = O(d^3e^{-d^2})$.
\end{itemize}
\end{lemma}

{\bf \noindent Proof of Lemma \ref{lma_CostDifferenceBound}.} For ease of exposition, let $\hat{\theta}_i^\mathrm{AP} = \frac{\alpha^\mathrm{AP}(\hat{C}_i)}{N-N_1+\alpha^\mathrm{AP}(\hat{C}_i)}$ ,$\theta_i^\mathrm{AP} = \frac{\alpha^\mathrm{AP}(C_i)}{N-N_1+\alpha^\mathrm{AP}(C_i)}$ and $\sigma_\mathrm{max} = \max_k\{ \sigma_k \}$. Let $\mu(C) = \frac{1}{|C|}\sum_{i\in C}\mu_i$, and thus, we have, $\mu_{i,0}^\mathrm{AP}(C_i) = \mu(C_i)$ and $\mu_{i,0}^\mathrm{AP}(\hat{C}_i) = \mu(\hat{C}_i)$. By definition, the difference in costs defined in Part (i) of Lemma \ref{lma_CostDifferenceBound} can be decomposed into four cases as follows. 
\begin{align}\label{eqn_CostDifferenceDecompose}
\begin{split}
     & \frac{1}{K}\sum_{i \in \{1,2\}}\sum_{k \in \hat{C}_i}(x_k(\alpha_i^{\mathrm{AP}}(\hat{C}_i),\mu_i^{\mathrm{AP}}(\hat{C}_i),\hat{\mu}_k^p) -\mu_k)^2 - \frac{1}{K}\sum_{i \in \{1,2\}} \sum_{k \in C_i}(x_k(\alpha_i^{\mathrm{AP}}(C_i),\mu_i^{\mathrm{AP}}(C_i),\hat{\mu}_k^p) -\mu_k)^2 \\
    = & \frac{1}{K} \sum_{k=1}^K  
    \sum_{i=1}^2\sum_{j=1}^2 \mathds{1}_{k \in \hat{C}_i,k \in C_j} \Big((\mu_k -\hat{\mu}_k^p+ \hat{\theta}_i^\mathrm{AP}(\mu(\hat{C_i})-\hat{\mu}_k^p))^2-(\mu_k -\hat{\mu}_k^p+ \theta_j^\mathrm{AP}(\mu(C_j)-\hat{\mu}_k^p))^2 \Big). 
    \end{split}
\end{align}
To bound each component in the above equation, we analyze the following four cases respectively. For ease of exposition, we denote, for  $i,j \in \{1,2\}$,
\begin{equation*}
\hat{m}_1(\hat{C}_i,C_j) = \frac{\sum_{k = 1}^{K}\mathds{1}_{k\in \hat{C}_i,k \in C_j}\hat{\mu}_k^p}{\sum_{k = 1}^{K}\mathds{1}_{k\in \hat{C}_i,k \in C_j}}, \hat{m}_2(\hat{C}_i,C_j) = \frac{\sum_{k = 1}^{K}\mathds{1}_{k\in \hat{C}_i,k \in C_j}(\hat{\mu}_k^p)^2}{\sum_{k = 1}^{K}\mathds{1}_{k\in \hat{C}_i,k \in C_j}},\hat{m}_3(\hat{C}_i,C_j) = \frac{\sum_{k = 1}^{K}\mathds{1}_{k\in \hat{C}_i,k \in C_j}\hat{\mu}_k^p \mu_k}{\sum_{k = 1}^{K}\mathds{1}_{k\in \hat{C}_i,k \in C_j}},
\end{equation*}
and 
\begin{equation*}
m_1(\hat{C}_i,C_j) = \frac{\sum_{k = 1}^{K}\mathds{1}_{k\in \hat{C}_i,k \in C_j}\mu_k}{\sum_{k = 1}^{K}\mathds{1}_{k\in \hat{C}_i,k \in C_j}}, m_2(\hat{C}_i,C_j) = \frac{\sum_{k = 1}^{K}\mathds{1}_{k\in \hat{C}_i,k \in C_j}\big(\mu_k^2+\frac{\sigma_k^2}{N-N_1}\big)}{\sum_{k = 1}^{K}\mathds{1}_{k\in \hat{C}_i,k \in C_j}},m_3(\hat{C}_i,C_j) = \frac{\sum_{k = 1}^{K}\mathds{1}_{k\in \hat{C}_i,k \in C_j}\mu_k^2}{\sum_{k = 1}^{K}\mathds{1}_{k\in \hat{C}_i,k \in C_j}} .
\end{equation*}
According to Assumption \ref{asmp_2}, we have $m_1(\hat{C}_i,C_j) \le d(b-a) + b$ , $m_2(\hat{C}_i,C_j) \le (d(b-a) + b)^2 + \frac{\sigma_{max}^2}{N-1}$ and $m_3(\hat{C}_i,C_j) \le (d(b-a) + b)^2$.

{\noindent \em Case 1.} For the set of problems $\{k\in\mathcal{K}: k\in\hat{C}_1$, $k\in C_1\}$, we have,
\begin{align*}
&\frac{1}{K}\sum_{k\in \hat{C}_1,k\in C_1}(\hat{\mu}_k^p - \mu_k + \hat{\theta}_1^\mathrm{AP}(\mu(\hat{C_1})- \hat{\mu}_k^p ))^2-(\hat{\mu}_k^p - \mu_k + \theta_1^\mathrm{AP}(\mu(C_1)-\hat{\mu}_k^p ))^2 \\
= &  \frac{1}{K}\sum_{k\in \hat{C}_1,k\in C_1} \Big( \hat{\mu}_k^p(2-\hat{\theta}_1^\mathrm{AP}-\theta_1^\mathrm{AP})(\hat{\theta}_1^\mathrm{AP}\mu(\hat{C}_1)-\theta_1^\mathrm{AP}\mu(C_1))-(\hat{\mu}_k^p)^2(2-\hat{\theta}_1^\mathrm{AP}-\theta_1^\mathrm{AP})(\hat{\theta}_1^\mathrm{AP}-\theta_1^\mathrm{AP}) \\
    &~~~~~~~~~~~~~~~~+(\hat{\theta}_1^\mathrm{AP}\mu(\hat{C}_1) + \theta_1^\mathrm{AP}\mu(C_1)-2\mu_k)(\hat{\theta}_1^\mathrm{AP}\mu(\hat{C}_1)-\theta_1^\mathrm{AP}\mu(C_1))\\ 
    &~~~~~~~~~~~~~~~~- \hat{\mu}_k^p(\hat{\theta}_1^\mathrm{AP}-\theta_1^\mathrm{AP})(\hat{\theta}_1^\mathrm{AP}\mu(\hat{C}_1) + \theta_1^\mathrm{AP}\mu(C_1)-2\mu_k)\Big)\\
= &  R_{1,1}\big(\hat{m}_1(\hat{C}_1,C_1)(2-\hat{\theta}_1^\mathrm{AP}-\theta_1^\mathrm{AP})(\hat{\theta}_1^\mathrm{AP}\mu(\hat{C}_1)-\theta_1^\mathrm{AP}\mu(C_1))-\hat{m}_2(\hat{C}_1,C_1)(2-\hat{\theta}_1^\mathrm{AP}-\theta_1^\mathrm{AP})(\hat{\theta}_1^\mathrm{AP}-\theta_1^\mathrm{AP}) \\
    &~~~~~~+(\hat{\theta}_1^\mathrm{AP}\mu(\hat{C}_1) + \theta_1^\mathrm{AP}\mu(C_1)-2m_1(\hat{C}_1,C_1))(\hat{\theta}_1^\mathrm{AP}\mu(\hat{C}_1)-\theta_1^\mathrm{AP}\mu(C_1))\\
    &~~~~~~- \hat{m}_1(\hat{C}_1,C_1)(\hat{\theta}_1^\mathrm{AP}-\theta_1^\mathrm{AP})(\hat{\theta}_1^\mathrm{AP}\mu(\hat{C}_1) + \theta_1^\mathrm{AP}\mu(C_1))  + 2\hat{m}_3(\hat{C}_1,C_1)(\hat{\theta}_1^\mathrm{AP}-\theta_1^\mathrm{AP})\big) \\
\le & R_{1,1} \Big(2|\hat{m}_1(\hat{C}_1,C_1)||\hat{\theta}_1^\mathrm{AP}\mu(\hat{C}_1)-\theta_1^\mathrm{AP}\mu(C_1)| +2|\hat{m}_2(\hat{C}_1,C_1)||\hat{\theta}_1^\mathrm{AP}-\theta_1^\mathrm{AP}|\\
&~~~~~~+ 2(d(b-a) + b) |\hat{\theta}_1^\mathrm{AP}\mu(\hat{C}_1)-\theta_1^\mathrm{AP}\mu(C_1)|+|\hat{m}_1(\hat{C}_1,C_1)|2(d(b-a) + b)|\hat{\theta}_1^\mathrm{AP}-\theta_1^\mathrm{AP}|\\
&~~~~~~ +2|\hat{m}_3(\hat{C}_1,C_1)||\hat{\theta}_1^\mathrm{AP}-\theta_1^\mathrm{AP}| \Big)\\
 = & R_{1,1} \Big(\big(2|\hat{m}_1(\hat{C}_1,C_1)| + 2(d(b-a) + b) \big) |\hat{\theta}_1^\mathrm{AP}\mu(\hat{C}_1)-\theta_1^\mathrm{AP}\mu(C_1)|\\
&~~~~~~+ \big(2|\hat{m}_2(\hat{C}_1,C_1)| +2|\hat{m}_3(\hat{C}_1,C_1)|+ |\hat{m}_1(\hat{C}_1,C_1)|2(d(b-a) + b) \big)|\hat{\theta}_1^\mathrm{AP}-\theta_1^\mathrm{AP}| \Big).
\end{align*}
{\noindent \em Case 2.} For the set of problems $\{k\in\mathcal{K}: k\in\hat{C}_2,$  $k\in C_2\}$, similar to {\em Case 1}, we have,
\begin{align*}
&\frac{1}{K}\sum_{k\in \hat{C}_2,k\in C_2}\big(\hat{\mu}_k^p - \mu_k + \hat{\theta}_2^\mathrm{AP}(\mu(\hat{C_2})- \hat{\mu}_k^p )\big)^2-\big(\hat{\mu}_k^p - \mu_k + \theta_2^\mathrm{AP}(\mu(C_2)-\hat{\mu}_k^p)\big)^2 \\
\le &  R_{2,2} \Big(\big(2|\hat{m}_1(\hat{C}_2,C_2)| + 2(d(b-a) + b) \big) |\hat{\theta}_2^\mathrm{AP}\mu(\hat{C}_2)-\theta_2^\mathrm{AP}\mu(C_2)|\\
&~~~~~~+ \big(2|\hat{m}_2(\hat{C}_2,C_2)| +2|\hat{m}_3(\hat{C}_2,C_2)|+ |\hat{m}_1(\hat{C}_2,C_2)|2(d(b-a) + b) \big)|\hat{\theta}_2^\mathrm{AP}-\theta_2^\mathrm{AP}| \Big).
\end{align*}
{\noindent \em Case 3.} For the set of problems $\{k\in\mathcal{K}: k\in\hat{C}_1, k\in C_2\}$, we have,
\begin{align*}
&\frac{1}{K}\sum_{k\in \hat{C}_1,k\in C_2}\big(\hat{\mu}_k^p - \mu_k + \hat{\theta}_1^\mathrm{AP}(\mu(\hat{C_1})- \hat{\mu}_k^p )\big)^2-\big(\hat{\mu}_k^p - \mu_k + \theta_2^\mathrm{AP}(\mu(C_2)-\hat{\mu}_k^p )\big)^2\\
= & \frac{1}{K}\sum_{k\in \hat{C}_1,k\in C_2}\Big(\hat{\mu}_k^p(2-\hat{\theta}_1^\mathrm{AP}-\theta_2^\mathrm{AP})(\hat{\theta}_1^\mathrm{AP}\mu(\hat{C}_1)-\theta_2^\mathrm{AP}\mu(C_2))-(\hat{\mu}_k^p)^2(2-\hat{\theta}_1^\mathrm{AP}-\theta_2^\mathrm{AP})(\hat{\theta}_1^\mathrm{AP}-\theta_2^\mathrm{AP}) \\
    &~~~~~~~~~~~~~~~~+(\hat{\theta}_1^\mathrm{AP}\mu(\hat{C}_1) + \theta_2^\mathrm{AP}\mu(C_2)-2\mu_k)(\hat{\theta}_1^\mathrm{AP}\mu(\hat{C}_1)
    -\theta_2^\mathrm{AP}\mu(C_2))\\
    &~~~~~~~~~~~~~~~~- \hat{\mu}_k^p(\hat{\theta}_1^\mathrm{AP}-\theta_2^\mathrm{AP})(\hat{\theta}_1^\mathrm{AP}\mu(\hat{C}_1) + \theta_2^\mathrm{AP}\mu(C_2)-2\mu_k)\Big)\\
= & R_{2,1}\Big(\hat{m}_1(\hat{C}_1,C_2)(2-\hat{\theta}_1^\mathrm{AP}-\theta_2^\mathrm{AP})(\hat{\theta}_1^\mathrm{AP}\mu(\hat{C}_1)-\theta_2^\mathrm{AP}\mu(C_2))-\hat{m}_2(\hat{C}_1,C_2)(2-\hat{\theta}_1^\mathrm{AP}-\theta_2^\mathrm{AP})(\hat{\theta}_1^\mathrm{AP}-\theta_2^\mathrm{AP}) \\
    &~~~~~~+(\hat{\theta}_1^\mathrm{AP}\mu(\hat{C}_1) + \theta_2^\mathrm{AP}\mu(C_2)-2m_1(\hat{C}_1,C_2))(\hat{\theta}_1^\mathrm{AP}\mu(\hat{C}_1) -\theta_2^\mathrm{AP}\mu(C_2)) \\
    &~~~~~~- \hat{m}_1(\hat{C}_1,C_2)(\hat{\theta}_1^\mathrm{AP}-\theta_2^\mathrm{AP})(\hat{\theta}_1^\mathrm{AP}\mu(\hat{C}_1) + \theta_2^\mathrm{AP}\mu(C_2)) + 2\hat{m}_3(\hat{C}_1,C_2)(\hat{\theta}_1^\mathrm{AP}-\theta_2^\mathrm{AP})\Big) \\
 \le & R_{2,1}\Big(4|\hat{m}_1(\hat{C}_1,C_2)|(d(b-a) + b) + 2|\hat{m}_2(\hat{C}_1,C_2)|+4(d(b-a)+b)^2 + 2|\hat{m}_3(\hat{C}_1,C_2)|\Big).
\end{align*}
{\noindent\em Case 4.} For the set of problems $\{k\in\mathcal{K}: k\in\hat{C}_2$, $k\in C_1\}$, similar to {\em Case 3}, we can show that,
\begin{align*}
&\frac{1}{K}\sum_{k\in \hat{C}_2,k\in C_1}(\hat{\mu}_k^p - \mu_k + \hat{\theta}_2^\mathrm{AP}(\mu(\hat{C_1})- \hat{\mu}_k^p ))^2-(\hat{\mu}_k^p - \mu_k + \theta_1^\mathrm{AP}(\mu(C_2)-\hat{\mu}_k^p ))^2\\
\le &R_{1,2}\Big(4|\hat{m}_1(\hat{C}_2,C_1)|(d(b-a) + b) + 2|\hat{m}_2(\hat{C}_2,C_1)|+4(d(b-a)+b)^2 + 2|\hat{m}_3(\hat{C}_2,C_1)|\Big).
\end{align*}
Based on the analysis of the above four cases and Equation \eqref{eqn_CostDifferenceDecompose}, we have,
\begin{align}\label{eqn_CostDifference}
\begin{split}
     & \frac{1}{K}\sum_{i \in \{1,2\}}\sum_{k \in \hat{C}_i}(x_k(\alpha_i^{\mathrm{AP}}(\hat{C}_i),\mu_i^{\mathrm{AP}}(\hat{C}_i),\hat{\mu}_k^p) -\mu_k)^2 - \frac{1}{K}\sum_{i \in \{1,2\}} \sum_{k \in C_i}(x_k(\alpha_i^{\mathrm{AP}}(C_i),\mu_i^{\mathrm{AP}}(C_i),\hat{\mu}_k^p) -\mu_k)^2 \\
    \leq & R_{1,1}\Big( \big(2|\hat{m}_1(\hat{C}_1,C_1)| + 2(d(b-a) + b) \big)|\hat{\theta}_1^\mathrm{AP}\mu(\hat{C}_1)-\theta_1^\mathrm{AP}\mu(C_1)|\\
&~~~~~~+ \big(2|\hat{m}_2(\hat{C}_1,C_1)| + |\hat{m}_1(\hat{C}_1,C_1)|2(d(b-a) + b) + 2|\hat{m}_3(\hat{C}_1,C_1)|\big)|\hat{\theta}_1^\mathrm{AP}-\theta_1^\mathrm{AP}|\Big) \\
&+R_{2,2}\Big( \big(2|\hat{m}_1(\hat{C}_2,C_2)| + 2(d(b-a) + b) \big) |\hat{\theta}_2^\mathrm{AP}\mu(\hat{C}_2)-\theta_2^\mathrm{AP}\mu(C_2)|\\
&~~~~~~~~~+ \big(2|\hat{m}_2(\hat{C}_2,C_2)| + |\hat{m}_1(\hat{C}_2,C_2)|2(d(b-a) + b) + 2|\hat{m}_3(\hat{C}_2,C_2)|\big)|\hat{\theta}_2^\mathrm{AP}-\theta_2^\mathrm{AP}| \Big)\\
    &+ R_{2,1}\Big(4|\hat{m}_1(\hat{C}_1,C_2)|(d(b-a) + b) + 2|\hat{m}_2(\hat{C}_1,C_2)|+4(d(b-a)+b)^2   + 2|\hat{m}_3(\hat{C}_2,C_1)|\Big) \\
    & + R_{1,2}\Big(4|\hat{m}_1(\hat{C}_2,C_1)|(d(b-a) + b) + 2|\hat{m}_2(\hat{C}_2,C_1|+4(d(b-a)+b)^2 + 2|\hat{m}_3(\hat{C}_1,C_2)|\Big).
    \end{split}
\end{align}
Next, we show there exists a random variable $\overline{G}$ that provides an upper bound of the right-hand side of the above equation \eqref{eqn_CostDifference} with probability 1. Specifically, we first show that there exist random variables that provide pathwise upper bounds for $|\hat{\theta}_i^\mathrm{AP}-\theta_i^\mathrm{AP}|$ and $|\hat{\theta}_i^\mathrm{AP}\mu(\hat{C}_i)-\theta_i^\mathrm{AP}\mu(C_i)|$, respectively. As the discussion unfolds, to bound the two terms mentioned above, it reduces to bound the terms, $|\mu(C_i)-\mu(\hat{C}_i)|$ and $|\frac{1}{K}\sum_{k \in C_1}(\mu_k-\mu(C_1) )^2 - \frac{1}{K}\sum_{k \in \hat{C}_1}(\mu_k-\mu(\hat{C}_1))^2 |$.

Recall that,
\begin{equation}
    {\alpha}^\mathrm{AP}(\hat{C}_1) = \frac{\sum_{k\in \hat{C}_1 }\sigma_k^2}{\sum_{k \in \hat{C}_1}(\mu_k-\mu(\hat{C}_1))^2}, \quad \alpha^\mathrm{AP}(C_1) = \frac{\sum_{k\in C_1 }\sigma_k^2}{\sum_{k \in C_1}(\mu_k-\mu(C_1))^2}.
\end{equation}
In addition, we have,
\begin{align}\label{eqn_muDifference}
\begin{split}
    &|\mu(C_1)-\mu(\hat{C}_1)| \\
    = &\frac{1}{|\hat{C_1}||C_1|} \Big | |\hat{C_1}|\sum_{k\in C_1}\mu_k -|C_1|\sum_{k\in \hat{C}_1}\mu_k \Big|\\
     = & \frac{1}{|\hat{C_1}||C_1|}\Big | \sum_{k \in C_1}\mathds{1}_{k\in \hat{C}_1, k \in C_1}(|\hat{C_1}| -|C_1|)\mu_k +\sum_{k\in C_1 }\mathds{1}_{k\in C_1, k\notin \hat{C}_1}|\hat{C}_1|\mu_k -\sum_{k\in \hat{C}_1 }\mathds{1}_{k\notin C_1,k\in \hat{C}_1}|C_1|\mu_k\Big|\\
    \le &  \frac{1}{|\hat{C_1}||C_1|} \Big(\Big|\sum_{k \in C_1}\mathds{1}_{k\in \hat{C}_1, k \in C_1}(|\hat{C_1}| -|C_1|)\mu_k\Big| + \Big|\sum_{k\in C_1 }\mathds{1}_{k\in C_1,k\notin \hat{C}_1}|\hat{C}_1|\mu_k\Big|  + \Big| \sum_{k\in \hat{C}_1 }\mathds{1}_{k\notin C_1,k\in \hat{C}_1}|C_1|\mu_k \Big| \Big) \\ 
    \le & K\big(d(b-a)+b\big)\Bigg(  \Bigg| \frac{1}{|C_1| } - \frac{1}{|\hat{C_1}|} \Bigg|   + \frac{R_{1,2}}{|C_1|} + \frac{ R_{2,1}}{|\hat{C}_1|}\Bigg).
    \end{split}
\end{align}
The first inequality holds due to the triangle inequality. The second inequality holds due to the assumption that $\mu_k$ is bounded and the definition of $R_{i,j}$. Therefore, we have,
\begin{align}\label{eqn_MeanSquareDifference}
\begin{split}
&\Big|\frac{1}{K}\sum_{k \in C_1}(\mu_k-\mu(C_1) )^2 - \frac{1}{K}\sum_{k \in \hat{C}_1}(\mu_k-\mu(\hat{C}_1))^2 \Big| \\
    \le &  4(d(b-a)+b)|\mu(C_1)-\mu(\hat{C}_1)|R_{1,1} + 4(d(b-a)+b)^2  (R_{1,2}+R_{2,1}) \\
    \le & 4(d(b-a)+b)|\mu(C_1)-\mu(\hat{C}_1)| + 4(d(b-a)+b)^2  (R_{1,2}+R_{2,1})\\
    \le & 4(d(b-a)+b)^2  \bigg(  \Big| \frac{K}{|C_1| } - \frac{K}{|\hat{C_1}|} \Big|   + \frac{R_{1,2}K}{|C_1|} + \frac{ R_{2,1}K}{|\hat{C}_1|}+R_{1,2}+R_{2,1}\bigg).
\end{split}
\end{align}
The first inequality holds due to Assumption \ref{asmp_2} such that $\mu_k < (d(b-a)+b)$. The second inequality holds because $R_{1,1}\leq 1$ by definition. The third inequality holds due to \eqref{eqn_muDifference}.

Thus, we have,
\begin{align*}
    |\hat{\theta}_1^\mathrm{AP}-\theta_1^\mathrm{AP}| &= \frac{N-N_1}{(N-N_1+\hat{\alpha}_{1}^\mathrm{AP}(\hat{C}_1))(N-N_1+\alpha^\mathrm{AP}(C_1))}|\hat{\alpha}_{1}^\mathrm{AP}(\hat{C}_1)-\alpha_{1}^\mathrm{AP}(C_1)|\\
    &\le |\hat{\alpha}_{1}^\mathrm{AP}(\hat{C}_1)-\alpha_{1}^\mathrm{AP}(C_1)|\\
    &\le \max\Big\{\frac{1}{K}\sum_{k\in \hat{C}_1 }\sigma_k^2,\frac{1}{K}\sum_{k\in C_1 }\sigma_k^2\Big\}\Big|\frac{1}{\frac{1}{K}\sum_{k \in \hat{C}_1}(\mu_k-\mu(\hat{C}_1))^2}-\frac{1}{\frac{1}{K}\sum_{k \in C_1}(\mu_k-\mu(C_1))^2}\Big| \\
    & \le \frac{\Big|\frac{1}{K}\sum_{k \in C_1}(\mu_k-\mu(C_1))^2 - \frac{1}{K}\sum_{k \in \hat{C}_1}(\mu_k-\mu(\hat{C}_1))^2 \Big|}{(\frac{1}{K}\sum_{k \in \hat{C}_1}(\mu_k-\mu(\hat{C}_1))^2)(\frac{1}{K}\sum_{k \in C_1}(\mu_k-\mu(C_1))^2)}\sigma_\mathrm{max}^2 \\
    & \le \frac{4(d(b-a)+b)^2  \Big(  \Big| \frac{K}{|C_1| } - \frac{K}{|\hat{C_1}|} \Big|   + \frac{R_{1,2}K}{|C_1|} + \frac{ R_{2,1}K}{|\hat{C}_1|}+R_{1,2}+R_{2,1}\Big)}{(\frac{1}{K}\sum_{k \in \hat{C}_1}(\mu_k-\mu(\hat{C}_1))^2)(\frac{1}{K}\sum_{k \in C_1}(\mu_k-\mu(C_1))^2)}\sigma_\mathrm{max}^2 \\
    & \le \frac{4(d(b-a)+b)^2  \Big(  \Big| \frac{K}{|C_1| } - \frac{K}{|\hat{C_1}|} \Big|   + \frac{R_{1,2}K}{|C_1|} + \frac{ R_{2,1}K}{|\hat{C}_1|}+R_{1,2}+R_{2,1}\Big)}{\min \Big \{(\frac{1}{K}\sum_{k \in \hat{C}_1}(\mu_k-\mu(\hat{C}_1))^2)^2, (\frac{1}{K}\sum_{k \in C_1}(\mu_k-\mu(C_1))^2)^2 \Big \}}\sigma_\mathrm{max}^2. 
\end{align*}
The first inequality holds because the shrinkage parameter is always positive. The last inequality holds due to Equation \eqref{eqn_MeanSquareDifference}. Furthermore, to bound the term $|\theta_1^\mathrm{AP}\mu(C_1)-\hat{\theta}_1^\mathrm{AP}\mu(\hat{C}_1)|$,
we observe that,
\begin{align*}
    \frac{ |\theta_1^\mathrm{AP}\mu(C_1)-\hat{\theta}_1^\mathrm{AP}\mu(\hat{C}_1)|}{K(d(b-a)+b)} &\le R_{1,1}\frac{ \Big| \theta_1^\mathrm{AP} |\hat{C_1}|-\hat{\theta}_1^\mathrm{AP} |C_1| \Big|}{|\hat{C_1}| |C_1|} + \frac{R_{1,2}}{|C_1|} + \frac{R_{2,1}}{|\hat{C}_1|}\\
    &\le\frac{\Big| \theta_1^\mathrm{AP}|\hat{C_1}|-\theta_1^\mathrm{AP}|C_1|+\theta_1^\mathrm{AP}|C_1|-\hat{\theta}_1^\mathrm{AP}|C_1| \Big|}{|\hat{C_1}||C_1|}
    + \frac{R_{1,2}}{|C_1|} + \frac{R_{2,1}}{|\hat{C}_1|}\\
    & \le \frac{\theta_1^\mathrm{AP}\Big| |\hat{C_1}|-|C_1|\Big |+|C_1| \Big|\theta_1^\mathrm{AP}-\hat{\theta}_1^\mathrm{AP} \Big| }{|\hat{C_1}||C_1|}
    + \frac{R_{1,2}}{|C_1|} + \frac{R_{2,1}}{|\hat{C}_1|}\\
    &\le  \bigg| \frac{1}{|C_1|} - \frac{1}{|\hat{C_1}|} \bigg| + \frac{1}{|\hat{C}_1|} \big|\hat{\alpha}_{\mu(\hat{C}_1)}^\mathrm{AP}-\alpha_{\mu(C_1)}^\mathrm{AP}\big| + \frac{R_{1,2}}{|C_1|} + \frac{R_{2,1}}{|\hat{C}_1|}.
\end{align*}
The second inequality holds because $R_{1,1} < 1$. The third inequality holds due to triangle inequality. The last inequality holds because $\theta_1^\mathrm{AP} <1 $ and $|\hat{\theta}_1^\mathrm{AP}-\theta_1^\mathrm{AP}| < |\hat{\alpha}_{i}^\mathrm{AP}(\hat{C}_1)-\alpha_{1}^\mathrm{AP}(C_1)|$. 

Define random variables, for $i=1,2$, $j=3-i$,
\begin{align*}
\tilde{Y}_i &= \frac{4(d(b-a)+b)^2  \Big(  \Big| \frac{K}{|C_i| } - \frac{K}{|\hat{C_i}|} \Big|   + \frac{R_{i,j}K}{|C_i|} + \frac{ R_{j,i}K}{|\hat{C}_i|}+R_{i,j}+R_{j,i}\Big)}{\min \Big \{(\frac{1}{K}\sum_{k \in \hat{C}_1}(\mu_k-\mu(\hat{C}_1))^2)^2, (\frac{1}{K}\sum_{k \in C_1}(\mu_k-\mu(C_1))^2)^2 \Big \}}\sigma_\mathrm{max}^2,\\
Y_i &= K(d(b-a)+b) \Big(\Big| \frac{1}{|C_i|} - \frac{1}{|\hat{C_i}|} \Big| + \frac{1}{|\hat{C}_i|} \Tilde{Y}_i + \frac{R_{i,j}}{|C_i|} + \frac{R_{j,i}}{|\hat{C}_i|}\Big),\\
\overline{G}  &= \sum_{i=1}^2\Big(\Big(2|\hat{m}_1(\hat{C}_i,C_i)| + 2(d(b-a) + b) \Big)Y_i+\Big(2|\hat{m}_2(\hat{C}_i,C_i)| + 2|\hat{m}_3(\hat{C}_i,C_i)| + |\hat{m}_1(\hat{C}_i,C_i)|2(d(b-a) + b) \Big)\tilde{Y}_i\Big) \\
&+\sum_{i=1}^2 R_{i,3-i}\Big(4|\hat{m}_1(\hat{C}_i,C_{3-i})|(d(b-a) + b) + 2|\hat{m}_2(\hat{C}_i,C_{3-i})|+4(d(b-a)+b)^2 + 2|\hat{m}_3(\hat{C}_i,C_{3-i})|\Big).
\end{align*}
By the above definitions and Equation \eqref{eqn_CostDifference}, we can find the random variable $\overline{G}$ such that 
\begin{eqnarray*}
    &&\frac{1}{K}\sum_{i \in \{1,2\}}\sum_{k \in \hat{C}_i}\big(x_k(\alpha_i^{\mathrm{AP}}(\hat{C}_i),\mu_i^{\mathrm{AP}}(\hat{C}_i),\hat{\mu}_k^p) -\mu_k\big)^2 - \frac{1}{K}\sum_{i \in \{1,2\}} \sum_{k \in C_i}\big(x_k(\alpha_i^{\mathrm{AP}}(C_i),\mu_i^{\mathrm{AP}}(C_i),\hat{\mu}_k^p) -\mu_k\big)^2\\
    &\le & \overline{G},\quad \text{w.p. } 1.
\end{eqnarray*}
This completes Part (i) of Lemma \ref{lma_CostDifferenceBound}. 

As $K \to \infty$, by the proof of Lemma \ref{lma_ClusterRatio}, the boundary of clusters goes to $\frac{a+b}{2} + \frac{d(b-a)}{2}$. Condition on $S^c$, the indicator function $\mathds{1}(k\in \hat{C}_i, k\in C_j)$ is fixed. For the set of problems, $\{k\in \mathcal{K}: k\in \hat{C}_i, k\in C_j\}$, one can construct a sequence of random variables $\tilde\mu_k$ such that $\tilde{\mu}_k =_d \hat{\mu}_k^p$ on a common probability space where $=_d$ stands for equal in distribution. By Lemma \ref{lma_SubGaussinConcentration} and replacing $\hat{\mu}_k^p$ with $\tilde{\mu}_k$, one can derive that  $\hat{m}_1(\hat{C}_i,C_j) \to_p m_1(\hat{C}_i,C_j)$, $\hat{m}_2(\hat{C}_i,C_j) \to_p m_2(\hat{C}_i,C_j)$, and $\hat{m}_3(\hat{C}_i,C_j) \to_p m_3(\hat{C}_i,C_j)$ for $i,j \in \{1,2\}$.
In addition, as $K\to \infty$, by Lemma \ref{lma_ClusterRatio}, we have, $\frac{|C_1|}{K}\to_p \frac{1}{2}$ and $\frac{|\hat{C}_1|}{K}\to_p \frac{1}{2}$, we have $\frac{K^2}{ |C_1||\hat{C}_1|} \to_p 4$ and $\frac{K}{|C_i| } - \frac{K}{|\hat{C_i}|} \to_p 0$. 
In addition, when Assumption \ref{asmp_2} holds, $(C_1, C_2)$ is the oracle's cluster structure, and consequently,
\begin{equation}
\lim_{K\to \infty} \min \Bigg \{\Big(\frac{1}{|\hat{C}_1|}\sum_{k \in \hat{C}_1}\big(\mu_k-\mu(\hat{C}_1)\big)^2\Big)^2, \Big(\frac{1}{|C_1|}\sum_{k \in C_1}(\mu_k-\mu(C_1))^2\Big)^2 \Bigg \} = \lim_{K\to \infty}  \Big(\frac{1}{|C_1|}\sum_{k \in C_1}\big(\mu_k-\mu(C_1)\big)^2\Big)^2.
\end{equation}
In addition, $\frac{1}{|C_1|}\sum_{k \in C_1}(\mu_k-\mu(C_1))^2 \to \frac{(b-a)^2}{48}$ as $K\to \infty$. Therefore, by Sluskty's Theorem, we have,
\begin{equation}
\frac{K^2 }{|C_1||\hat{C}_1|\min \Big \{(\frac{1}{|\hat{C}_1|}\sum_{k \in \hat{C}_1}(\mu_k-\mu(\hat{C}_1))^2)^2, (\frac{1}{|C_1|}\sum_{k \in C_1}(\mu_k-\mu(C_1))^2)^2 \Big \}} \to_p  \frac{9216}{(b-a)^4}.
\end{equation}
By Sluskty's Theorem, we can show that, 
\begin{equation}
    \tilde{Y}_i \to_p  \tilde{y}_i = 12\frac{9216}{(b-a)^4}(d(b-a)+b)^2\sigma_\mathrm{max}^2   ({r}_{i,j}+{r}_{j,i}),
\end{equation}
where ${r}_{i,j} = \lim_{K\to\infty} R_{i,j}$.
Similarly, we can show that
\begin{equation}
Y_i  \to_p y_i =  2(d(b-a)+b)(\tilde{y}_i + {r}_{i,j}  + {r}_{j,i} ), \quad
\overline{G}  \to_p \overline{g},
\end{equation}
where 
\begin{align*}
\overline{g} &=  \sum_{i=1}^2\Big(\big(2|m_1(\hat{C}_i,C_i)| + 2(d(b-a) + b) \big)y_i+\big(2|m_2(\hat{C}_i,C_i)|+2|m_3(\hat{C}_i,C_i)| + |m_1(\hat{C}_i,C_i)|2(d(b-a) + b) \big)\tilde{y}_i\Big) \\
&~~~~~~~~~~+\sum_{i=1}^2 r_{i,3-i}\Big(4|m_1(\hat{C}_1,C_2)|(d(b-a) + b) + 2|m_2(\hat{C}_1,C_2)|+4(d(b-a)+b)^2  + 2|m_3(\hat{C}_1,C_2)|\Big).
\end{align*}

With the above relations, we have,
\begin{align*}
    &\lim_{K\to\infty} \overline{G}  \\
    &\leq 2\Bigg(24\frac{9216(d(b-a)+b)^2\sigma_\mathrm{max}^2}{(b-a)^4}\bigg((18d(b-a)+b)^2+\frac{2\sigma_\mathrm{max}^2}{N-N_1}\bigg)+48(d(b-a)+b)^2 + \frac{4\sigma_\mathrm{max}^2}{N-N_1}\Bigg)\\
    &~~~~\cdot\mathrm{Erfc}\Big[ \frac{(b-a)d\sqrt{N_1}}{2\sqrt{2}\sigma_\mathrm{max}}\Big] = O(d^3 e^{-d^2}).
\end{align*}
According to Lemma \ref{eqn_WrongClusterRatio}, we have, ${r}_{2,1} = {r}_{1,2} \leq \frac{1}{4} \mathrm{Erfc} \Big[ \frac{(b-a)d\sqrt{N_1}}{2\sqrt{2}\sigma_\mathrm{max}}\Big]$. Thus, the above inequality holds. The last equality holds because the general bound of the error function, that is, $\mathrm{Erfc}[x] \le \frac{e^{-x^2}}{x}$.
This completes the proof.
\QED

{\bf \noindent Proof of Theorem \ref{thm_DataDrivenClusterBenefitThreshold}.}
The difference between the expected out-of-sample cost of cluster-based direct data pooling approach with the estimated clusters and that of the direct data pooling approach is computed as,
\begin{align*}
     & \mathbb{E}_{S}\Big[\overline{Z}(\hat{\alpha},\hat{\mu}_0)\Big] - \mathbb{E}_{S}\Big[\overline{Z}^c(\hat{\bm{\alpha}}( \hat{\bm{C}}),\hat{\bm{\mu}}(\hat{\bm{C}}),\hat{\bm{C}})\Big] \\
      = & \mathbb{E}_{S}\Big[\overline{Z}(\hat{\alpha},\hat{\mu}_0)\Big] - \mathbb{E}_{S}\Big[\overline{Z}(\alpha^{\mathrm{AP}},\mu_0^{\mathrm{AP}})\Big]  + \mathbb{E}_{S}\Big[\overline{Z}(\alpha^{\mathrm{AP}},\mu_0^{\mathrm{AP}})\Big] -  \mathbb{E}_{S}\Big[\overline{Z}^c(\bm{\alpha}^{\mathrm{AP}}(\bm{C}),\bm{\mu}^{\mathrm{AP}}(\bm{C}),\bm{C})\Big] \\
     &+ \mathbb{E}_{S}\Big[\overline{Z}^c(\bm{\alpha}^{\mathrm{AP}}(\bm{C}),\bm{\mu}^{\mathrm{AP}}(\bm{C}),\bm{C})\Big]  - \mathbb{E}_{S}\Big[\overline{Z}^c(\bm{\alpha}^{\mathrm{AP}}(\hat{\bm{C}}),\bm{\mu}^{\mathrm{AP}}(\hat{\bm{C}}),\hat{\bm{C}})\Big] \\
     &+ \mathbb{E}_{S}\Big[\overline{Z}^c(\bm{\alpha}^{\mathrm{AP}}(\hat{\bm{C}}),\bm{\mu}^{\mathrm{AP}}(\hat{\bm{C}}),\hat{\bm{C}})\Big] - \mathbb{E}_{S}\Big[\overline{Z}^c(\hat{\bm{\alpha}}( \hat{\bm{C}}),\hat{\bm{\mu}}(\hat{\bm{C}}),\hat{\bm{C}})\Big].
\end{align*}
From Proposition \ref{prop_DataDrivenPooling} and Lemma \ref{lemma_1}, we have 
\begin{align}
&\mathbb{E}_{S}\Big[\overline{Z}(\hat{\alpha},\hat{\mu}_0)\Big] = \mathbb{E}_{S}\Big[\overline{Z}(\alpha^{\mathrm{AP}},\mu_0^{\mathrm{AP}})\Big],\\
&\mathbb{E}_{S}\Big[\overline{Z}^c(\bm{\alpha}^{\mathrm{AP}}(\hat{\bm{C}}),\bm{\mu}^{\mathrm{AP}}(\hat{\bm{C}}),\hat{\bm{C}})\Big] = \mathbb{E}_{S}\Big[\overline{Z}^c(\hat{\bm{\alpha}}( \hat{\bm{C}}),\hat{\bm{\mu}}(\hat{\bm{C}}),\hat{\bm{C}})\Big].
\end{align}
From the Proposition \ref{prop_BenefitUnknownClusterN1}, we have, 
\begin{eqnarray*}
&&\mathbb{E}_{S}\Big[\overline{Z}(\alpha^{\mathrm{AP}},\mu_0^{\mathrm{AP}})\Big] -  \mathbb{E}_{S}\Big[\overline{Z}^c(\bm{\alpha}^{\mathrm{AP}}(\bm{C}),\bm{\mu}^{\mathrm{AP}}(\bm{C}),\bm{C})\Big] \\
&=&\lim_{K\to\infty} \frac{1}{K}Z(\alpha^\mathrm{AP}, \mu_0^\mathrm{AP}, \mathcal{K}) - \frac{1}{K}\sum_{i=1}^2 Z^{p}(\alpha_{i}^\mathrm{AP}, \mu_{i,0}^\mathrm{AP}, C_i) > 0,
\end{eqnarray*}
and this difference increases as $d$ increases when the conditions stated in  Proposition \ref{prop_BenefitUnknownClusterN1} hold. Hence, there exists a constant $\gamma >0$ and $\bar{d}_1$ such that when $d \ge \bar{d}_1$ and $K \to \infty$, we have 
\begin{equation}
\mathbb{E}_{S}\Big[\overline{Z}(\alpha^{\mathrm{AP}},\mu_0^{\mathrm{AP}})\Big] -  \mathbb{E}_{S}\Big[\overline{Z}^c(\bm{\alpha}^{\mathrm{AP}}(\bm{C}),\bm{\mu}^{\mathrm{AP}}(\bm{C}),\bm{C})\Big]  > \gamma.
\end{equation}
 We have, $\mathbb{E}_{S}\Big[\overline{Z}^c(\bm{\alpha}^{\mathrm{AP}}(\bm{C}),\bm{\mu}^{\mathrm{AP}}(\bm{C}),\bm{C})\Big]  - \mathbb{E}_{S}\Big[\overline{Z}^c(\bm{\alpha}^{\mathrm{AP}}(\hat{\bm{C}}),\bm{\mu}^{\mathrm{AP}}(\hat{\bm{C}}),\hat{\bm{C}})\Big] $ is negative and its absolute value is $O(d^3 e^{-d^2})$ from the Lemma \ref{lma_CostDifferenceBound}. Hence, there exists $\overline{d}_2$, when $d \ge \overline{d}_2$,  the absolute value of the above difference in costs is smaller than $\gamma$ when $d \to \infty$.  

Let $d^* = \max\{\overline{d}_1, \overline{d}_2\}$. When $d \ge d^*$, we have,
\begin{equation}
      \mathbb{E}_{S}\Big[\overline{Z}(\hat{\alpha},\hat{\mu}_0)\Big] - \mathbb{E}_{S}\Big[\overline{Z}^c(\hat{\bm{\alpha}}( \hat{\bm{C}}),\hat{\bm{\mu}}(\hat{\bm{C}}),\hat{\bm{C}})\Big] > 0.
\end{equation}
This completes the proof.
\QED

{\bf \noindent Proof of Theorem \ref{thm_DataDrivenClusterWorse}.} When (i) $\tilde{y} > 6$, by Proposition \ref{prop_BenefitUnknownClusterN1}, we have $\lim_{K\to\infty} \frac{1}{K}Z(\alpha^\mathrm{AP}, \mu_0^\mathrm{AP}, \mathcal{K}) - \frac{1}{K}\sum_{i=1}^2 Z^{p}(\alpha_{i}^\mathrm{AP}, \mu_{i,0}^\mathrm{AP}, C_i)\leq 0$. We focus on the proof for condition (ii).

We first characterize the expected out-of-sample cost of the cluster-based data pooling approach. By the proof of Lemma \ref{lma_ClusterRatio}, for any problem $k$, the probability of grouping $k$ into the cluster $\hat{C}_2$ is given by, 
\begin{align}
    P(k \in \hat{C}_2) = \int_{\frac{a+b}{2}}^{\infty}\frac{\sqrt{N_1}}{\sqrt{2\pi}\sigma} e^{-\frac{N_1(x-\mu_k)^2}{2\sigma^2}} dx =\frac{1}{2} \mathrm{Erfc}\Big[ \frac{((a+b)/2-\mu_k)\sqrt{N_1}}{\sqrt{2}\sigma}\Big].
\end{align}
Thus, we have, as $K \to \infty$,
\begin{align*}
\frac{1}{|\hat{C}_2|}\sum_{k\in \hat{C}_2}(\mu_k - \mu(\hat{C}_2))^2 \to_p  \Gamma,
\end{align*}
where
\begin{align*}
\Gamma = & \frac{\sigma^2  \mathrm{Erf}\left[\frac{(a-b) \sqrt{N_1}}{2
\sqrt{2} \sigma }\right]^2 \left((a-b)^2 N_1 -2 \sigma ^2 \right)}{2(a-b)^2N_1^2}+ \frac{(a-b)^2}{48} \left(4-3 \mathrm{Erf}\Big[\frac{(a-b) \sqrt{N_1}}{2 \sqrt{2} \sigma }\Big]^2\right)  -\frac{ 
e^{-\frac{(a-b)^2 N_1}{\sigma^2}} \sigma^2 }{2 \pi  N_1} \\
&  -\frac{e^{-\frac{(a-b)^2 N_1}{8 \sigma ^2}}  \mathrm{Erf}\Big[\frac{(a-b)
\sqrt{N_1}}{2 \sqrt{2} \sigma }\Big]  \left((a-b)^2 N_1-4 \sigma ^2\right)\sigma}{\sqrt{8\pi}(a-b)N_1^{3/2}} .
\end{align*}
Since $\alpha_1^\mathrm{AP}(\hat{C}_1)$ is independent of the data set $S_k^p$, we have, as $K\to\infty$,
\begin{equation}
    {\alpha}_1^\mathrm{AP}(\hat{C}_1) = \frac{\sum_{k\in \hat{C}_1 }\sigma_k^2}{\sum_{k \in \hat{C}_1}(\mu_k-\mu_{i,0}^\mathrm{AP}(\hat{C}_1))^2} \to_p \frac{\sigma^2}{\Gamma}.
\end{equation}
By symmetry, we also have, $\alpha^\mathrm{AP}_{2}(\hat{C}_2) \to_p \frac{\sigma^2}{\Gamma}$.

Similar to the proof of Proposition \ref{prop_BenefitUnknownClusterN1}, the expected out-of-sample cost of the cluster-based data pooling approach satisfies,
\begin{align}
   \mathbb{E}_{S}\Big[\overline{Z}^c(\hat{\bm{\alpha}}^\mathrm{AP}(\hat{\bm{C}}),\hat{\bm{\mu}}^\mathrm{AP}(\hat{\bm{C}}),\hat{\bm{C}})\Big] = \sigma^2\Big(1+\Big(N-N_1+\frac{\sigma^2}{\Gamma}\Big)^{-1}\Big).
\end{align}
Recall from the proof of Proposition \ref{prop_BenefitUnknownClusterN1},  the expected out-of-sample cost with the direct data pooling approach,  is given by
\begin{equation}
\mathbb{E}_{S}[\overline{Z}(\alpha^\mathrm{AP},\mu_0^\mathrm{AP})] = \sigma^2\Big(1+\Big(N+\frac{12\sigma^2}{(b-a)^2(1+3d+3d^2)}\Big)^{-1}\Big).
\end{equation}
When $0\leq \tilde{y} \leq 6$, and $L(\tilde{y}) < 0$, we have,
\begin{equation}
    N-N_1+\frac{\sigma^2}{\Gamma} < N+\frac{12\sigma^2}{(b-a)^2(1+3d+3d^2)},
\end{equation}
and thus, 
\begin{equation}
\mathbb{E}_{S}\Big[\overline{Z}(\alpha^\mathrm{AP},\mu_0^\mathrm{AP})\Big] - \mathbb{E}_{S}\Big[\overline{Z}^c(\hat{\bm{\alpha}}^\mathrm{AP}(\hat{\bm{C}}),\hat{\bm{\mu}}^\mathrm{AP}(\hat{\bm{C}}),\hat{\bm{C}})\Big] < 0
\end{equation}
This completes the proof.
\QED

{\noindent \bf Proof of Theorem \ref{thm_GeneralCost}.}
   By the definition of $\text{ SubOpt}_{h_1,h_2}(\alpha_1,\alpha_2, C_1,C_2)$, we have,
\begin{eqnarray*}
    &&\text{SubOpt}_{h_1,h_2}(\alpha_{h_1,\hat{C}_1}^\mathrm{CS-SAA},\alpha_{h_2,\hat{C}_2}^\mathrm{CS-SAA}, \hat{C}_1, \hat{C}_2)\\
    && = \frac{1}{K}  \sum_{i\in \{1,2\}} \Big( {Z}_{\hat{C}_i}\big(\alpha_{h_i,\hat{C}_i}^\mathrm{CS-SAA},h_i\big) - {Z}_{\hat{C}_i}\big(\alpha_{h_i,\hat{C}_i}^\mathrm{OR},h_i\big) \Big)    +  \frac{1}{K} \sum_{i\in \{1,2\}} \Big( {Z}_{\hat{C}_i}\big(\alpha_{h_i,\hat{C}_i}^\mathrm{OR},h_i\big) -  {Z}_{C_i}\big(\alpha_{h_i,C_i}^\mathrm{OR},h_i\big) \Big). 
\end{eqnarray*}
We first bound the difference between the first two terms. Based on Theorem F.1 in \cite{GuptaKallus2022}, for any cluster $\hat{C_i}$, there exist universal constants $\tilde{A}_i$ such that for any $0<\delta<1/2$, with probability at least $1-\delta$ such that
\begin{align*}
    \frac{1}{|\hat{C}_i|} \Big( {Z}_{\hat{C}_i}\big(\alpha_{h_i,\hat{C}_i}^\mathrm{CS-SAA},h_i\big) - {Z}_{\hat{C}_i}\big(\alpha_{h_i,\hat{C}_i}^\mathrm{OR}, h_i\big) \Big) \le  \tilde{A}_i \max\{\Pi_2,L\sqrt{\Pi_2/\gamma}\}\cdot \frac{\log^2(1/\delta) \log^{3/2}(|\hat{C}_i|)}{\sqrt{|\hat{C}_i|}}.
\end{align*}
Therefore, the following inequality holds with probability at least $1 - 2\delta$,
\begin{align*}
    \frac{1}{K} \sum_{i\in \{1,2\}} \Big( {Z}_{\hat{C}_i}\big(\alpha_{h_i,\hat{C}_i}^\mathrm{CS-SAA},h_i\big) - {Z}_{\hat{C}_i}\big(\alpha_{h_i,\hat{C}_i}^\mathrm{OR},h_i\big) \Big) \le \sum_{i\in \{1,2\}} \frac{|\hat{C}_i|}{K}  \tilde{A}_i \max\{\Pi_2,L\sqrt{\Pi_2/\gamma}\}\cdot \frac{\log^2(1/\delta) \log^{3/2}(|\hat{C}_i|)}{\sqrt{|\hat{C}_i|}}.
\end{align*}
Furthermore, for the right-hand side of the above inequality, we have,
\begin{align*}
  \sum_{i\in \{1,2\}} \tilde{A}_i \max\{\Pi_2,L\sqrt{\Pi_2/\gamma}\}  \sqrt{\frac{|\hat{C}_i|}{K}}  \frac{\log^2(1/\delta) \log^{3/2}(|\hat{C}_i|)}{\sqrt{K}}
   \le A \max\{\Pi_2,L\sqrt{\Pi_2/\gamma}\}\cdot \frac{\log^2(1/\delta) \log^{3/2}(K)}{\sqrt{K}},
\end{align*}
where $A = \max\{\tilde{A}_1, \tilde{A}_2\}$. The last inequality holds because $\sqrt{\frac{|\hat{C}_i|}{K}} \le 1$ and $\log^{3/2}(x)$ is an increasing function when $x \ge 1$.

For the difference between the last two terms in the equation that decomposes the sub-optimality, we have,
\begin{align*}
    &\frac{1}{K}\sum_{i\in \{1,2\}} \Big( {Z}_{\hat{C}_i}(\alpha_{h_i,\hat{C}_i}^\mathrm{OR},h_i) - {Z}_{C_i}(\alpha_{h_i,C_i}^\mathrm{OR},h_i) \Big) \\
    =  & \frac{1}{K}\sum_{i\in \{1,2\}}\Big({Z}_{\hat{C}_i}(\alpha_{h_i,\hat{C}_i}^\mathrm{OR},h_i) -  {Z}_{\hat{C}_i}(\alpha_{h_i,C_i}^\mathrm{OR},h_i) +  {Z}_{\hat{C}_i}(\alpha_{h_i,C_i}^\mathrm{OR},h_i)  -  {Z}_{C_i}(\alpha_{h_i,C_i}^\mathrm{OR},h_i) \Big) \\ 
   \le  &\frac{1}{K} \sum_{i\in \{1,2\}} \sum_{k = 1}^{K}\mathds{1}_{k\in C_i,k \in \hat{C}_{3-i}}\Big(\mathbb{E}_{\xi_k}\big[c_k(x_k(\alpha_{h_{3-i},C_{3-i}}^\mathrm{OR},h_{3-i}, {S}^p_k),\xi_k) - c_k(x_k(\alpha_{h_i,C_i}^\mathrm{OR},h_i,{S}^p_k),\xi_k)\big]  \Big) \\
    \le & \frac{2\Pi_2}{K}\sum_{i\in \{1,2\}}\sum_{k = 1}^{K}\mathds{1}_{k\in C_i,k \in \hat{C}_{3-i}} = 2(R_{1,2}+R_{2,1})\Pi_2.
\end{align*}
The first inequality holds because ${Z}_{\hat{C}_i}(\alpha_{h_i,\hat{C}_i}^\mathrm{OR},h_i) \leq  {Z}_{\hat{C}_i}(\alpha_{h_i,C_i}^\mathrm{OR},h_i)$ by definition of the oracle's shrinkage parameter. The second inequality holds by Assumption \ref{asmp_4}. 
Recall $R_{i,j} = \frac{1}{K} \sum_{k=1}^K \mathds{1}(k \in C_i, k\in \hat{C}_j), i = 1,2, j =3-i$ indicates the portion of problems that belongs to cluster $i$ but grouped into cluster $j$. The key step now is to provide an upper bound of the sum of ratios $R = R_{1,2}+R_{2,1}$. Based on Hofflding's inequality (Lemma \ref{lma_hoffding_inequality}), we have,
\begin{align*}
    P\Big(R - \mathbb{E}[R] \ge t\Big)  = P\Big(R_{1,2} + R_{2,1} - \mathbb{E}(R_{1,2}+R_{2,1}) \ge t \Big) \le \exp(- 2Kt^2).\label{eqn_G_bound} 
\end{align*}
Thus, for any $0<\delta<1/2$, we have, with probability at least $1-\delta$, 
\begin{equation}
    R \le \mathbb{E}[R] + \frac{\sqrt{2}\log^{1/2}(1/\delta)}{2\sqrt{K}}. \label{prob_bound}
\end{equation}

Next, we need to bound the term $\mathbb{E}[R]$. We first bound the term $\mathbb{E}[R_{1,2}]$, and by symmetry, the bound can be applied to $\mathbb{E}[R_{2,1}]$. For ease of exposition, we define the event  $\mathcal{E} = \Big\{\frac{1 }{K}\sum_{k\in \mathcal{K}} \mu_k > \frac{a+b+d(b-a)}{2} - t \Big\} $ for some $t$ such that $0 \le t \le \frac{d(b-a)}{2}$. Consequently, the complement of the event $\mathcal{E}$ is denoted as  $\mathcal{\bar{E}}$. Let $D_k = \frac{1}{N_1}\sum_{i = 1}^{N_1} \hat{\xi}_{k,i} - \frac{1}{K}\sum_{k^{'}\in \mathcal{K}}\frac{1}{N_1} \sum_{i=1}^{N_1} \hat{\xi}_{k^{'},i}$. By the definition of the clustering algorithm \ref{alg_Cluster}, we have,
\begin{align}
    \mathbb{E}[R_{1,2}] & = \mathbb{E}\Big[\frac{1}{K}\sum_{k = 1}^{K}\mathds{1}_{k\in C_1,k \in \hat{C}_{2}}\Big] \\
    &= \frac{1}{K}\sum_{k \in C_1} P\Big(\frac{1}{N_1}\sum_{i = 1}^{N_1} \hat{\xi}_{k,i}\ge \frac{1}{K}\sum_{k^{'}\in \mathcal{K}}\frac{1}{N_1} \sum_{j=1}^{N_1} \hat{\xi}_{k^{'},j}\Big)
     \\
      &=  \frac{1}{K}\sum_{k \in C_1} \Big( P\Big(D_k \ge 0\Big| \mathcal{E}\Big)P(\mathcal{E}) +   P\Big(D_k \ge 0\Big| \mathcal{\bar{E}}\Big)P(\mathcal{\bar{E}})\Big).\label{eqn_R12}
\end{align}
Let $Y_{k,i} = \hat{\xi}_{k,i} - \frac{1}{K}\sum_{k^{'}\in \mathcal{K}}\hat{\xi}_{k^{'},i}$. For the first part of Equation \eqref{eqn_R12}, we have,
\begin{align*}
    \frac{1}{K}\sum_{k \in C_1} P\Big(\mathcal{D}_k \ge 0\Big| \mathcal{E}\Big)  & =  \frac{1}{K}\sum_{k \in C_1}P\Big(D_k - \mathbb{E}[D_k]\ge -  \mathbb{E}[D_k] \Big| \mathcal{E}\Big) \\
    &\le \frac{1}{K}\sum_{k \in C_1}P\Big(\frac{1}{N_1}\sum_{i = 1}^{N_1}Y_{k,i} -  \mathbb{E}\Big[\frac{1}{N_1}\sum_{i = 1}^{N_1}Y_{k,i}\Big]\ge \frac{a+b+d(b-a)}{2} - t -\mu_k \Big| \mathcal{E}\Big)\\
   & \le  \frac{1}{K}\sum_{k \in C_1}\exp\Big(-N_1\Big(\frac{a+b+d(b-a)}{2} - t -\mu_k\Big)^2/2\Pi_1^2\Big) \\
   & \le \frac{|C_1|}{K}\exp\Big(\frac{-N_1(d(b-a)-2t)^2}{8\Pi_1^2}\Big).
\end{align*}
The first inequality holds because, $\mathbb{E}[D_k] = \mu_k - \frac{1}{K}\sum_{k^{'}\in \mathcal{K}}\mu_{k^{'}}$. Thus, when $k \in C_1$ and the event $\mathcal{E}$ is true, we have,  $-\mathbb{E}[D_k] > \frac{a+b+d(b-a)}{2} - t -\mu_k \ge 0$. The second inequality holds due to Hoeffding's inequality (Lemma \ref{lma_hoffding_inequality}) by defining the random variable  $Y_{k, i}$.

For the second part of Equation \eqref{eqn_R12},
by definition of $\overline{\mathcal{E}}$, we first observe that,
\begin{align*}
     \frac{1}{K}\sum_{k \in C_1}P\Big(\mathcal{D}(d,k) \ge 0\Big| \mathcal{\bar{E}}\Big)\le \frac{|C_1|}{K}.
\end{align*}
In addition, we have,
\begin{align*}
    P(\overline{\mathcal{E}}) &= P\Big( \frac{1}{K}\sum_{k\in \mathcal{K}} \mu_k  \le \frac{a+b+d(b-a)}{2} - t \Big)\\
    &\le P\Big( \Big|\frac{a+b+d(b-a)}{2} -\frac{ 1 }{K} \sum_{k\in \mathcal{K}} \mu_k \Big| \ge t \Big) \\
    &\le 2\exp\Bigg(\frac{-2Kt^2}{(d+1)^2(b-a)^2}\Bigg).
\end{align*}
The last inequality holds because, by Assumption \ref{asmp_2}, $\mu_k$ is uniformly sampled from the intervals, $(a, \frac{a+b}{2})\cup (\frac{a+b}{2}+d(b-a), b + d(b-a))$ with mean equal to $\frac{a+b+d(b-a)}{2}$.

Thus, combining the above analysis, we present an upper bound of $\mathbb{E}[R_{1,2}]$,
\begin{align*}
    \mathbb{E}[R_{1,2}] & \le  \frac{|C_1|}{K}\exp\Big(\frac{-N_1(d(b-a)-2t)^2}{8\Pi_1^2}\Big) + \frac{|C_1|}{K}2\exp\Big(\frac{-2Kt^2}{(d+1)^2(b-a)^2}\Big)\\
      &  \le  \frac{|C_1|}{K}\exp\Big(\frac{-N_1d^2(b-a)^2}{32\Pi_1^2}\Big) + \frac{|C_1|}{K}2\exp\Big(\frac{-Kd^2}{8(d+1)^2}\Big).
\end{align*}
The second inequality holds because the upper bound increases in $t$ and we set the $t = \frac{d(b-a)}{4}$.
Note that the upper bound decreases in $d$ and $K$, that is, the more distant the two intervals are and the larger the number of problems is, the less likely the problem is grouped into a misclassified cluster.
By symmetry, we can bound the expectation $\mathbb{E}[R] = \mathbb{E}[R_{1,2}] + \mathbb{E}[R_{2,1}]$. Combined with Equation \eqref{prob_bound}, we have, with probability at least $1-\delta$, 
\begin{align*}
    R & \le \mathbb{E}[R] + \frac{\sqrt{2}\log^{1/2}(1/\delta)}{2\sqrt{K}}\\
     & \le \exp\Big(\frac{-N_1d^2(b-a)^2}{32\Pi_1^2}\Big) + 2\exp\Big(\frac{-Kd^2}{8(d+1)^2}\Big) + \frac{\sqrt{2}\log^{1/2}(1/\delta)}{2\sqrt{K}}.
\end{align*}
Based on the above analysis, there exist universal constants $\tilde{A}_1, \tilde{A}_2$ such that for any $0<\delta<1/3$, with probability at least $1-3\delta$ such that:
\begin{align*}
    &\text{SubOpt}_{h_1,h_2}(\alpha_{h_1,\hat{C}_1}^\mathrm{CS-SAA},\alpha_{h_2,\hat{C}_2}^\mathrm{CS-SAA})\\
    \le & \max\{\tilde{A}_1, \tilde{A}_2\} \max\{\Pi_2,L\sqrt{\Pi_2/\gamma}\}\cdot \frac{\log^2(1/\delta) \log^{3/2}(K)}{\sqrt{K}}+ \sqrt{2}\Pi_2\frac{\log^{1/2}(1/\delta)}{\sqrt{K}}\\
    &  + 2\Pi_2\exp\Big(\frac{-N_1d^2(b-a)^2}{32\Pi_1^2}\Big) + 4\Pi_2\exp\Big(\frac{-Kd^2}{8(d+1)^2}\Big).
\end{align*}
By defining the constant $A = \max\{\tilde{A}_1, \tilde{A}_2\} \max\{\Pi_2,L\sqrt{\Pi_2/\gamma}\}$ and scaling $\delta$, we obtain the upper bound in Theorem \ref{thm_GeneralCost}. This completes the proof.
\QED

\section{Additional Numerical Experiments}\label{appx_Numerical}

\begin{figure}[htbp!]
    \centering
     \includegraphics[width=0.7\linewidth]{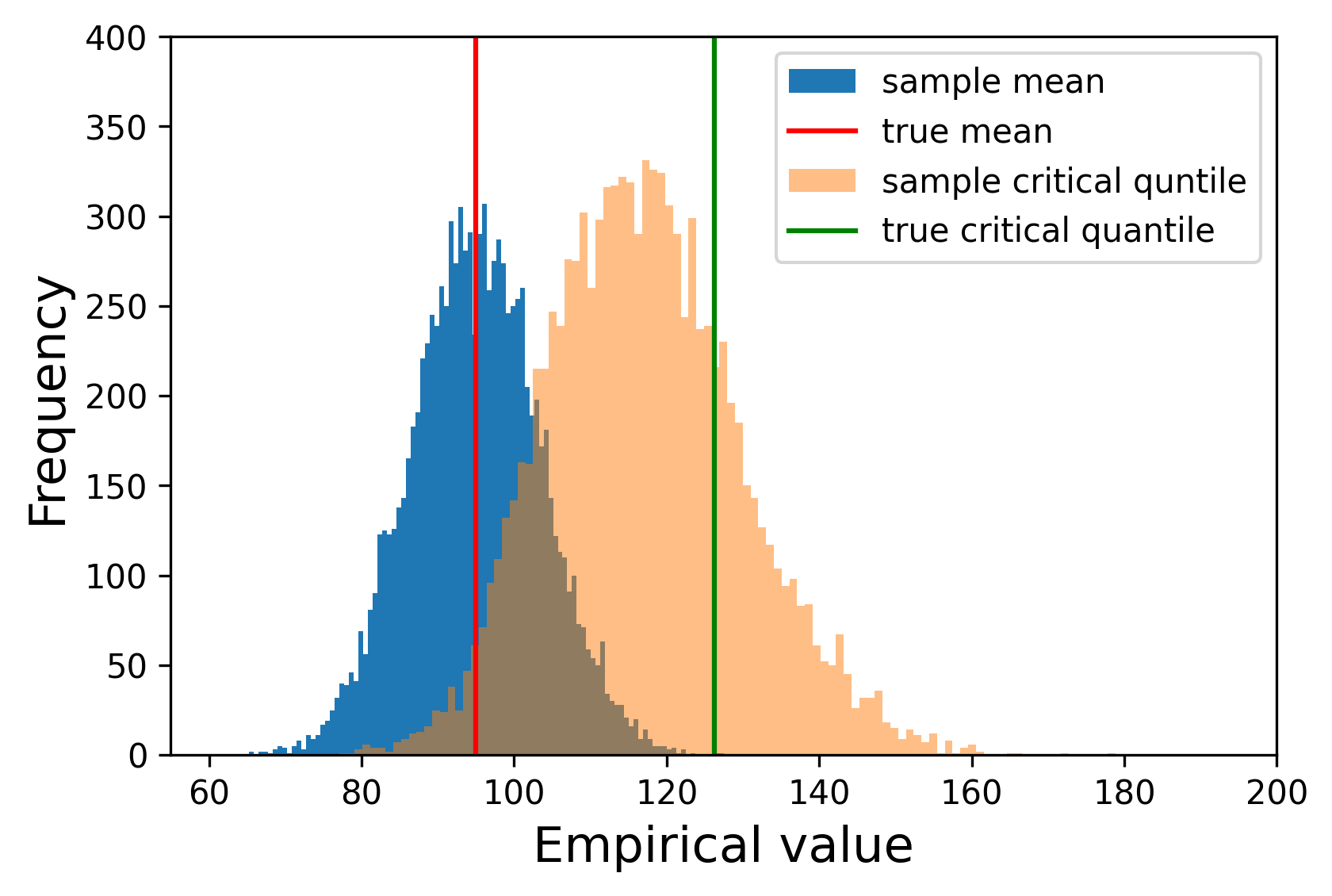}
    \caption{Empirical value for sample mean and sample quantile}
    \label{fig:hist_mean_quantile}
    {\footnotesize Note.  $\mu_k = 95$, $\xi_k \sim N(\mu_k,(\mu_k*0.2)^2)$ and $N_1 = 5$}
\end{figure}

We conducted an additional experiment for the problem with $\mu_k = 95$ and $v_k =0.2$. Thus, the critical quantile for $s=0.95$ is 126. We sample $N_1 = 5$ points from the distribution $N(\mu_k, (v_k \mu_k)^2)$, and compute the sample mean and the critical quantile based on the empirical distribution. We repeat this procedure for 10,000 iterations, and report the numerical results in Figure  \ref{fig:hist_mean_quantile}. The sample mean provides an unbiased estimate with lower variance, whereas the sample quantile is subject to bias with a larger variance. Since the sample mean provides a minimum-variance unbiased estimate of the true mean, when the CVs are the same, the data-driven clustering outcomes based on the sample means should be closer to the oracle cluster structure under Newsvendor loss than that of the critical quantile based metric.

\begin{figure}[htbp!]
    \centering
     \includegraphics[width=0.8\linewidth]{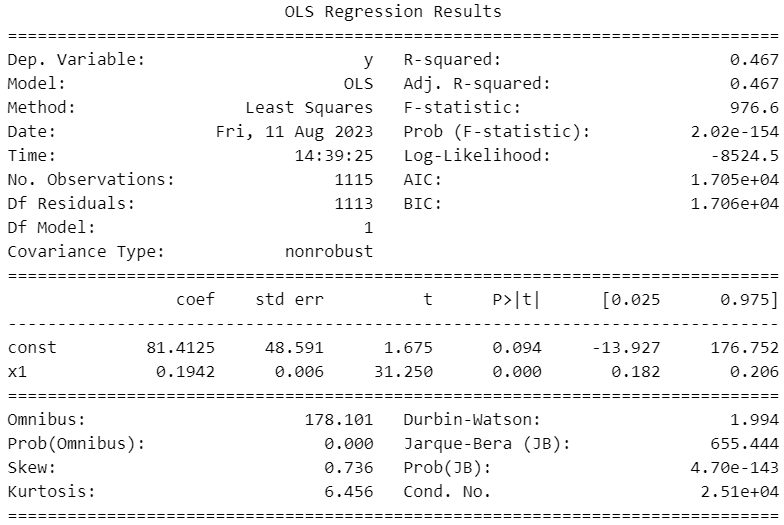}
    \caption{OLS results}
    \label{fig:ols_results}
\end{figure}

Illustrated in Figure \ref{fig:sample_mean_stde}, it becomes evident that the sample mean and sample standard deviation display a linear correlation, underscoring the close alignment of the coefficient of variation across all problems. To further investigate, an ordinary least squares analysis is conducted, employing the sample mean as the independent variable and the sample standard deviation as the dependent variable.\footnote{One can adopt more sophisticated tests to determine whether the CVs of random demands are similar \citep[]{FeltzMiller1996, KrishnamoorthyLee2014}.} The summary results of OLS are shown in Figure \ref{fig:ols_results}. The results reveal a significant parameter of approximately 0.2. Thus, we believe that using the sample mean as the metric for clustering will bring higher benefits.

%
%
%






\clearpage
\section{Table of Notations}
\begin{table}[htbp!]
    \centering
    \renewcommand\arraystretch{1.4}
    \begin{tabular}{ll}
    \hline
    Notation    & Definition \\\hline
    $K$ & number of problems\\
    $\mathcal{K}$ & set of all problems\\
    $N$ & number of data samples for each problem\\
    $N_1$ & number of data samples allocated for clustering\\
    $C_i\mbox{ }(\hat{C}_i)$ & set of problems that belong to the (estimated) cluster indexed by $i$\\
    $\xi_k$ & the random variable of the $k$-th problem\\
    $\mathbb{P}_k$ & the probability measure of the random variable of the $k$-th problem\\
    $\mu_k \mbox{ }(\sigma_k^2)$    & mean (variance) of the random variable of the $k$-th problem\\
    $x_k$ & the decision variable of the $k$-th problem\\
    $\mathcal{X}_k$ & the feasible region of the $k$-th problem\\
    $c_k(x_k,\xi_k)$ & the cost associated with decision $x_k$ and realized random variable $\xi_k$ for the $k$-th problem\\
    $S_k$ & the set of data samples of $k$-th problem \\
    $S_k^c \mbox{ }(S_k^p)$ & the set of data samples of $k$-th problem that are used for clustering (pooling)\\
    $\mu_0\mbox{ }(\mu_{i,0})$ & the anchor mean for all problems $\mathcal{K}$ (a cluster of problems $C_i$ or $\hat{C}_i$)\\
    $\mu_0^{\mathrm{AP}} \mbox{ }(\mu_{i,0}^{\mathrm{AP}})$ & the optimal a priori anchor mean for all problems $\mathcal{K}$ (problems of a true cluster $C_i$)\\
    $\mu_{i,0}^\mathrm{AP}(\hat{C}_i)$ & the optimal a priori anchor mean for problems in the estimated cluster $\hat{C}_i$\\
    $\alpha \mbox{ }(\alpha_i)$ & shrinkage parameter for all problems (problems in cluster $C_i$) for data pooling\\
    $\alpha^\mathrm{AP}\mbox{ }(\alpha_{i}^\mathrm{AP})$ & the optimal a priori shrinkage parameter for all problems $\mathcal{K}$ (problems in the true cluster $C_i$)\\
    $\alpha_{i}^\mathrm{AP}(\hat{C}_i)$ & the optimal a priori shrinkage parameter for the problems in the estimated cluster $\hat{C}_i$\\
    $\hat{\mu}_0\mbox{ }(\hat{\mu}_{i,0}(\hat{C}_i))$ & the data-driven anchor mean for all problems (the problems in the estimated cluster $\hat{C}_i$)\\
    $\hat{\alpha}_i\mbox{ }(\hat{\alpha}_i(\hat{C}_i))$ & the data-driven shrinkage parameter for all problems (the problems in the estimated cluster $\hat{C}_i$)\\
    $x_k(\alpha,\mu_0,\hat{\mu}_k)$ & the decision for problem $k$ given  shrinkage parameter $\alpha$, anchor mean $\mu_0$ and sample mean $\hat{\mu}_k$\\
    $Z_{\mathrm{SAA}}$ & expected out-of-sample cost of all problems using SAA \\
    $Z(\alpha,\mu_0, C)$ & expected out-of-sample cost of the set of problems $C$ of direct data pooling with $S$\\
    $Z^{p}(\alpha,\mu_0,C)$ & expected out-of-sample cost of the set of problems $C$ of direct data pooling with $S^p$\\
    $R_{i,j}$ &  the percentage of problems that belong to Cluster $i$ but are grouped into Cluster $j$ \\
    $r_{i,j}$ & the limit of $R_{i,j}$ when $ K \to \infty$
    \\\hline
    \end{tabular}
    \caption{Table of Notations}
    \label{T_Notation}
\end{table}

\end{document}